\documentclass[11pt,final,letterpaper,onecolumn]{IEEEtran}

\usepackage{setspace}
\usepackage[pagebackref=true,breaklinks=true,letterpaper=true,colorlinks,bookmarks=false]{hyperref}
\usepackage{epsfig}
\usepackage{graphicx}
\usepackage{amsmath}
\usepackage{amssymb}
\usepackage{amsthm}
\usepackage{subfig}

\newcommand{\ud}{\,\mathrm{d}}

\newcommand{\R}{\mathbb{R}}

\newcommand{\ip}[3]{\left< {#1}, {#2} \right>_{#3}}

\newcommand{\mean}[1]{\mbox{avg}({#1})}

\newcommand{\comment}[1]{ }
\newcommand{\der}[2]{\frac{\ud {#1}}{\ud {#2}}}

\newtheorem{defn}{Definition}

\begin{document}

\title{Shape Tracking With Occlusions via Coarse-To-Fine Region-Based Sobolev Descent}

\author{Yanchao Yang \, and \, Ganesh Sundaramoorthi %
\IEEEcompsocitemizethanks{\IEEEcompsocthanksitem 
  Y.~Yang is with the Department of Electrical Engineering,
  King Abdullah University of Science and Technology (KAUST), Thuwal, Saudi Arabia 
\IEEEcompsocthanksitem
G.~Sundaramoorthi is with the Department of Electrical Engineering and Department of Applied Mathematics and Computational Science, King Abdullah University of Science and Technology (KAUST), Thuwal, Saudi Arabia
  \protect\\
  E-mail: \{yanchao.yang, ganesh.sundaramoorthi\}@kaust.edu.sa}
}
\maketitle

\begin{abstract}
We present a method to track the precise shape of an object in video
based on new modeling and optimization on a new Riemannian manifold of
parameterized \emph{regions}.

Joint dynamic shape and appearance models, in which a template of the
object is propagated to match the object shape and radiance in the
next frame, are advantageous over methods employing global image
statistics in cases of complex object radiance and cluttered
background. In cases of 3D object motion and viewpoint change,
self-occlusions and dis-occlusions of the object are prominent, and
current methods employing joint shape and appearance models are unable
to adapt to new shape and appearance information, leading to
inaccurate shape detection. In this work, we model self-occlusions and
dis-occlusions in a joint shape and appearance tracking framework.

Self-occlusions and the warp to propagate the template are coupled,
thus a joint problem is formulated. We derive a coarse-to-fine
optimization scheme, advantageous in object tracking, that initially
perturbs the template by coarse perturbations before transitioning to
finer-scale perturbations, traversing all scales, seamlessly and
\emph{automatically}. The scheme is a gradient descent on a novel
infinite-dimensional Riemannian manifold that we introduce. The
manifold consists of planar parameterized \emph{regions}, and the
metric that we introduce is a novel Sobolev-type metric defined on
infinitesimal vector fields on regions. The metric has the property of
resulting in a gradient descent that automatically favors coarse-scale
deformations (when they reduce the energy) before moving to
finer-scale deformations.

Experiments on video exhibiting occlusion/dis-occlusion, complex
radiance and background show that occlusion/dis-occlusion modeling
leads to superior shape accuracy compared to recent methods employing
joint shape/appearance models or employing global statistics.
\end{abstract}

\markboth{Technical Report}%
{Yang and Sundaramoorthi: Coarse-to-Fine Region-Based Sobolev Descent}

\maketitle

\IEEEpeerreviewmaketitle

\section{Introduction}

In many video processing applications, such as post-production of
motion pictures, it is important to obtain the precise \emph{shape} of
the object of interest at each frame in a video. Although many methods
have been proposed, much work remains. Many existing tracking methods
(e.g.,
\cite{rathi2007tracking,cremers2006dynamical,bibby2010real,Fan:PAMI12})
are built on top of partitioning the image into foreground and
background based on global image statistics (e.g., color
distributions, edges, texture, motion), which is advantageous in
obtaining shape of the object. However, in tracking objects with
complex radiance and cluttered background, partitioning the image
based on global statistics may not yield the object as a partition. An
alternative approach is to deform a template (the radiance function
defined on the region of the projected object) to match the object in
shape and radiance in the next frame (the deformed shape yields the
object of interest). We will refer to this alternative approach as
\emph{joint shape/appearance matching}.

A difficulty in tracking by joint shape/appearance matching is that 3D
object and/or camera motion imply that parts of the object come into view
(\emph{dis-occlusions}) and go out of view (\emph{occlusions});
therefore, an initially accurate template, even when warped through a
non-rigid deformation, becomes an inaccurate model of the object in
later frames. Thus, it is necessary to update the template by removing
occluded regions and including dis-occluded regions.

In this work, we model self-occlusions and dis-occlusions in
\emph{tracking by joint shape/appearance matching}.  Small frame rate
implies moderately large non-rigid deformation of the projected object
between frames. Thus, we represent the large non-rigid warp as an
integration of a time-varying vector field defined on \emph{evolving
  region (or domain of interest)}.  Since an occlusion is the part of
the template that does not correspond to the next frame, occlusions
and the deformation are coupled, and thus, a joint optimization
problem in the large deformation and occlusion is setup, and a simple,
efficient algorithm is derived. We note that dis-occlusions can be
detected only with priors on the object. We show how to use a prior
that the object radiance is self-similar, so that dis-occluded regions
between frames can be detected by measuring image similarity to the
current template. To ensure robust estimates of the object's radiance
across frames, recursive filtering is used.

In order to perform optimization in the warp and the occlusion, we
introduce a novel coarse-to-fine optimization scheme that is
well-suited for object tracking. The scheme is \emph{simply} a
gradient descent on a novel Riemannian manifold that we introduce. The
(infinite-dimensional) manifold consists of parametric \emph{regions},
represented as warps from an initial region to arbitrary regions
defined in the plane. The choice of regions is particularly suited for
object tracking as the object in the imaging plane is described by
both its shape and its radiance function, the later defined on the
\emph{region} in the imaging plane determined by all points on the 3D
object that project into the imaging plane. The Riemannian metric that
we introduce is defined on vector fields on regions, and it is a
Sobolev-type metric. We show how to compute gradients of energies
defined on warps with respect to this metric, and that a gradient
descent with respect to this metric leads to the extremely beneficial
property for tracking: an initial region to match an unknown subset of
an image deforms in a coarse-to-fine fashion, initially moving the
region according to a coarse-scale deformations before transitioning
continuously and seamlessly to finer-scale deformations.  This
coarse-to-fine property is inherent in the gradient descent and no
methodology (other than the gradient computation) is needed to enforce
this property.

\subsection{Key Contributions}
Our main contributions are two-fold: modeling and theory. First, we
formulate self-occlusions and dis-occlusions in tracking by joint
shape/appearance matching. Occlusions have been modeled in shape
tracking, but existing works do so either in a framework with simpler
models of radiance (e.g., \cite{bibby2010real}), i.e., color
histograms, or are layered models with complex radiance (e.g.,
\cite{jackson2008dynamic}) that can cope with occlusions of one layer
on another, but not \emph{self}-occlusions or dis-occlusions.  We also
solve dis-occlusions with the similarity prior mentioned above.

The second main contribution is the novel general optimization scheme
for determining the warp and occlusion that has an automatic
coarse-to-fine behavior. This new optimization technique is based on
new theoretical advances, including our novel Riemannian manifold of
regions, and a novel Sobolev-type metric on infinitesimal
perturbations of regions.

\subsection{Related Work: Tracking and Occlusions}

Most shape tracking techniques (e.g.,
\cite{isard1998condensation,rathi2007tracking,cremers2006dynamical,bibby2010real})
extend image segmentation techniques such as active contours (e.g.,
\cite{kass1988snakes,caselles1997geodesic,kichenassamy1995gradient,paragios2002geodesic,chan2001active}). These
techniques build on discriminating the foreground and background using
global image statistics (e.g., color distributions, texture, edges,
motion). However, when the object has complex radiance and is within
cluttered background, discriminating global image statistics leads to
errors in the segmentation. Some methods try to resolve this issue by
using local statistics (e.g.,
\cite{mumford1989optimal,lankton2008localizing,bai2009video}). Other
methods use temporal consistency to predict the object location /
shape in the next frame (e.g.,
\cite{isard1998condensation,rathi2007tracking,niethammer2008geometric,sundaramoorthi2011new})
to provide better initialization to frame partitioning. In
\cite{cremers2006dynamical} (based on a dynamic extension of active
shape and appearance models\cite{cootes1995active}), dynamics of shape
are modeled from training data, constraining the solution of frame
partitioning; however, training data is only available in restricted
scenarios. While providing improvements, images with complex object
radiance and cluttered background still pose a significant challenge.

We approach shape tracking by joint shape/appearance matching. We use
a radiance model that is a dense function defined on the projected
object. Dense radiance functions have been used (e.g.,
\cite{black1998eigentracking,hager1998efficient}) for tracking via
matching to the next frame. However, they are box trackers, and do not
provide shape. In \cite{jackson2008dynamic,bai2010dynamic}, a joint
model of radiance and shape of the object \emph{and} background is
used, however, \emph{self}-occlusions and \emph{dis-occlusions} are
not modeled.

Occlusions have been considered in optical flow. In
\cite{alvarez2002symmetrical,ben2007variational}, forward and backward
optical flows are computed, and the occluded region is the set where
the composition of these flows is not the identity. In
\cite{strecha2004probabilistic,xiao2006bilateral}, an occlusion is the
set where the optical flow residual is large. In
\cite{sundberg2011occlusion}, occlusion boundaries are detected by
discontinuities of optical flow. In \cite{ayvaci2011sparse}, joint
estimation of the optical flow and occlusions is performed. In
\cite{ricco2012dense}, dense trajectory estimation across multiple
frames with occlusions is solved. We use ideas of occlusions in
\cite{ayvaci2011sparse}, and apply them to shape tracking where
additional considerations must be made for evolving the shape,
dis-occlusions, and larger deformations.

\subsection{Related Work: Shape Metrics}

The optimization technique for joint warp, occlusion, and region
estimation that we introduce is a gradient descent on a Riemannian
manifold, and thus our work relates to the literature on \emph{shape
  metrics} by modeling shapes on a Riemannian manifold.  The
literature on shape is large, and we do not give a full survey, only
the most relevant works of shape based on Riemannian manifolds. There
have been two primary uses for shape metrics. One is \emph{shape
  optimization}, that is, minimization of energies defined on shapes
usually to segment shapes from images. The other is \emph{shape
  matching and analysis}, i.e., computing distances and morphs between
given shapes (usually already segmented from images) or decomposing
given shapes into constituent components (e.g., via a PCA) that is
made possible by a shape metric.

Active contours (e.g.,
\cite{kass1988snakes,caselles1997geodesic,kichenassamy1995gradient,chan2001active,paragios2000geodesic}),
where shape is defined as a planar contour, are an instance of shape
optimization. Active contours are usually based on a gradient descent
of an energy, and the gradient depends on a choice of a metric on
perturbations of planar contours. The metric implicitly chosen is a
geometric $\mathbb{L}^2$ metric. Other metrics for active contours
were considered by
\cite{sundaramoorthi2007sobolev,charpiat2007generalized}, in
particular, Sobolev-type metrics on contours, which favor spatially
regular flows for gradient descent, and typically avoid un-desirable
local minima due to fine-scale structures of an image. In
\cite{sundaramoorthi2008coarse}, it was shown that Sobolev-type
metrics are ideally suited for tracking applications since they have
an automatic coarse-to-fine property in comparison to the
$\mathbb{L}^2$ metric. The novel Riemannian metric that we introduce
in this paper, is motivated by the coarse-to-fine property noticed in
\cite{sundaramoorthi2008coarse}. However, the energies considered in
this paper are \emph{not defined on contours}; they are defined on
parameterized \emph{regions} (since we are interested in both shape
and the radiance function of the object, which is defined on the
interior of a \emph{region}), and the framework of
\cite{sundaramoorthi2008coarse} does not apply. Thus, we define a new
Riemannian manifold and a Sobolev-type metric on parameterized regions
(i.e., warps of a region to arbitrary regions).

In shape matching and analysis, several Riemannian metrics have been
proposed. In \cite{klassen2004analysis}, an $\mathbb{L}^2$ Riemannian
metric is proposed on the tangent vector field of planar curves. In
\cite{michor2007metric,michor2007overview}, Sobolev-type metrics are
proposed on planar curves, which induces meaningful shape morphings as
geodesic paths (shortest paths where paths are defined in the manifold
of shapes), unlike the standard $\mathbb{L}^2$ metric, which does not
yield geodesics \cite{michor2003riemannian,yezzi2005conformal}. The
work of deformable templates
\cite{beg2005computing,miller2006geodesic} defines a Riemannian
manifold on the space of warps (diffeomorphisms) from the entire
domain of the image to itself, and shape matching can be performed by
diffeomorphisms that map a characteristic function of one shape
defined on the entire domain of the image onto another. Sobolev
metrics on vectors fields of the fixed domain are defined, and
geodesic paths are computed.

Our work relates to deformable templates, since we also define a
Riemannian metric on a space of warps, but there are two differences
in our mathematical framework (besides the obvious fact that we are
interested in object tracking rather than image registration or shape
matching of already segmented shapes as in
\cite{beg2005computing,miller2006geodesic}). First, our set of warps
are defined on a region of an object (not the entire image) to all
regions (compact subsets) in the imaging domain. This choice is used
because we model only the object of interest. Modeling the entire
image is more difficult, as it consists of various different motions
(of other objects and the background). The smoothness assumption on
entire domain made in \cite{beg2005computing,miller2006geodesic} is
thus not appropriate for video from natural scenes where there are
discontinuities in deformation between boundaries of objects, but
rather to medical images where the deformation of the entire image can
be approximated by a globally smooth deformation. Moreover, occlusions
are not considered in \cite{beg2005computing,miller2006geodesic}.  The
second difference from \cite{beg2005computing,miller2006geodesic} is
that we are not interested in computing geodesic paths on the
Riemannian manifold of warps, rather we are interested in computing a
gradient descent on warps (in contrast to a gradient descent on
\emph{paths} of warps). The latter may be computationally more efficient
(since computing geodesics requires searching for a minimal path over
all paths, whereas a gradient descent simply chooses a path based on
the energy and the metric and does not solve an expensive optimization
problem over all possible paths), is simpler, and induces a
coarse-to-fine descent, which is extremely beneficial in object
tracking.

Lastly, the work of \cite{wirth2011continuum} introduces a
Sobolev-type Riemannian metric on regions for shape matching (e.g.,
computing geodesics between shapes) rather than \emph{shape
  optimization}, which is the focus of this work. We compute gradients
of energies defined on warps of regions, which is not considered in
\cite{wirth2011continuum}. The particular form of the Sobolev-metric
that we construct is different than \cite{wirth2011continuum} as it
has a natural decomposition of perturbations of a region into
translations and orthogonal deformations, which is well suited for
object tracking, and leads to convenience in the computation of the
gradient.

\subsection{Extensions to Conference Version}
This work is an extension of a preliminary conference paper
\cite{YangS2013occlusions}. One extension in this paper is a
significant theoretical advancement leading to the automatic
coarse-to-fine optimization scheme based on a novel region-based
Sobolev metric, which is extremely convenient in practice, in
particular, it is \emph{parameter free}. In contrast, the scheme in
\cite{YangS2013occlusions} was only an approximation
of the coarse-to-fine property, and not based on a unified
energy. Other extensions include extra experiments, analysis of
parameter sensitivity for occlusion and dis-occlusion energy
thresholds, and detailed numerical discretization.

\section{Dynamic Model of the Projected Object}
In this section, we give our dynamic model of the shape and radiance
of the 3D object projected in the imaging plane. From this, the notion
of occlusions and dis-occlusions is clear. The dynamic model is
necessary for the recursive estimation algorithm in
Section~\ref{sec:filtering}.

\begin{figure}
  \centering
  \includegraphics[clip,trim=0 220 390 120,totalheight=1.6in]{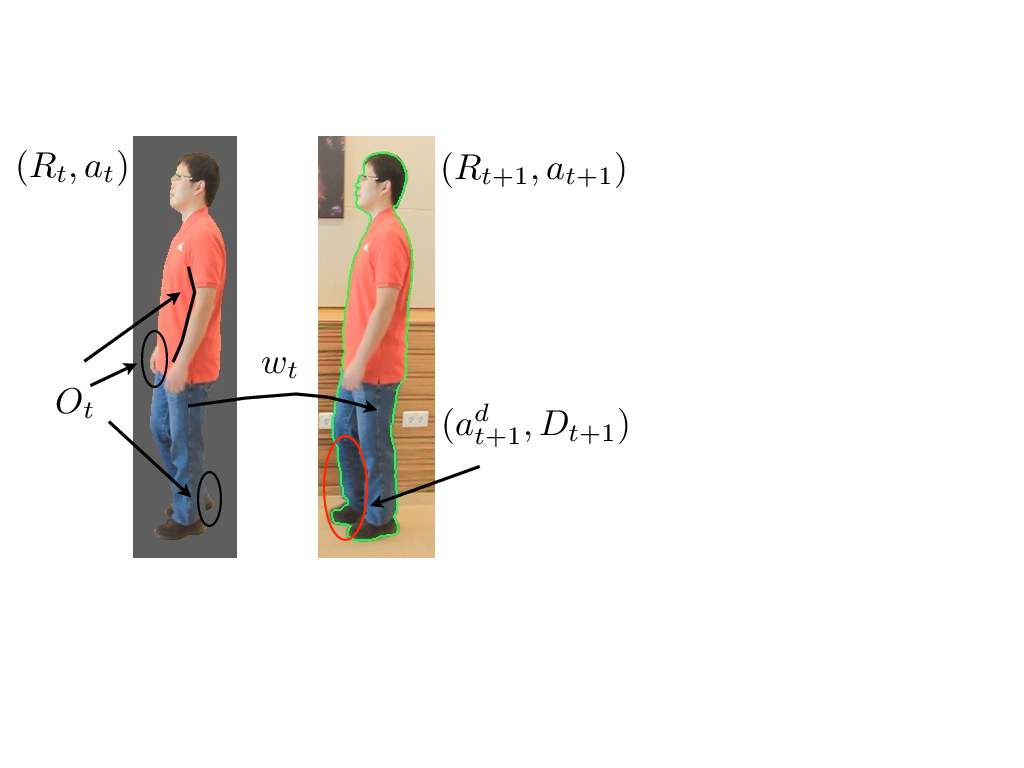}
  \caption{{\bf Diagram illustrating our dynamic model.} Left:
    template $(R_{t}, a_{t})$ (non-gray), right:
    $I_{t+1}$. Self-occlusions $O_{t}$, dis-occlusions $D_{t+1}$ and
    its radiance $a_d^{t+1}$, the region at frame $t+1$ is $R_{t+1}$
    (inside the green contour), and the warp is $w_t$, which is
    defined in $R_t\backslash O_{t}$. The curved black line is a
    self-occlusion since the arm moves towards the left.}
  \label{fig:schematic_occlusions_model}
\end{figure}

Let $\Omega \subset \R^2$, and $I : \{1, 2, \ldots, N\} \times \Omega
\to \R^k$ denote the image sequence ($N$ frames) that has $k$
channels. We denote frame $t$ by $I_t$.  The camera projection of
visible points on the 3D object at time $t$ is denoted by $R_t$, which
we refer to as ``shape'' or region. The projected object's radiance
is denoted $a_t$, and $a_t : R_t \to \R^k$. Our dynamic model of the
region and radiance (see Fig.~\ref{fig:schematic_occlusions_model}
for a diagram) is
\begin{align}
  R_{t+1} &= w_{t}( R_{t} \backslash O_{t} ) \cup D_{t+1}
  \label{eq:region_model}
  \\
  a_{t+1}(x) &= 
  \begin{cases}
    a_{t}(w_{t}^{-1}(x)) + \eta_t(x) & x\in w_{t}(R_{t} \backslash O_{t}) \\
    a^{d}_{t+1}(x) + \eta_t(x)& x \in  D_{t+1}
  \end{cases}
  \label{eq:appearance_model}
\end{align}%
where $O_t$ denotes the subset of $R_t$ that is occluded from view in
frame $t+1$, $D_{t+1}$ denotes the subset of the projected object that
is disoccluded (comes into view) at frame $t+1$, $a^{d}_{t+1} :
D_{t+1} \to \R^k$ is the radiance of the disoccluded region, and
$w_{t}$ maps points that are not occluded in $R_{t}$ to $R_{t+1}$ in
the next frame.  The warp $w_{t}$ is a diffeomorphism on the
\emph{un-occluded} region $R_{t}\backslash O_{t}$ (it will be extended
to all of $R_t$: see Section~\ref{sec:energy_occlusion} for details),
which is a transformation arising from viewpoint change and 3D
deformation.

The region $R_{t}\backslash O_{t}$, is warped by $w_{t}$ and the
dis-occlusion of the projected object, $D_{t+1}$, is appended to the
warped region to form $R_{t+1}$. The relevant portion of the
radiance, $a_{t}| (R_{t} \backslash O_{t})$ is transfered via the
warp $w_{t}$ to $R_{t+1}$ (as usual brightness constancy), noise
added, and then a newly visible radiance is obtained in
$D_{t+1}$. The noise models deviation from brightness constancy (e.g.,
non-Lambertian reflectance, small illumination change, noise, etc...).

{\bf Organization of the rest of the paper}: A template ($a_0, R_0$)
of the object is given. Our goal is, given an estimate of $R_t$,
$a_t$, and $I_{t+1}$ to estimate $R_{t+1}$ in $I_{t+1}$.  In
Section~\ref{sec:energy_occlusion}, we formulate an optimization
problem to determine $w_t$ and the occlusion $O_t$ given $a_t,R_t$,
and $I_{t+1}$. In Section~\ref{sec:energy_disocclusion}, we formulate
an optimization problem to determine the dis-occlusion $D_{t+1}$ given
$w_t(R_t\backslash O_t)$ and $I_{t+1}$. The joint energy for $w_t$ and
$O_t$ presented in Section~\ref{sec:energy_occlusion} involves an
alternating optimization. In Section~\ref{sec:sobolev_optimization},
we present a new general optimization scheme for energies defined on
warps, which requires introducing a new Riemannian manifold and a
novel Riemannian metric, a \emph{Sobolev-type region based metric},
whose corresponding gradient descent we show has a coarse-to-fine
property. This optimization scheme is a relevant sub-problem for the
energy of interest in Section~\ref{sec:energy_occlusion}, and the full
optimization scheme for the joint energy in the warp and occlusion is
presented in Section~\ref{sec:final_optimization_warp_occlusion}. The
optimization for the dis-occlusion energy is presented in
Section~\ref{subsec:disocclusion_optimization}. Finally, in
Section~\ref{sec:filtering}, we derive a recursive estimation
procedure and integrate all steps. See Fig.~\ref{fig:overview} for a
system overview.

\def\fHobExa{figures2/example_hobits}
\def\fHobExSa{0.79in}
\begin{figure*}
  \footnotesize
  \centering
  \comment{
  \begin{tabular}{c@{ }c|c@{ }c@{ }c|c|c}
    template & target &  warped template & occlusion map & occlusion removed &
    dis-occlusion & dis-occlusion added\\
    $(a_t,R_t)$ & $I_{t+1}$ & $a_t\circ w_t, \,w_t(R_t)$ & $w_t(O_t)$ &
    $w_t(R_t\backslash O_t)$ & $D_{t+1}$ & ($a_{t+1}, R_{t+1}$)\\
    \includegraphics[clip,trim=230 0 120 0,totalheight=\fHobExSa]{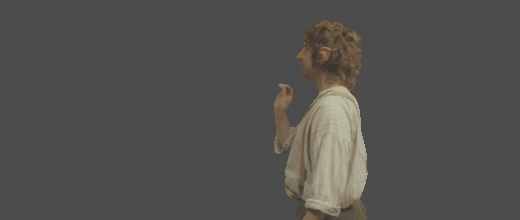} 
    &
    \includegraphics[clip,trim=230 0 120 0,totalheight=\fHobExSa]{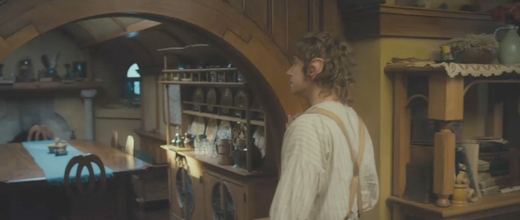}
    &
    \includegraphics[clip,trim=230 0 120 0,totalheight=\fHobExSa]{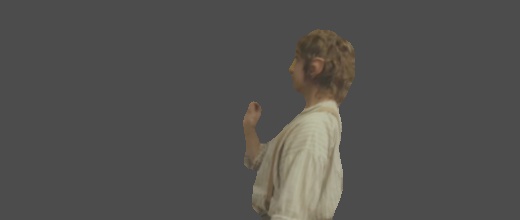}%
    \includegraphics[clip,trim=230 0 120 0,totalheight=\fHobExSa]{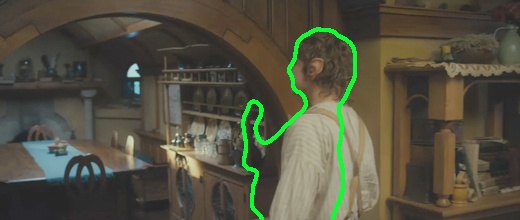}
    &
    \includegraphics[clip,trim=230 0 120 0,totalheight=\fHobExSa]{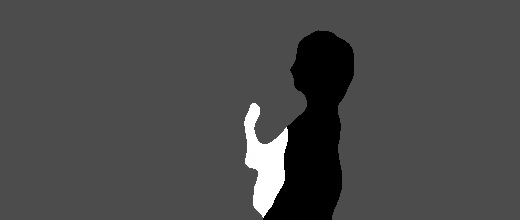}
    &
    \includegraphics[clip,trim=230 0 120 0,totalheight=\fHobExSa]{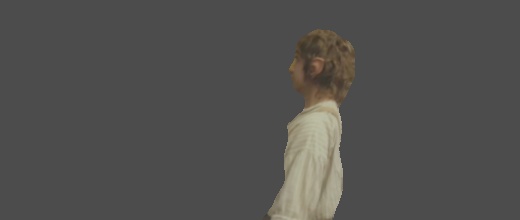}%
    \includegraphics[clip,trim=230 0 120 0,totalheight=\fHobExSa]{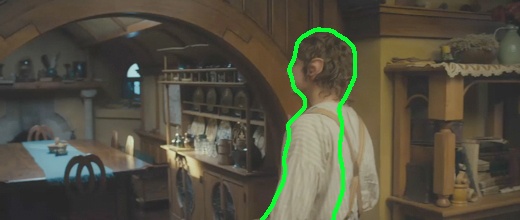}
    &
    \includegraphics[clip,trim=230 0 120 0,totalheight=\fHobExSa]{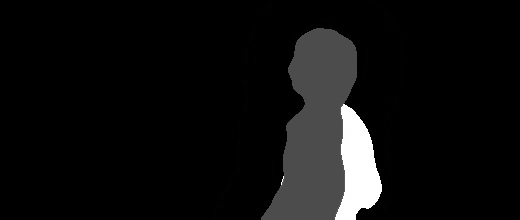}
    &
    \includegraphics[clip,trim=230 0 120 0,totalheight=\fHobExSa]{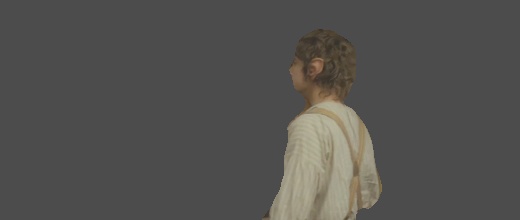}%
    \includegraphics[clip,trim=230 0 120 0,totalheight=\fHobExSa]{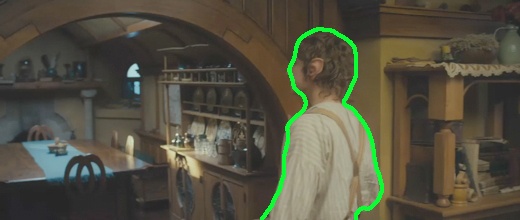} \\
    \multicolumn{2}{c|}{(a) Input} &  \multicolumn{3}{c|}{(b) Section~\ref{sec:occlusion}} &
    (c) Section~\ref{sec:disocclusions} &
    (d) Section~\ref{sec:filtering}
  \end{tabular}\\
}
  \begin{tabular}{c@{ }c|c@{ }c@{ }c}
    template & target &  warped template & occlusion map & occlusion removed \\
    $(a_t,R_t)$ & $I_{t+1}$ & $a_t\circ w_t, \,w_t(R_t)$ & $w_t(O_t)$ &
    $w_t(R_t\backslash O_t)$ \\
    \includegraphics[clip,trim=230 0 120 0,totalheight=\fHobExSa]{\fHobExa/a/Appearance_t} 
    &
    \includegraphics[clip,trim=230 0 120 0,totalheight=\fHobExSa]{\fHobExa/a/Image_t+1}
    &
    \includegraphics[clip,trim=230 0 120 0,totalheight=\fHobExSa]{\fHobExa/c/templateevolve/templateevolve076}%
    \includegraphics[clip,trim=230 0 120 0,totalheight=\fHobExSa]{\fHobExa/c/contourevolve/coutourevolve076}
    &
    \includegraphics[clip,trim=230 0 120 0,totalheight=\fHobExSa]{\fHobExa/c/converged_occlusion}
    &
    \includegraphics[clip,trim=230 0 120 0,totalheight=\fHobExSa]{\fHobExa/c/template_after_occlusiondetection}%
    \includegraphics[clip,trim=230 0 120 0,totalheight=\fHobExSa]{\fHobExa/c/region_after_occlusiondetection}
    \\
    \multicolumn{2}{c|}{(a) Input} &  \multicolumn{3}{c}{(b)}
  \end{tabular}\\\vspace{0.18in}
  \begin{tabular}{c|c}
    dis-occlusion & dis-occlusion added\\
    $D_{t+1}$ & ($a_{t+1}, R_{t+1}$) \\
    \includegraphics[clip,trim=230 0 120 0,totalheight=\fHobExSa]{\fHobExa/d/detected_disocclusion}
    &
    \includegraphics[clip,trim=230 0 120 0,totalheight=\fHobExSa]{\fHobExa/d/Final_Appearance}%
    \includegraphics[clip,trim=230 0 120 0,totalheight=\fHobExSa]{\fHobExa/d/Final_Region} \\
    (c)  & (d)
  \end{tabular}
  
  \caption{{\bf Illustration of frame processing in our algorithm.}  {\bf
      (a)}: Estimate at frame $t$ of the shape and radiance
    $(a_t,R_t)$, and the next image $I_{t+1}$. {\bf (b)}: Simultaneous
    non-rigid warping and occlusion estimation is performed (first
    image: warped template $a_t\circ w_t$, second: boundary of warped
    template in $I_{t+1}$, third: warped occlusion $w_t(O_t)$
    determined, fourth: warped template with warped occlusion removed
    $w_t(R_t\backslash O_t)$, fifth: boundary of $w_t(R_t\backslash
    O_t)$). {\bf (c)}: Dis-Occlusion $D_{t+1}$ in $I_{t+1}$ determined
    from input $w_t(R_t\backslash O_t)$. {\bf(d)}: Final shape and
    radiance $(a_{t+1}, R_{t+1})$ in frame $t+1$ (adding
    dis-occlusion $D_{t+1}$ to $w_t(R_t\backslash O_t)$). Shaded gray
    regions indicates not defined.}
  \label{fig:overview}
\end{figure*}

\section{Energy Formulation}
\label{sec:energies}

This section concerns formulation of a joint energy for the warp of a
given template to a subset to be determined in an image and the
occluded subset of the template in the first sub-section, and in the
subsequent sub-section, a formulation of an energy for the
dis-occlusion.

\subsection{Joint Energy for the Warp and Occlusion}
\label{sec:energy_occlusion}

We model the warp $w_t$ as a diffeomorphism (smooth invertible
non-rigid transformation) from $R_t\backslash O_t$ (the co-visible
region) to an unknown target set (that must be determined) in the
domain of $I_{t+1}$.  An occlusion of region $R_t$ is the subset of
$R_t$ that goes out of view in frame $t+1$. We compute occlusions as
the subset of $R_t$ that \emph{does not register} to $I_{t+1}$ under a
viable warp. Thus, the occlusion depends on the warp, but to determine
an accurate warp, data from the occluded region must be excluded,
hence a circular problem. Therefore, occlusion detection and
registration should be computed jointly.

We avoid subscripts $t$ for ease of notation in the rest of this
section, and all sections until Section~\ref{sec:filtering}.  We
formulate the problem of given a region $R \subset \Omega$, the
radiance $a : R \to \R^k$, and $I : \Omega \to\R^k$ to compute the
occluded part $O$ of $R$, the warp $w$ defined on $R\backslash O$, and
$w(R\backslash O)$ such that $I(x) = a(w^{-1}(x)) + \eta(x)$ for $x\in
w(R\backslash O)$ (where $\eta$ is noise modeled in
\eqref{eq:appearance_model}).

The warp $w$ is a diffeomorphism in the un-occluded region $R\backslash
O$. For ease in the optimization, we consider $w$ to be extended to a
diffeomorphism on all of $R$; the warp of interest will be the
restriction to $R\backslash O$. We setup an optimization problem to
determine $w$ so that $w(R\backslash O)$ is the object region in $I$,
i.e., $a|R\backslash O$ should correspond to $I|w(R\backslash O)$ via
the warp $w$. We thus formulate the energy (to be minimized in $O,w$)
as
\begin{align} \label{eq:energy_o}
  E_o(O,w; I, a, R) &= \int_{R} f(w(x),x) \ud x
  + \beta_o\mbox{Area}(O) \\
  f(y,z) &= \rho( (I(y)-a(z))^2 ) \bar \chi_{O}(y)
\end{align}
where $\beta_o>0$ is a weight, $\bar \chi_O(x) = 1 - \chi_O(x)$,
$\chi_O$ is the characteristic function of $O$, and $\rho : \R\to \R$
is some monotonic function (e.g., $\rho(x)=x$ for a quadratic penalty
or $\rho(x)=\sqrt{ x + \varepsilon }$ where $\varepsilon > 0$ for a
robust penalty \cite{black1996robust}; the choice of $\rho$ will
depend on the actual noise model $\eta$ chosen in
\eqref{eq:appearance_model}). The first term penalizes deviation of
the object radiance, $a$, to the pull-back of the image intensity
$I|w(R)$ under $w$ onto the region $R$. The term $\bar \chi_O(x)$
implies that $w$ is only required to warp the radiance to match the
image intensity $I$ in the \emph{un-occluded region} $R\backslash
O$. The occlusion area penalty is needed to avoid the trivial solution
$O = R$. Given a moderate frame rate of the camera, it is realistic to
assume that the occlusion is small in area compared to the object.

Due to the aperture problem, multiple warps $w$ can optimize the
energy $E_o$, and typically a regularization term is added directly
into the energy (e.g., for small warps as in optical flow
\cite{horn1981determining}, or for large warps
\cite{beg2005computing}), changing the energy. Rather than
regularizing the energy, we regularize the \emph{flow optimizing
  $E_o$} in a way that optimizes $E_o$ without changing it, leading to
a favorable solution; this is described in
Section~\ref{sec:sobolev_optimization}.

\subsection{Energy Formulation of Dis-Occlusion}
\label{sec:energy_disocclusion}

We now describe the energy formulation of the dis-occlusion
$D_{t+1}\subset \Omega$ of the object at frame $t+1$ given the warped
un-occluded part of the region $w_{t}(R_{t}\backslash O_{t})$
determined from the optimization of the energy in the previous
section, and the image $I_{t+1}$.  To determine the disoccluded region
of the object (the region of the projected object that comes into view
in the next frame that is not seen in the current template), it is
necessary to make a prior assumption on the 3D object.

A realistic assumption is self-similarity of the 3D object's radiance
(that is, the radiance of the 3D object in a patch is similar to other
patches). To translate this prior into determining the dis-occlusion
of the object $D_{t+1}$, we assume that the image in the disoccluded
region of the object is similar to parts of the image $I_{t+1}$ in
$w_{t}(R_{t}\backslash O_{t})$, and for computationally efficiency, we
assume similarity to close-by parts of the template. This is true in
many cases, and is effective as shown in the experiments.

Although dis-occlusions in image $I_{t+1}$ are parts of the image that
do not correspond to $I_{t}$ (i.e., an occlusion backward in time),
these parts may be a dis-occlusion of the object or the
\emph{background}. It is not possible to determine without additional
priors which dis-occlusions are of the object of interest. Our method
works directly from the prior without having to compute a backward
warp.

We now setup an optimization problem for the dis-occlusion. To simplify
notation, we avoid subscripts in $D_{t+1}$ and $I_{t+1}$, and denote
$R'=w_{t}(R_{t}\backslash O_{t})$. The energy is
\begin{equation} \label{eq:energy_d}
  E_d(D) = -\int_{D} p(x) \ud x + \beta_d \mbox{Area}(D)
\end{equation}
where $D \subset \Omega \backslash R'$, $p(x)\geq 0$ denotes the
likelihood that $x\in \Omega \backslash R'$ belongs to the
dis-occluded region, and $\beta_d>0$ is a weight. The dis-occluded
region, assuming a moderate camera frame rate, is small in area
compared to the projected object, hence the penalty on area.

\begin{figure}
  \centering
  \includegraphics[clip,trim=10 205 10 250, totalheight=0.9in]{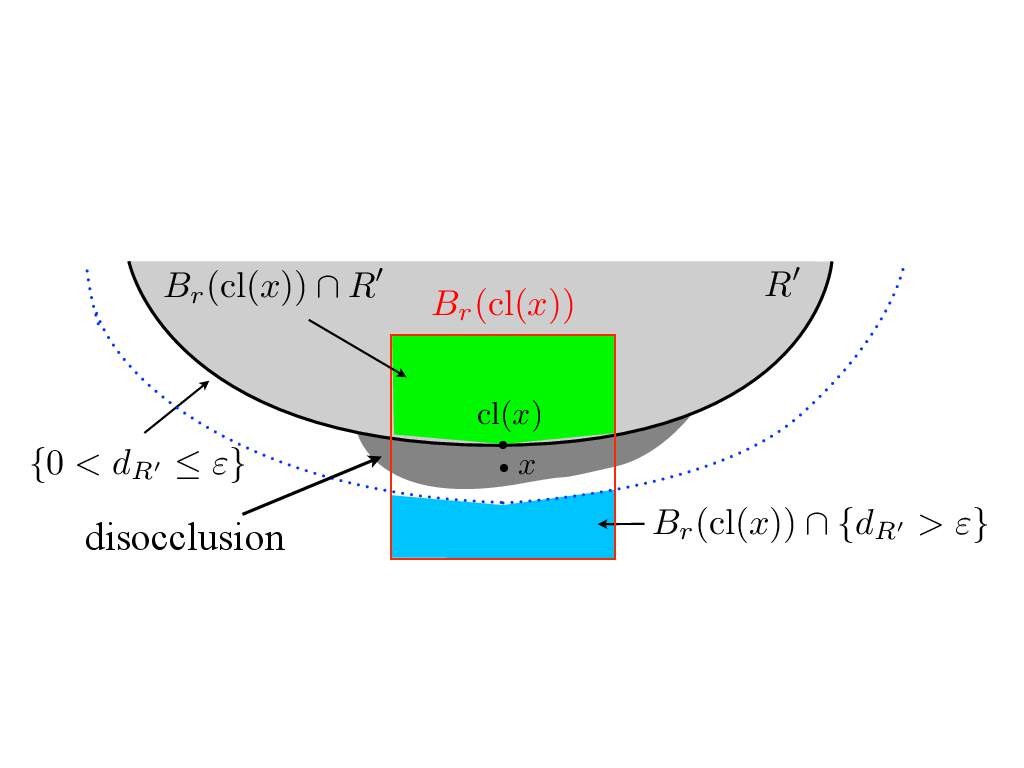}
  \caption{{\bf Diagram of quantities used in the likelihood
      $p(x)$ of a disoccluded pixel.} The dark gray region is the
    dis-occlusion to be determined. Light gray region
    is $R'$, region before the dis-occlusion is determined. A pixel $x$
    within the band $\{ 0 < d_{R'} \leq \varepsilon \}$ is depicted,
    and its closest pixel to $R'$, $\mbox{cl}(x)$. The green (blue) region is
    where the foreground (background) distribution
    $p_{\text{cl},f}(x)$ ($p_{\text{cl},b}(x)$) is determined.}
  \label{fig:schematic_disocclusion_energy}
\end{figure}

Let $\text{cl}(x)$ denote the closest point of $R'$ to $x$, and let
$B_r(x)$ denote the ball of radius $r$ about the point $x$.  We choose
$p(x)$ to have two components (see diagram in
Fig.~\ref{fig:schematic_disocclusion_energy}.): one that measures the
fit of $I(x)$ to the local distribution of $I$ within
$B_r(\text{cl}(x)) \cap R'$ versus the background $B_r(\text{cl}(x))
\cap \{ d_{R'} > \varepsilon \}$ in $I$, and the second that measures
nearness of $x$ to $R'$.  One choice of $p$ is
\begin{equation} \label{eq:p_disc}
  p(x) \propto \exp{\left[-\frac{d_{R'}(x)^2}{2\sigma_d^2} + 
      p_{\text{cl}(x),f}(I(x)) - p_{\text{cl}(x),b}(I(x))  \right] } 
\end{equation}
where $d_{R'}(x)$ indicates the Euclidean distance from $x$ to $R'$,
$\sigma_d>0$ is a weighting factor, $p_{\text{cl}(x),f}, \,
p_{\text{cl}(x),b}$ are Parzen estimates of the intensity distribution
of $I$ in $B_r(\text{cl}(x)) \cap R'$ (resp.~$B_r(\text{cl}(x)) \cap
\{ d_{R'} > \varepsilon \}$) where $\varepsilon$ is chosen large
enough so that the region includes some background beyond the
dis-occlusion.

\section{Coarse-To-Fine Optimization of Energies Defined on Warps}
\label{sec:sobolev_optimization}

In order to optimize $E_o$, we will apply an alternating scheme,
alternating between optimization of $O$ and $w$, which will be
presented in Section~\ref{sec:final_algorithm}. This section will
focus on the general problem of optimizing an energy defined on warps
of the form 
\begin{equation}
  E(w) = \int_{R} f(w(x),x) \ud x
\end{equation}
where $f : \Omega\times \Omega \to \R$. Note that this sub-problem is
relevant in optimizing $E_o$. The optimization with
respect to $w$ is done using a steepest descent scheme. Steepest
descent depends on a Riemannian metric on the space of warps, $w$. The
Riemannian metric is defined on infinitesimal perturbations of the
warp $w$, and the metric controls the type of motions/deformations
that are favored in optimizing the energy. We will design a novel
Sobolev-type metric, and use it in the steepest descent of $E$. 

The motivation for the design of this metric comes from the active
contours literature
\cite{sundaramoorthi2007sobolev,sundaramoorthi2008coarse}, where it
was shown that Sobolev-type metrics defined on \emph{curves}
(boundaries of regions) result in flows that optimize the energy in a
coarse-to-fine manner, initially optimizing the energy with respect to
coarse perturbations, and then moving on to finer perturbations when
coarse deformations no-longer optimize the energy. This coarse-to-fine
behavior is done in an automatic fashion simply by using the Sobolev
metric to compute the gradient of the energy.  Motivated by this
coarse-to-fine property, we design a new-Sobolev metric that is suited
for energies defined on warps (rather than on curves in the active
contour literature), that is, a \emph{region-based metric}. The metric
used in \cite{sundaramoorthi2007sobolev,sundaramoorthi2008coarse} does
not apply to the energy of interest in this paper as $E$ is defined on
the space of warps (the point-wise correspondence of the interior is
essential) rather than on boundaries of closed curves as the energies
considered in
\cite{sundaramoorthi2007sobolev,sundaramoorthi2008coarse}.

\subsection{Sobolev Region-Based Metric and Gradient}

We start by presenting some theoretical background so that the metric
can be defined and the computation of the gradient of the energy with respect to the
metric can be done. The space where our energy is defined is
\begin{equation}
  M =  \{ w: R\to \Omega \subset \R^2 \, : \,  w : R \to w(R) \mbox{ is a diffeomorphism} \},
\end{equation}
where $R\subset \Omega \subset \R^2$ is a compact set with smooth
boundary (and thus also the range of $w$'s are compact and have smooth
boundary).  A diffeomorphism is a smooth invertible map whose inverse is
also smooth.  The range of $w\in M$ need not be all of $\Omega$, but
rather an arbitrary subset of $\Omega$.  We refer to $M$ as the
\emph{space of parameterized regions} since elements $w\in M$
parameterize regions $w(R)$ via the fixed region $R$. Note that the
actual parameterization of a region is important as the energy of
interest $E$ depends on the parameterization.

Infinitesimal perturbations of $w$ are given by smooth vector fields
$h: R \to \R^2$, which form the tangent space to $w$ and is denoted
$T_w M$. An infinitesimal perturbation of $w$ is $w_{\varepsilon}$,
given by
\begin{equation}
  w_{\varepsilon}(x) =w(x) + \varepsilon h(x).
\end{equation}
Note that if $\varepsilon > 0$ is small enough, then
$w_{\varepsilon}\in M$, i.e., $w_{\varepsilon}$ is a diffeomorphism,
which implies that $M$ is a manifold and thus, we may define a
Riemannian metric on $T_w M$, which in turn allows us to define
gradients of the energy. Perturbations $h$ are defined on $R$, and by
right translation, i.e., $h\circ w^{-1}: w(R) \to \R^2$, they are also
defined on $w(R)$. We now specify an inner product on $T_w M$, which
makes $M$ a Riemannian manifold:
\begin{defn}[Sobolev-type Inner Product on $M$]
  The inner product on the set of perturbations of
  $w$ (i.e., the metric) that we consider is defined as follows:
  \begin{equation} \label{eq:sobolev_inner}
    \ip{h_1}{h_2}{w} = \mean{\hat h_1} \cdot \mean{\hat h_2} + 
    \alpha \int_{w(R)} \mbox{tr}\left\{
      \nabla \hat h_1(x)^T\nabla \hat h_2(x) \right\} \ud x
  \end{equation}
  where $\alpha>0$, $\hat h := h\circ w^{-1}$ when $h : R\to \R^2$,
  $\nabla \hat h_1(x)$ denotes the spatial Jacobian of $\hat h_1(x)$,
  $\mbox{tr}$ denotes the trace of a matrix, $\ud x$ is the area
  measure on $w(R)$, and
  \begin{equation}
    \mean{\hat h} = \frac{1}{|w(R)|}\int_{w(R)} \hat h(x) \ud x.
  \end{equation}
\end{defn}
\noindent The first term in \eqref{eq:sobolev_inner} uses the mean value of the
perturbations rather than the $\mathbb{L}^2$ inner product of the
perturbations as in standard Sobolev inner products
\cite{evans1998partial}. This change is for convenience in the
algorithm that we present to optimize $E$, and an easy decomposition
of the gradient into orthogonal components as we shall see. The second
term of \eqref{eq:sobolev_inner} is the $\mathbb L^2$ inner product of
the Jacobian of the perturbations.

The goal now is to define a gradient (or steepest) descent approach to
minimize $E$. It should be noted that the gradient of an energy
depends on the choice of inner product on the space of perturbations
of the warp. The typical choice (either implicitly or explicitly) is
the $\mathbb L^2$ inner product, but this does not have desirable
properties for tracking applications. We therefore, compute the
gradient with respect to the Sobolev inner product defined above
\eqref{eq:sobolev_inner}. First, we state the definition of the
gradient, which shows the dependence on the inner product:
\begin{defn}[Gradient of Energy]
  Let $E : M \to \R$, $w\in M$, $h\in T_w M$, and $\ip{}{}{w}$ denote
  the inner product on $T_w M$. The {\bf directional derivative} of
  $E$ at $w$ in the direction $h$ denoted, $\ud E(w)\cdot h$, is 
  \begin{equation}
    \ud E(w)\cdot h = \der{}{\varepsilon} E(w+\varepsilon h)|_{\varepsilon = 0}.
  \end{equation}
  The gradient of $E$, denoted $\nabla_w E\in T_w M$, is the
  perturbation that satisfies the relation
  \begin{equation}
    \ud E(w)\cdot h = \ip{\nabla_w E}{h}{w}
  \end{equation}
  for all $h\in T_w M$.
\end{defn}
In order to see intuitively how the choice of inner product affects the
gradient, we give another interpretation of the gradient, i.e., it
is a perturbation that maximizes the following ratio:
\begin{equation}
  \frac{\ud E(w) \cdot h }{\|h\|_w}
\end{equation}
where $\|h\|_w = \sqrt{\ip{h}{h}{w}}$ is the norm induced by the inner
product. That is, the gradient is a perturbation $h$ that maximizes
the change in energy by perturbing in direction $h$ divided by the
norm of the perturbation. Therefore, while it is often stated that the
gradient is the direction that maximizes the energy the fastest, it is
actually the direction that maximizes energy while \emph{minimizing
  its cost} (measured by the norm). Since non-smooth perturbations
cost a lot according to the Sobolev norm, they are not typically
Sobolev gradients. Coarse perturbations are favored for Sobolev
gradients when they can increase the energy. Note that moving in the
negative gradient direction, $h=-\nabla_w E$, reduces the energy for
any choice of $\alpha$.

The Sobolev gradient of $E$, $G=\nabla_w E$ is a linear combination of
two (orthogonal) components, the translation and the deformation:
\begin{equation}
  G(x) = \mean{G} + \frac{1}{\alpha} \tilde G(x), \, x\in w(R)
\end{equation}
where $\tilde G$ (which is independent of $\alpha$) satisfies the
following Poisson PDE:
\begin{equation}\label{eq:deformation_SG_main}
  \begin{cases}
    -\Delta \tilde G(x) = f_1(x,w^{-1}(x)) \det{(\nabla w^{-1}(x))}^{-1} - \mean{ f_1(\cdot,w^{-1}(\cdot) \det{(\nabla w^{-1}(\cdot))})^{-1}
    } & x\in w(R) \\
    \nabla \tilde G(x) \cdot N = 0  & x\in \partial w(R) \\
    \mean{\tilde G}=0 &
  \end{cases},
\end{equation}
where $\Delta$ denotes the Laplacian, $N$ is the outward unit normal
to $\partial w(R)$,
\begin{equation}
  \mean{G} = \int_{R} f_1(x,w^{-1}(x)) \det{(\nabla w^{-1}(x))}^{-1} \ud x,
\end{equation}
and $f_1$ denotes the partial derivative of $f$ with respect to the
first argument of $f$. Details of the derivations for these
expressions can be found in Appendix~\ref{app:sobolev_gradients}.  The
numerical scheme to solve \eqref{eq:deformation_SG_main} is given in
Appendix~\ref{app:numerics_Poisson}. Note that larger $\alpha$
(implying more spatial regularity) implies the gradient approaches a
translation (the smoothest transformation), and smaller $\alpha$
implies a non-rigid deformation, which is spatially smooth and the
amount of smoothness depends on the data.

\subsection{Optimizing the Energy via Gradient Descent}
\label{subsec:gradient_descent_warp}

The gradient flow to optimize $E_o$ is then given by the following
partial differential equation 
\begin{equation} \label{eq:warp_evolution}
  \begin{cases}
    \partial_{\tau} \phi_{\tau}(x) = -\nabla_w E(\phi_{\tau}(x)) &  x\in R \\
    \phi_{0}(x) = x & x\in R
  \end{cases}
\end{equation}
where $\tau$ indicates an artificial time parameter parameterizing the
evolution of the warp $\phi_{\tau} : R \to \Omega$ at a given frame in
the image sequence (not to be confused with the frame number $t$). The
final converged $\phi_{\tau}$ is a local optimizer of the energy $E$.
It should be noted that the above equation maintains that $\phi_{\tau}
\in M$, i.e., that the final converged result is a
diffeomorphism. This can be seen since $\nabla_w E$ is smooth (it is
the solution of a Poisson equation and thus, $H^2$
\cite{evans1998partial}), and integrating a smooth vector field
results in diffeomorphism using classical results \cite{ebin1970groups}
(and in particular \cite{mennucci2008properties} for first order
Sobolev regularity), precise details for this fact are out of the
scope of this paper.

In implementing the gradient flow \eqref{eq:warp_evolution}, we are
interested in the final converged region, and thus we keep track of
$R_{\tau} = \phi_{\tau}(R)$. For numerical ease and accuracy, we keep
track of $R_{\tau}$ using a level set method \cite{osher1988fronts},
although it is not required. We also keep track of the backward map
$\phi_{\tau}^{-1}$, which is needed to evaluate the gradient $\nabla_w
E(\phi_{\tau}(x))$.

The level set function will be denoted $\Psi_{\tau} : \Omega \to
\R$. Its evolution is described by a transport PDE. The backward map
$\phi_{\tau}^{-1}$ also satisfies a transport equation. Therefore, the
optimization of $E$ is given by the coupled PDE:
\begin{align}
  \label{eq:Psi_init}
  \Psi_0(x) &= d_R(x), \, x\in B_2(R) \\
  \phi_{0}^{-1}(x) &= x, \, x\in R \\
  \label{eq:R_init}
  R_0 &= R \\
  \label{eq:gradient_compute}
  G_{\tau} &= \nabla_w E(\phi_{\tau})\\
  \label{eq:transport_backward_warp}
  \partial_{\tau} \phi_{\tau}^{-1} &= \nabla\phi_{\tau}^{-1}(x)\cdot
  G_{\tau}(x), x\in R_{\tau}\\
  \partial_{\tau} \Psi_{\tau} &= \nabla \Psi_{\tau}(x) \cdot
  G_{\tau}(x), x\in B_2(R_{\tau}) \\
  \label{eq:region_update}  
  R_{\tau} &= \{ \Psi_{\tau} < 0 \}
\end{align}
where $\partial_{\tau}$ denotes partial with respect to $\tau$, and
$B_2(R_{\tau}) = \{ x\in \Omega : |d_{R_{\tau}}(x)| \leq 2\}$ where
$d_{R_{\tau}}$ is the signed distance function of $R_{\tau}$. The
region $R_{\tau}$ is updated in direction of minus the gradient of
$E$, $-G_{\tau} : R_{\tau}\to \R^2$ via the level set evolution. Note
$G_{\tau}$ is extended to $B_2(R_{\tau})$ as in narrowband level set
methods.  The backward warp $\phi_{\tau}^{-1} : R_{\tau} \to R$ is
computed by flowing the identity map along the velocity field
$-G_{\tau}$ up to time $\tau$, and this is accomplished by the
transport equation \eqref{eq:transport_backward_warp}. At convergence
(when $E$ does not decrease), we denote this time $\tau_{\infty}$,
$w=(\phi^{-1}_{\tau_{\infty}})^{-1} : R \to R_{ \tau_{\infty} }$ is a
local optima of $E$, and $R_{ \tau_{\infty} } = w(R)$ is the region
matched in the image $I$.

The evolution above is automatically coarse-to-fine for any choice of
$\alpha$, that is, the gradient descent favors coarse
motions/deformations initially before transitioning to more finer
scale deformations. See Figure~\ref{fig:coarse_to_fine} in
Sub-Section~\ref{subsec:discussion_HS_LK} for an experimental
verification of this property.

\subsection{Parameter Independent Optimization}
One of the advantages of the particular form of the Sobolev-type
metric chosen in \eqref{eq:sobolev_inner} besides the coarse-to-fine
property is that one can eliminate the need for choosing the parameter
$\alpha$, while optimizing $E$. One can take $\alpha\to\infty$, in
which case $G \to \mean{G}$, a translation motion. One can optimize by
translating in the direction $-G=-\mean{G}$ when $\alpha\to\infty$,
until convergence. At convergence, $\mean{G} = 0$, then one can evolve
the warp infinitesimally in the negative gradient $-G = -\tilde
G/\alpha$ direction for any finite $\alpha$. Since the gradient
depends only on $\alpha$ by a scale factor, the choice of $\alpha$ is
just a time re-parameterization of the evolution, not changing the
geometry of the evolution, and does not impact the final converged
warp nor the converged region. The algorithm to optimize $E$ that is
not dependent on the choice of $\alpha$ is summarized in the following
steps:
\begin{enumerate}
\item Perform the initializations \eqref{eq:Psi_init}-\eqref{eq:R_init}.
\item Repeat the evolution
  \eqref{eq:gradient_compute}-\eqref{eq:region_update} with
  $\alpha\to\infty$, in which case $G_{\tau} = \mean{G_{\tau}}$, until
  convergence (when $\mean{G_{\tau}}=0$).
\item Perform one time step
  \eqref{eq:gradient_compute}-\eqref{eq:region_update} with the
  deformation of $G_{\tau} \propto \tilde G_{\tau}$ (one may choose
  $\alpha=1$, but any choice would give the same result).
\item Repeat Steps 2-3 until convergence (when $E$ does not decrease).
\end{enumerate}

The procedure above optimizes with respect to translations first until
convergence, then optimizes with respect to deformations that are not
translations (favoring coarse motions/deformations if they optimize
the energy), and the process is iterated. This results in a scheme
that is independent of a regularity parameter $\alpha$, and that
favors a coarse-to-fine evolution (like the gradient descent with any
fixed $\alpha$) of the region $R_{\tau}$ and coarse-to-fine
motion/deformation estimation.

\subsection{Discussion}
\label{subsec:discussion_HS_LK}

We now discuss the relation of our approach to classical optical flow
and tracking approaches, namely the approach by Lucas and Kanade
\cite{lucas1981iterative} and Horn and Schunck
\cite{horn1981determining}. 

Since there are multiple possible solutions optimizing $E$ (that
contains just data fidelity), regularization must be used to determine
a viable solution. The approach in \cite{lucas1981iterative} is to
restrict the possible warps to a smaller set rather than the space of
diffeomorphisms, i.e., translations, affine motions, or other
parametric groups. While providing less ambiguity in determining a
unique optimizer of $E$, this restricts the possible warps $w$ and
thus also the shape of the region. One may consider optimizing $E$ with
respect to translations first, thus getting a coarse estimate of the
desired region in image $I$, then resort to optimizing in more fine
transformations, e.g., Euclidean transformations (i.e., translations
and rotations), then affine transformations. However, one may go up to
the projective group, and then it becomes unclear what group to choose
to optimize further. The algorithm that we have presented to optimize
$E$ optimizes the energy by using coarse perturbations initially, it
then transitions \emph{continuously} and \emph{automatically} to more
finer-scale perturbations, in fact, it transitions through \emph{all}
possible scales of motions/deformations, eliminating the need to
choose groups of motions to optimize. This property of Sobolev-type
metrics for contours was shown analytically in particular cases using
a Fourier analysis in \cite{sundaramoorthi2008coarse}. In this work,
since we work with regions, the property is harder to show
analytically since the Fourier basis would need to be derived using
the eigenfunctions of the Laplacian defined on a region, difficult to
perform analysis analytically. We therefore demonstrate the property
in an experiment.

The method of optical flow computation in \cite{horn1981determining}
deals with multiple possible optimizers of $E$ by \emph{changing} the
original energy by adding regularization of the warp directly into the
energy; indeed the energy for infinitesimal warps is 
\begin{equation} \label{eq:energy_HS}
  E_{HS}(v;a,I,R) = 
  \int_R |I(x) - a(x) + \nabla a(x) \cdot v(x)|^2\ud x +
  \gamma \int_{R} |\nabla v(x)|^2 \ud x.
\end{equation}
An advantage of this approach over \cite{lucas1981iterative} is that,
the motions/deformations are not restricted to finitely parameterized
motions. The parameter $\gamma$ controls the scale of the estimated
motion (large $\gamma$ implying coarse motion, and small $\gamma$
implying finer motion). One can deform the region $R$ by $v$
infinitesimally to obtain $R_{\tau}$, then recalculate $v$ based on
the warped appearance $a\circ \phi_{\tau}^{-1}$, and iterate the
process to determine the region of the object in the next frame. While
the procedure allows the cumulative warp $w$ to be an arbitrary
diffeomorphism and therefore obtain arbitrarily shaped regions, the
technique relies on the choice of the parameter $\gamma$: large
$\gamma$ yields only coarse approximations of the region shape, and
small $\gamma$ yields finer details of shape, but is likely to be
trapped in fine details of the image before reaching the desired
region of interest. There is no principled way to choose $\gamma$, and
no one scale of motions/deformations, that is no one $\gamma$, is
sufficient. Further, the iterative procedure described does not
optimize an energy for the warp $w$ (although each iteration minimizes
an $E_{HS}$ for an infinitesimal warp).

One ad-hoc solution to the dilemma of choosing $\gamma$ is to attempt
a coarse-to-fine scheme by starting with $\gamma$ large until the
region in the procedure discussed converges, reduce $\gamma$ and then
deform the region until convergence, reduce $\gamma$, etc., (which is
the scheme considered in our preliminary conference paper
\cite{YangS2013occlusions}). While the procedure solves issue of the
choice of $\gamma$ and is coarse-to-fine, our proposed algorithm has
three advantages. First, our scheme does not rely on an ad-hoc scheme
to reduce the parameter $\gamma$. Second, our scheme
\emph{automatically} and \emph{continuously} traverses through
\emph{all} possible scales of motions/deformations favoring roughly
coarse-to-fine transition, whereas the ad-hoc scheme only traverses
through a discrete number of scales (chosen by the scheme to reduce
$\gamma$) and the transition is not automatic.  Reducing $\gamma$
monotonically in the ad-hoc scheme may not always be beneficial (e.g.,
when new coarse structure is ``discovered'' from the data during
evolution and larger $\gamma$ is then needed), our new scheme chooses
the appropriate scale of deformation automatically from computation of
the gradient, generally favoring coarse-to-fine. There is no added
complication in our new Sobolev descent: the gradient has similar
structure as the velocity in Horn \& Schunck, both have similar
numerical implementation, and same efficiency.  Our scheme is thus
much more convenient for practical applications. Lastly, our scheme is
minimizing the objective energy $E$, while the ad-hoc scheme
does not necessarily minimize an energy.

\begin{figure}
  \centering
  \includegraphics[clip,trim=0 0 0 0, totalheight=1.5in]{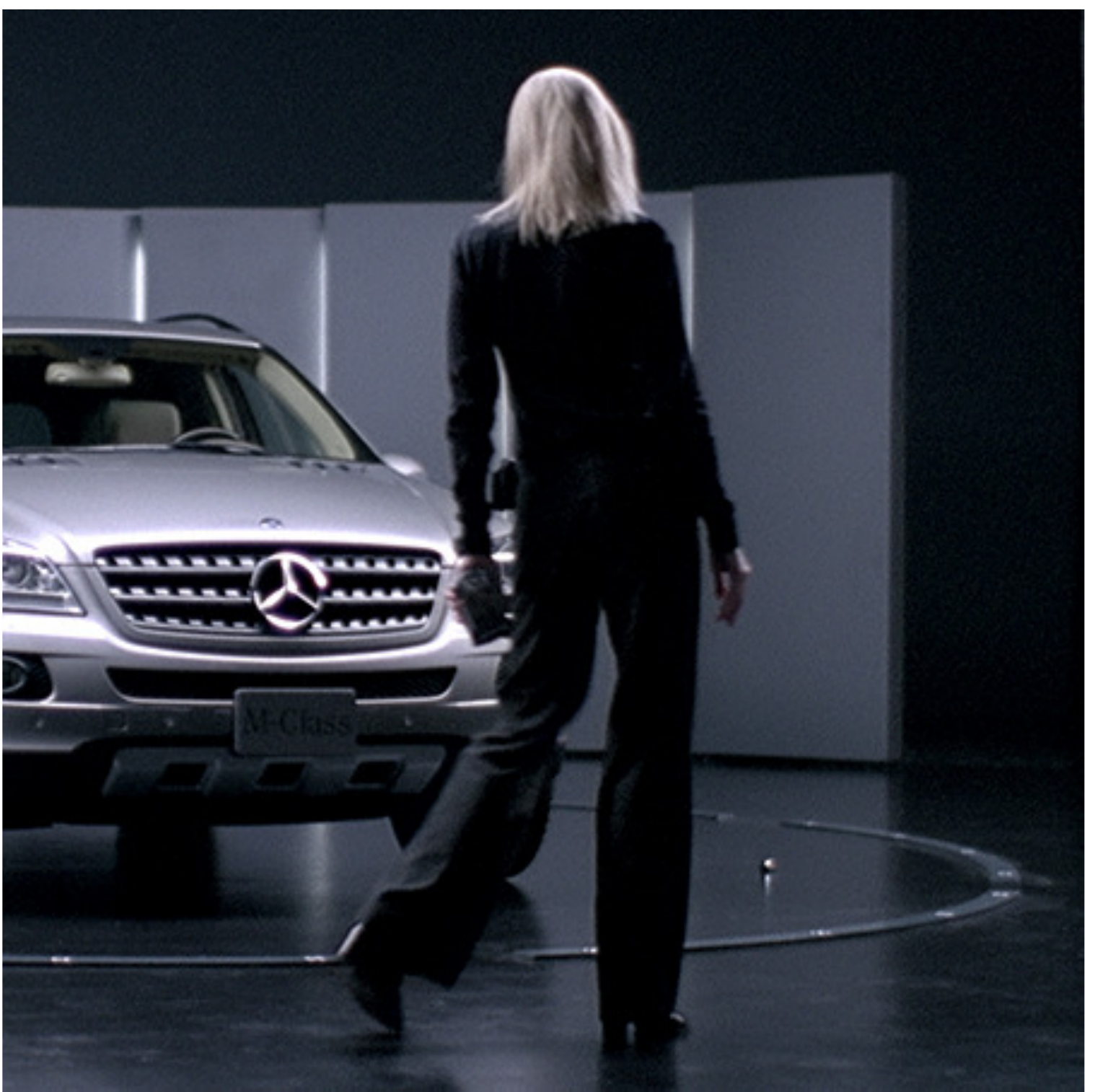}%
  \includegraphics[clip,trim=0 0 0 0, totalheight=1.5in]{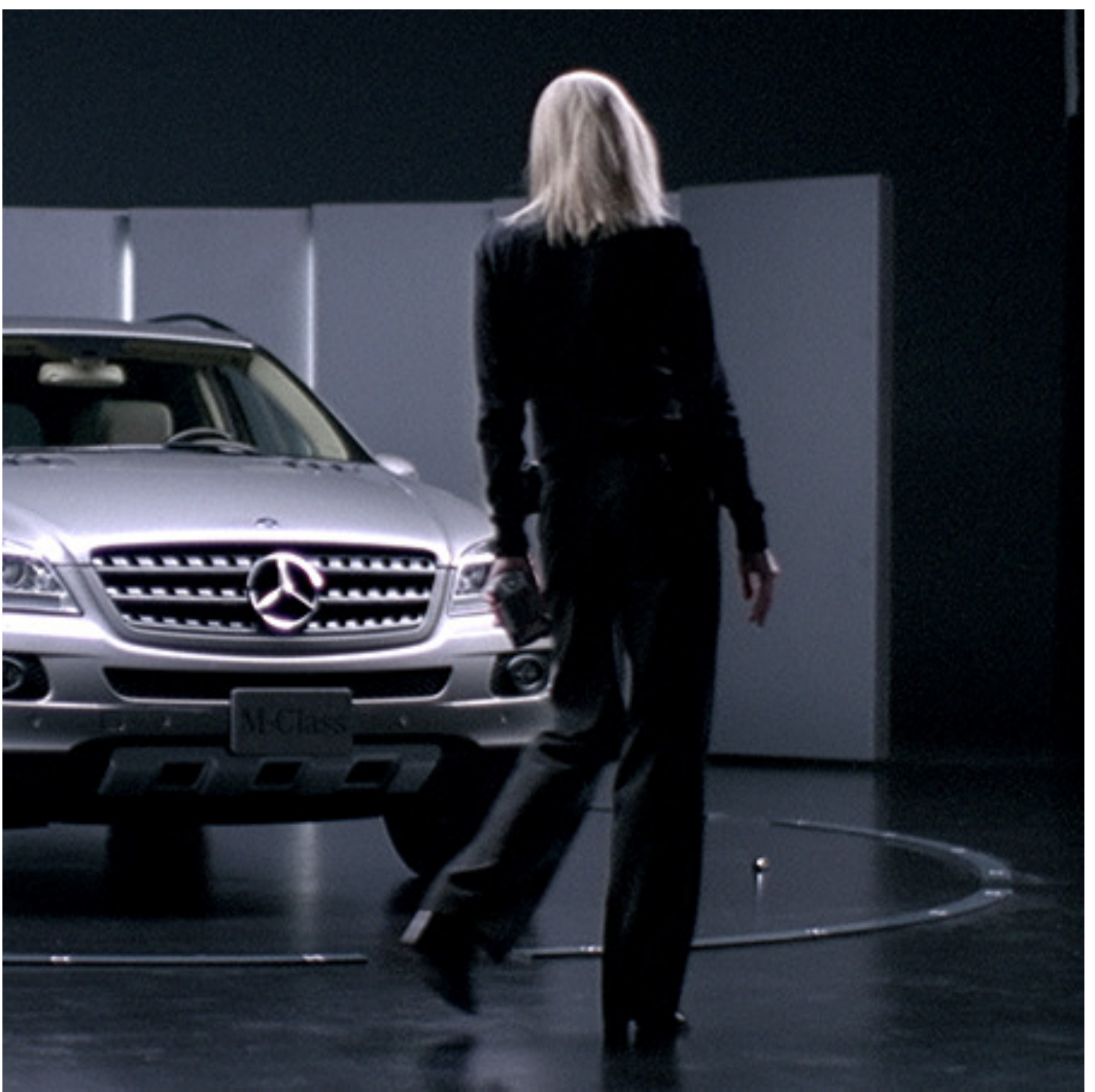}%
  \includegraphics[clip,trim=0 0 0 0, totalheight=1.5in]{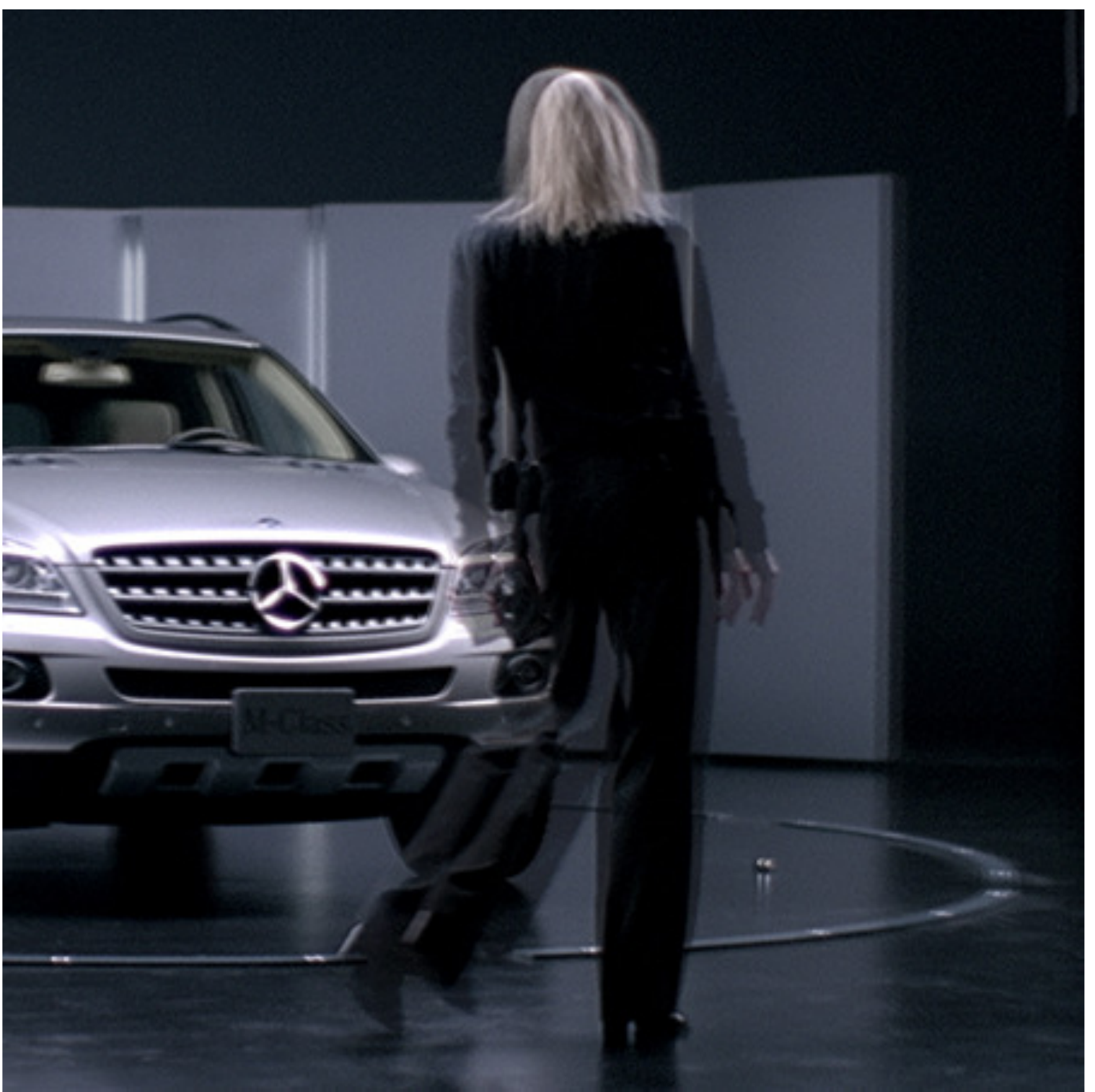}%
  \caption{Images $I_1$ (left) and $I_2$ (middle) used in the experiment in
    Figure~\ref{fig:coarse_to_fine}, and an overlay of $I_1$ on $I_2$
    to show the motion/deformation between frames, which is non-rigid
    and contains both coarse and fine motion/deformations.}
  \label{fig:coarse_to_fine_images}
\end{figure}

We illustrate the coarse-to-fine behavior of the region-based Sobolev
gradient descent by matching a template of the object (woman) obtained
from image 1 to image 2 shown in Fig.~\ref{fig:coarse_to_fine_images},
where there are both coarse-scale and fine-scale deformations. The
evolution (at various snapshots and ran until convergence) of
region-based Sobolev and a Horn and Schunck approach described above
with varying $\gamma$ is shown in Fig.~\ref{fig:coarse_to_fine}. Final
objects detected with these schemes in a zoomed region of interest is
shown in Fig.~\ref{fig:zoom_coarse_to_fine}.  The displacement between
two time instances $\tau_i$ and $\tau_{i+1}$, that is,
$d_{\tau_i,\tau_{i+1}}(x) = \phi_{\tau_i} \circ
\phi_{\tau_{i+1}}^{-1}(x)-x$, is is shown in optical flow code
\cite{baker2011database} (the color indicates direction and darkness
indicates magnitude; magnitude should not be compared across images as
they are re-scaled in each image) in
Fig.~\ref{fig:coarse_to_fine}. Notice that the region-based Sobolev
moves according to coarse motions (nearly constant color in the
visualization) before gradually resorting to finer-scale deformations
whereas the Horn \& Schunck approach has roughly the same scale of
motions/deformation at all stages of the evolution for each
$\gamma$. Small $\gamma$ does not capture regions of coarse
deformation and gets stuck in intermediate structures. Larger $\gamma$
captures regions of coarse deformation, but regions of finer scale
motion (e.g., the legs) are not captured. The Sobolev descent moves
from coarse-to-fine deformations, and thus captures both regions of
coarse and fine deformation.

\def\fPath{figures2/coarse_to_fine/}

\def\fPatha{figures2/coarse_to_fine/alpha500}
\def\fPathb{figures2/coarse_to_fine/alpha10}
\def\fPathc{figures2/coarse_to_fine/alpha100}
\def\fPathd{figures2/coarse_to_fine/alphafree}

\def\fWidth{36pt}
\def\fHeight{67pt}

\def\fWidth{66pt}
\def\fHeight{123pt}
\begin{figure}
  \centering
  \vspace{-0.1in}
  {
    \begin{tabular}{c@{}c@{}c@{}c@{ }c@{}c@{}c}
      \comment{
    \multicolumn{7}{c} { Energy regularization, $\gamma=1$ } \\
    \includegraphics[width=\fWidth,height=\fHeight]{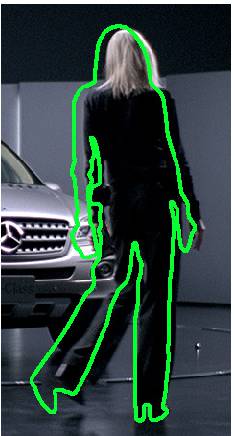} &
    \includegraphics[width=\fWidth,height=\fHeight]{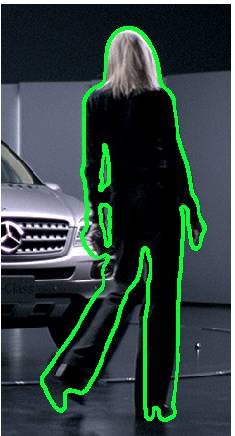} &
    \includegraphics[width=\fWidth,height=\fHeight]{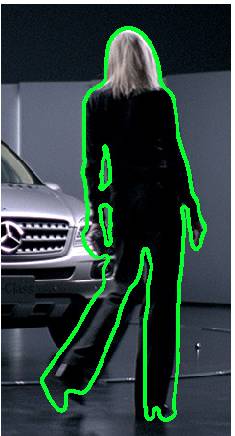} &
    \includegraphics[width=\fWidth,height=\fHeight]{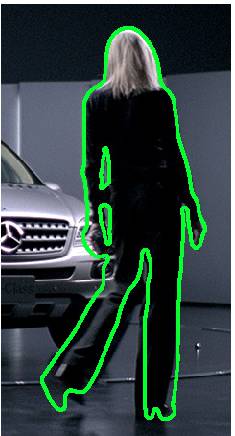} &
    \includegraphics[width=\fWidth,height=\fHeight]{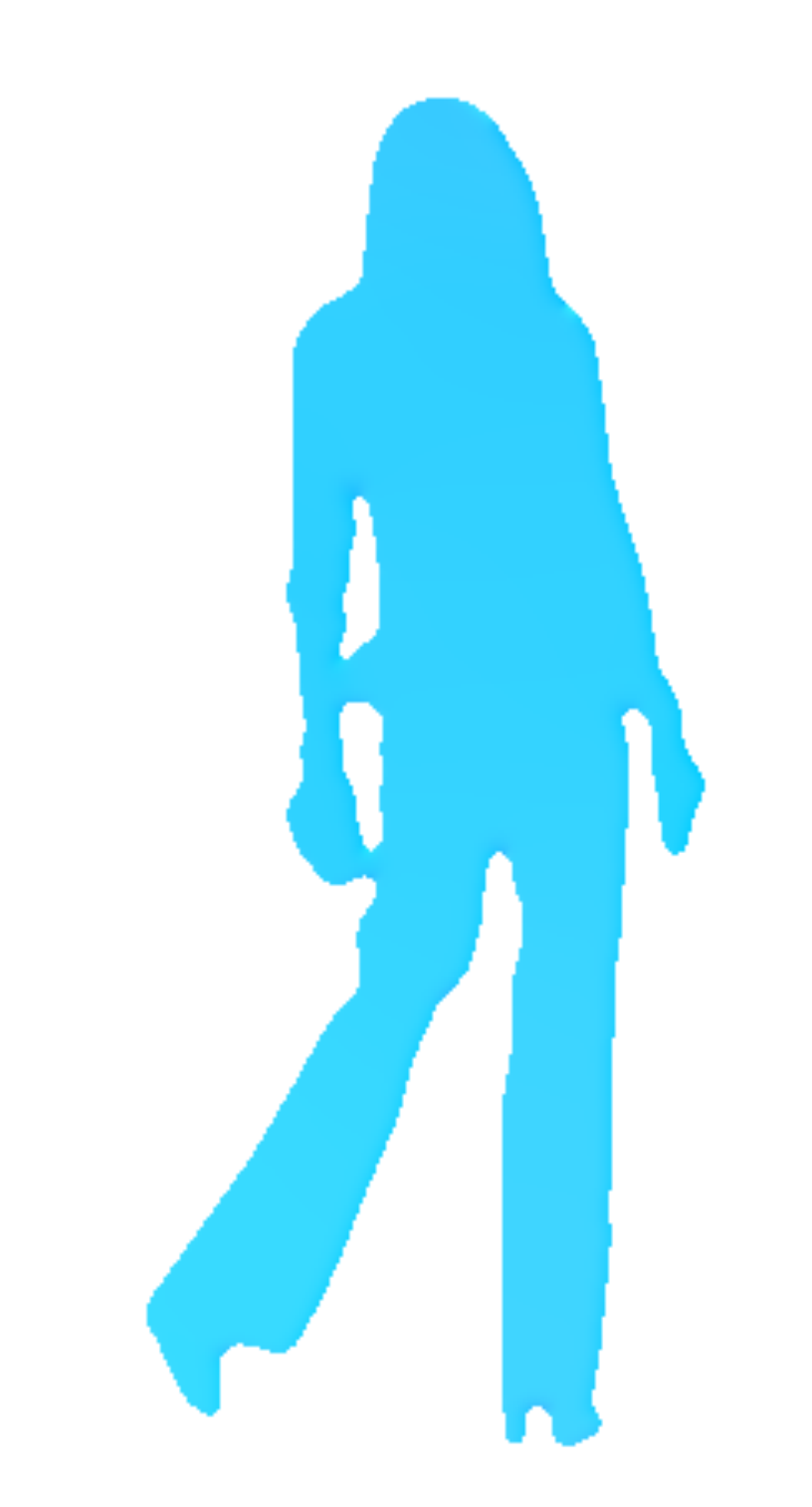} &
    \includegraphics[width=\fWidth,height=\fHeight]{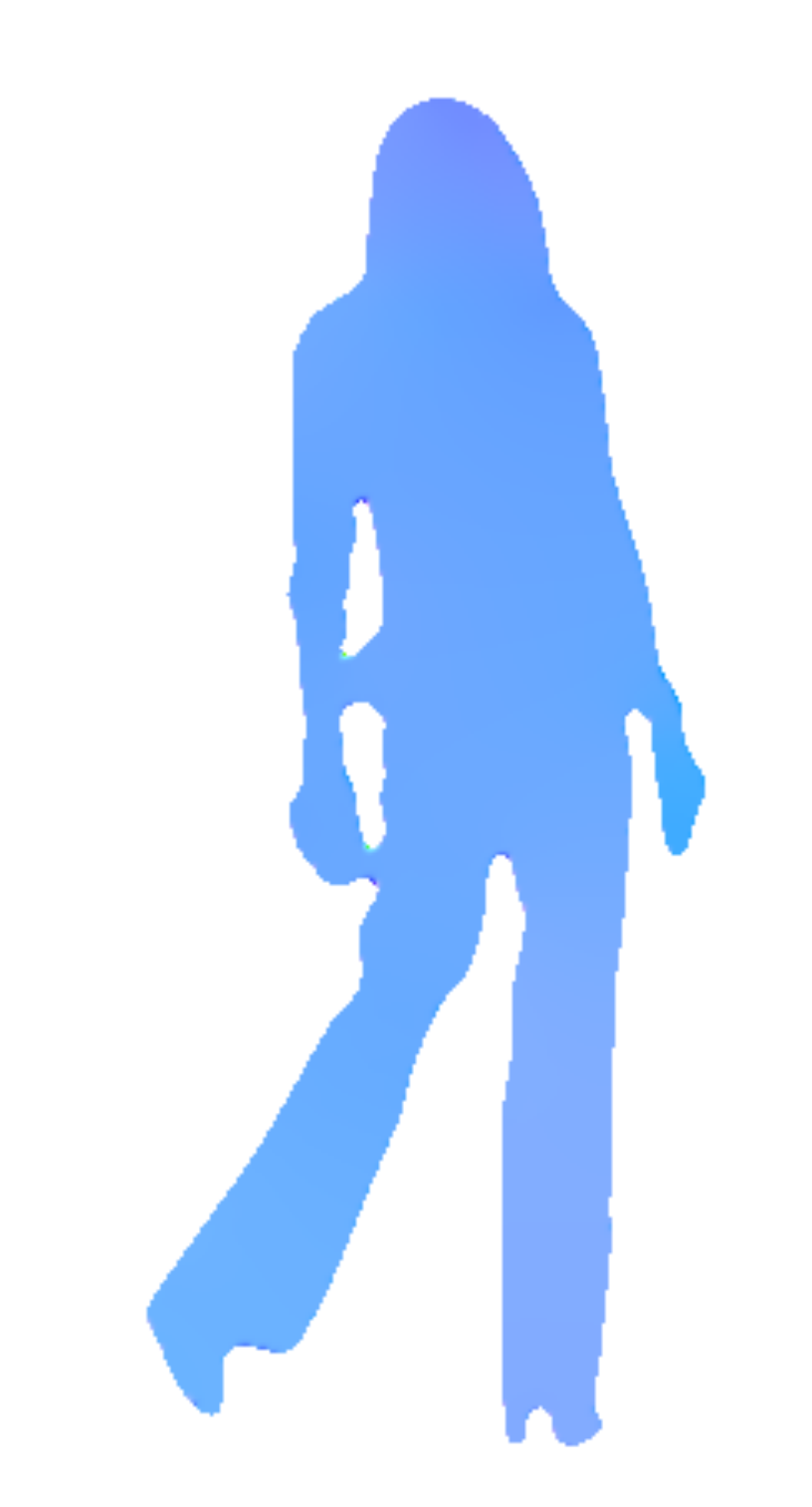} &
    \includegraphics[width=\fWidth,height=\fHeight]{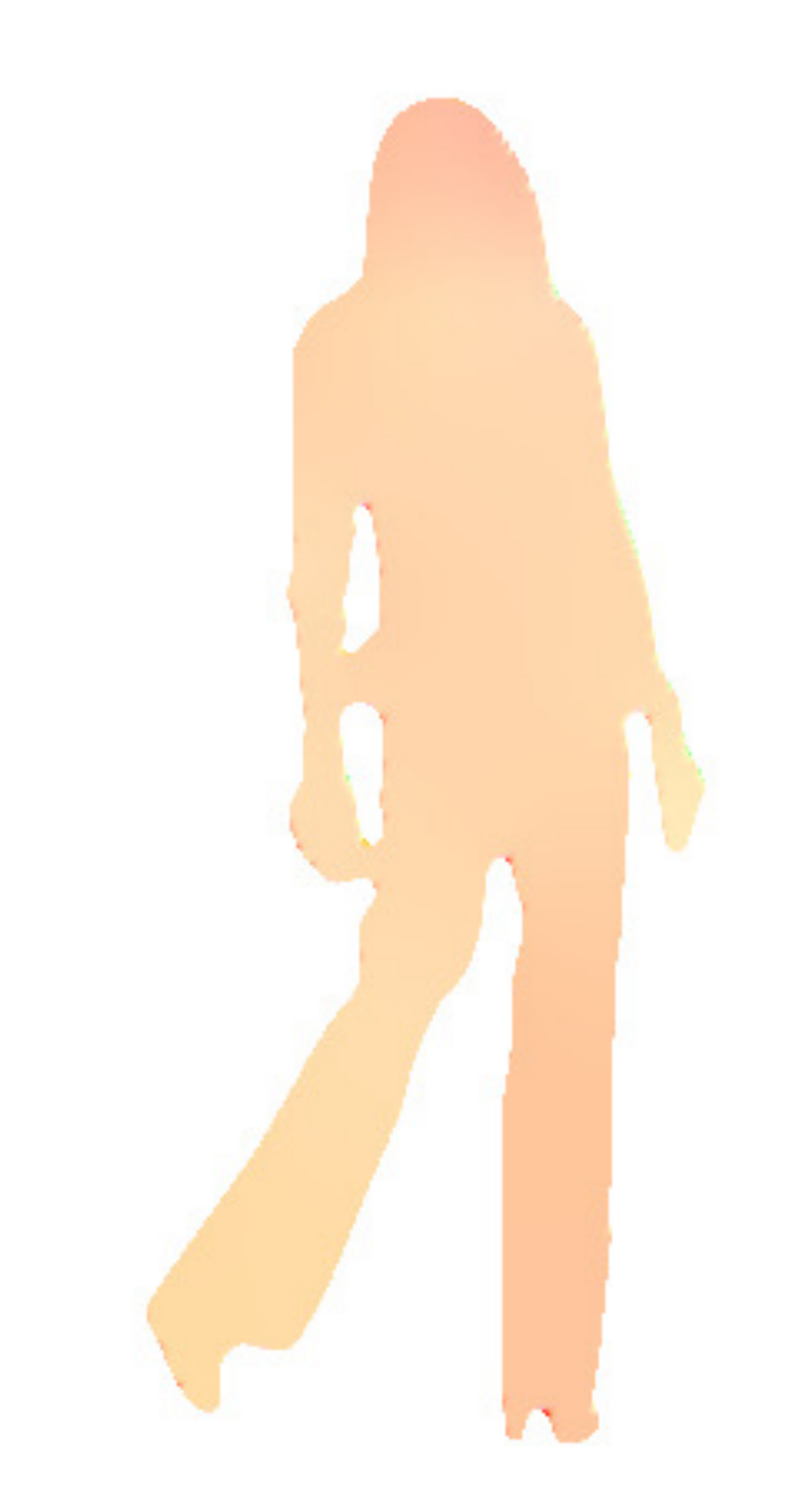}
    \\
  }
    \multicolumn{7}{c} { Energy regularization, $\gamma=10$ } \\
    \includegraphics[width=\fWidth,height=\fHeight]{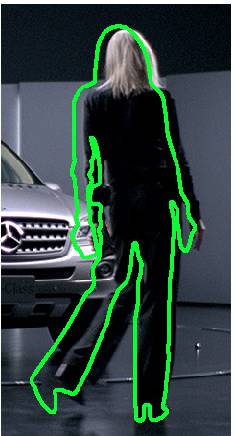} &
    \includegraphics[width=\fWidth,height=\fHeight]{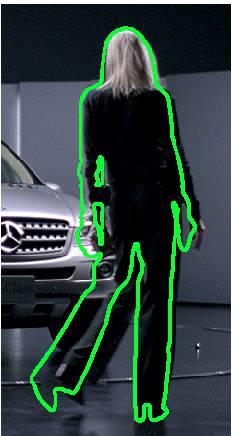} &
    \includegraphics[width=\fWidth,height=\fHeight]{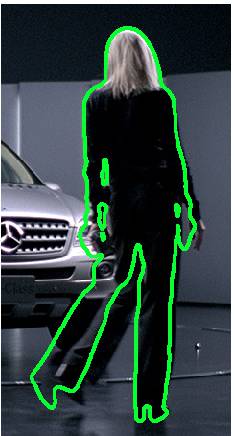} &
    \includegraphics[width=\fWidth,height=\fHeight]{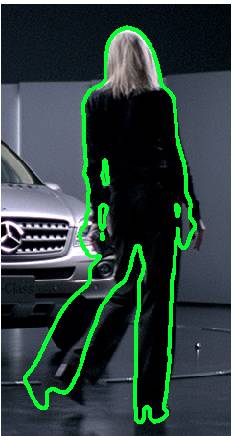} &
    \includegraphics[width=\fWidth,height=\fHeight]{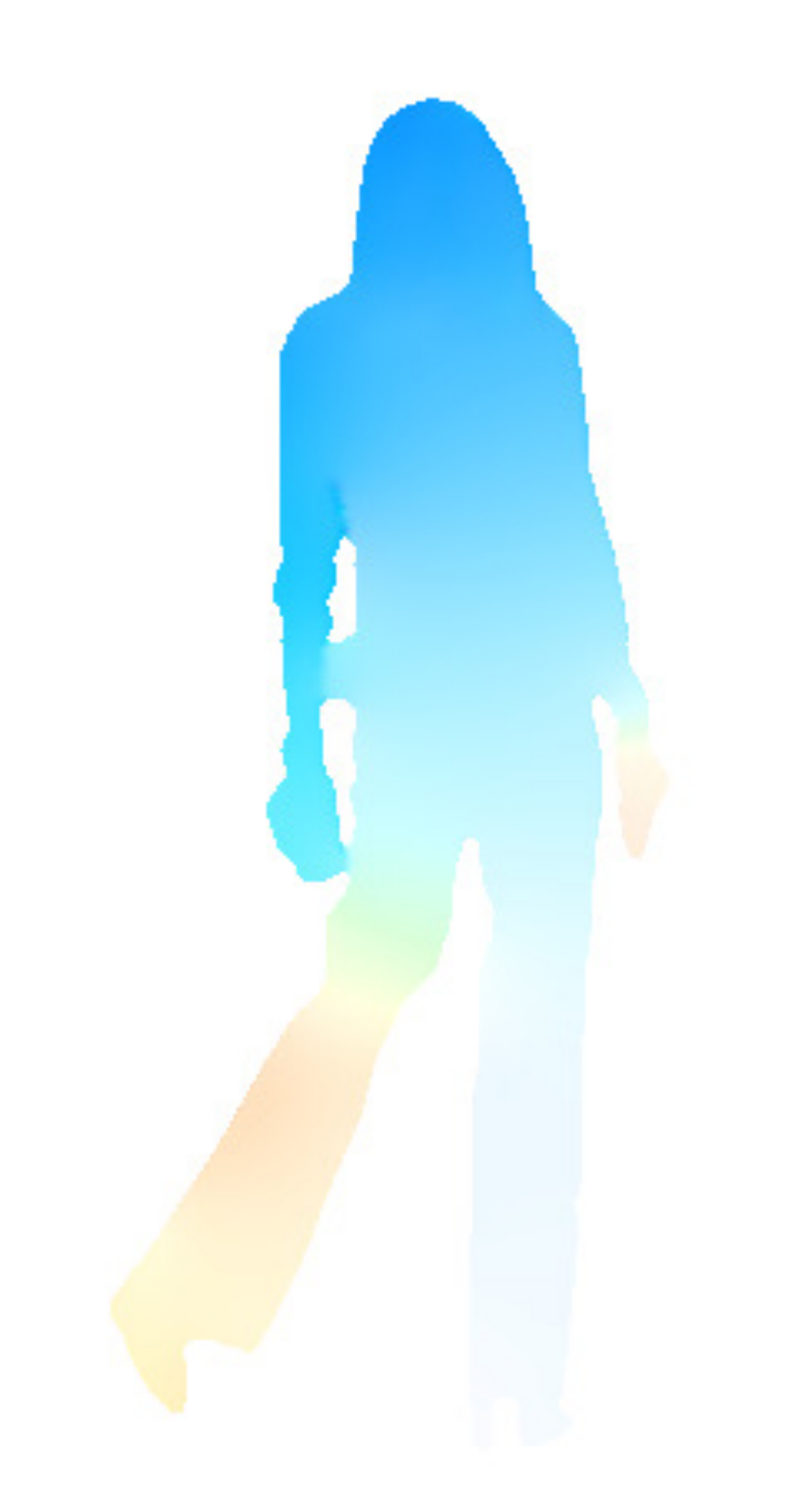} &
    \includegraphics[width=\fWidth,height=\fHeight]{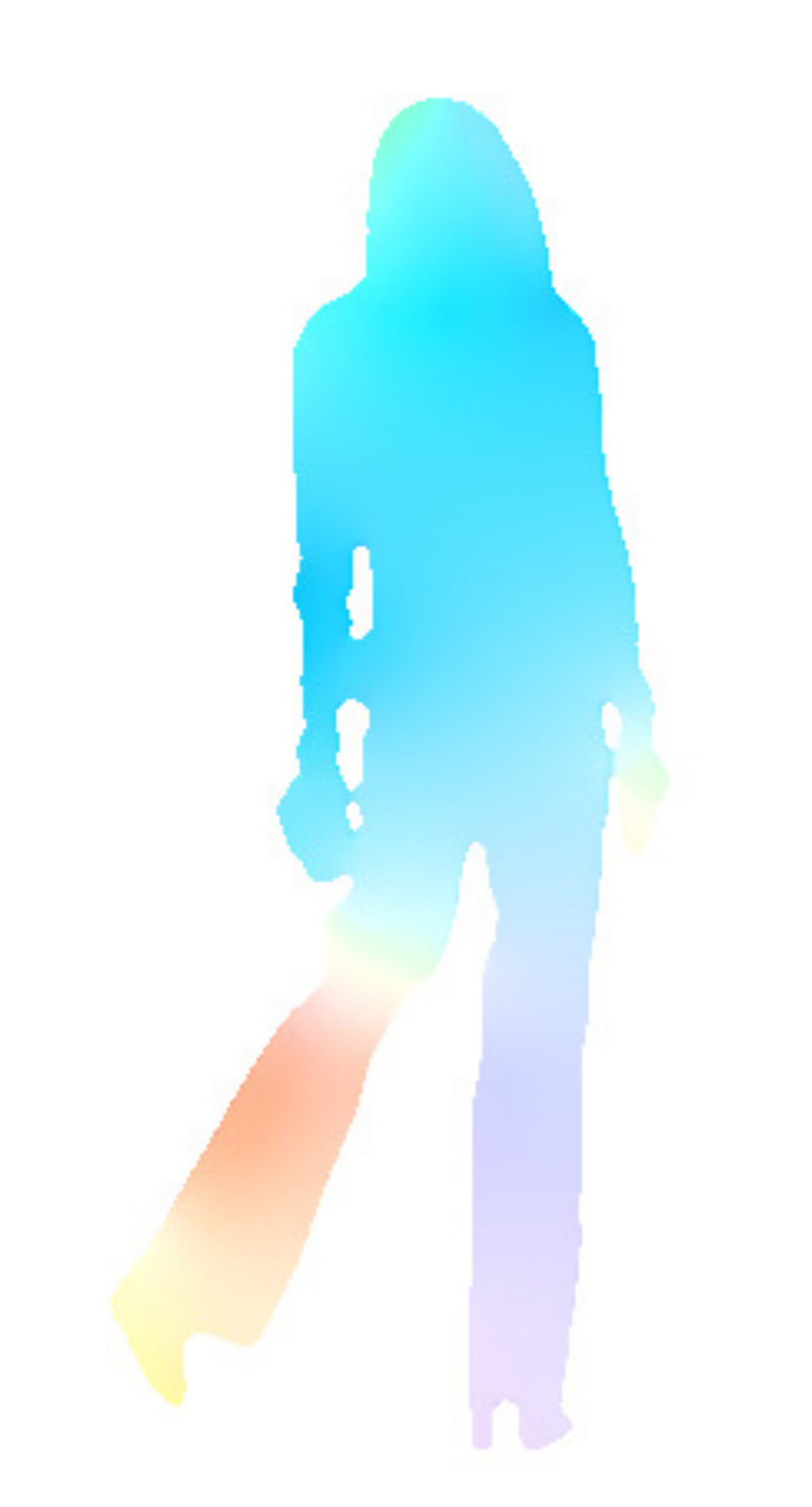} &
    \includegraphics[width=\fWidth,height=\fHeight]{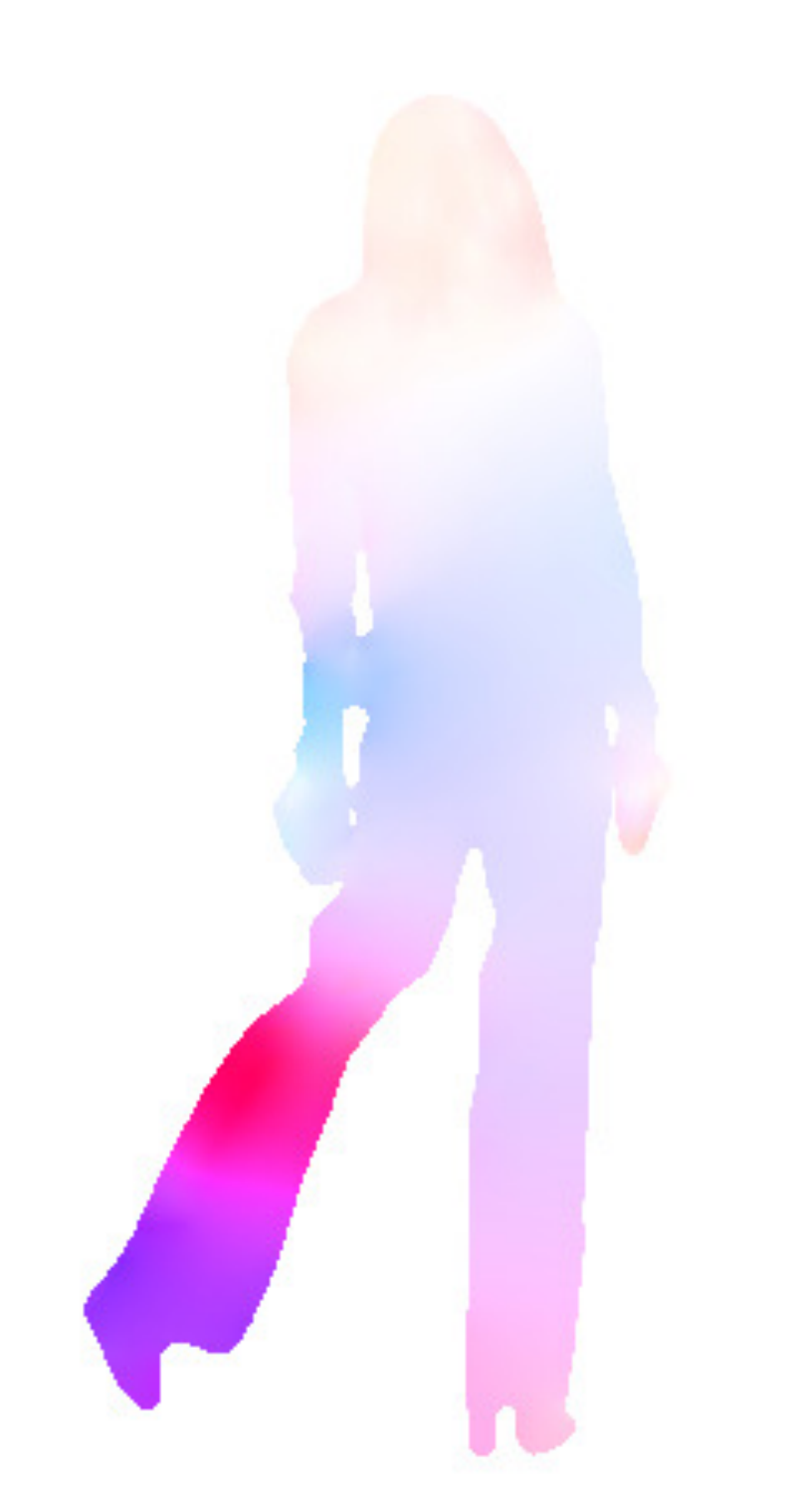} \\
    \multicolumn{7}{c} { Energy regularization, $\gamma=100$ } \\
    \includegraphics[width=\fWidth,height=\fHeight]{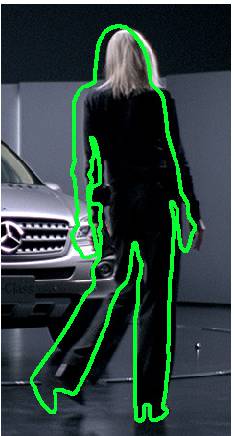} &
    \includegraphics[width=\fWidth,height=\fHeight]{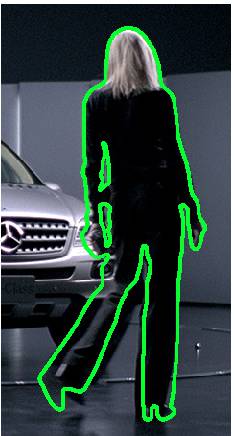} &
    \includegraphics[width=\fWidth,height=\fHeight]{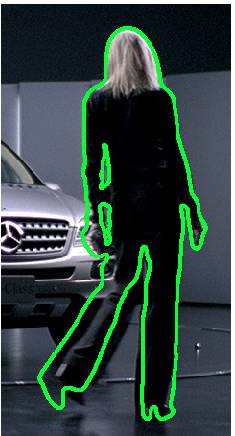} &

    \includegraphics[width=\fWidth,height=\fHeight]{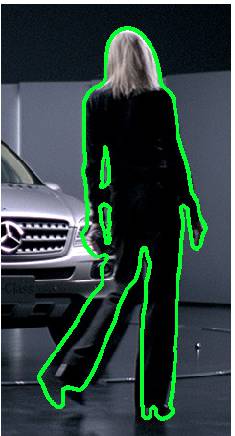} &
    \includegraphics[width=\fWidth,height=\fHeight]{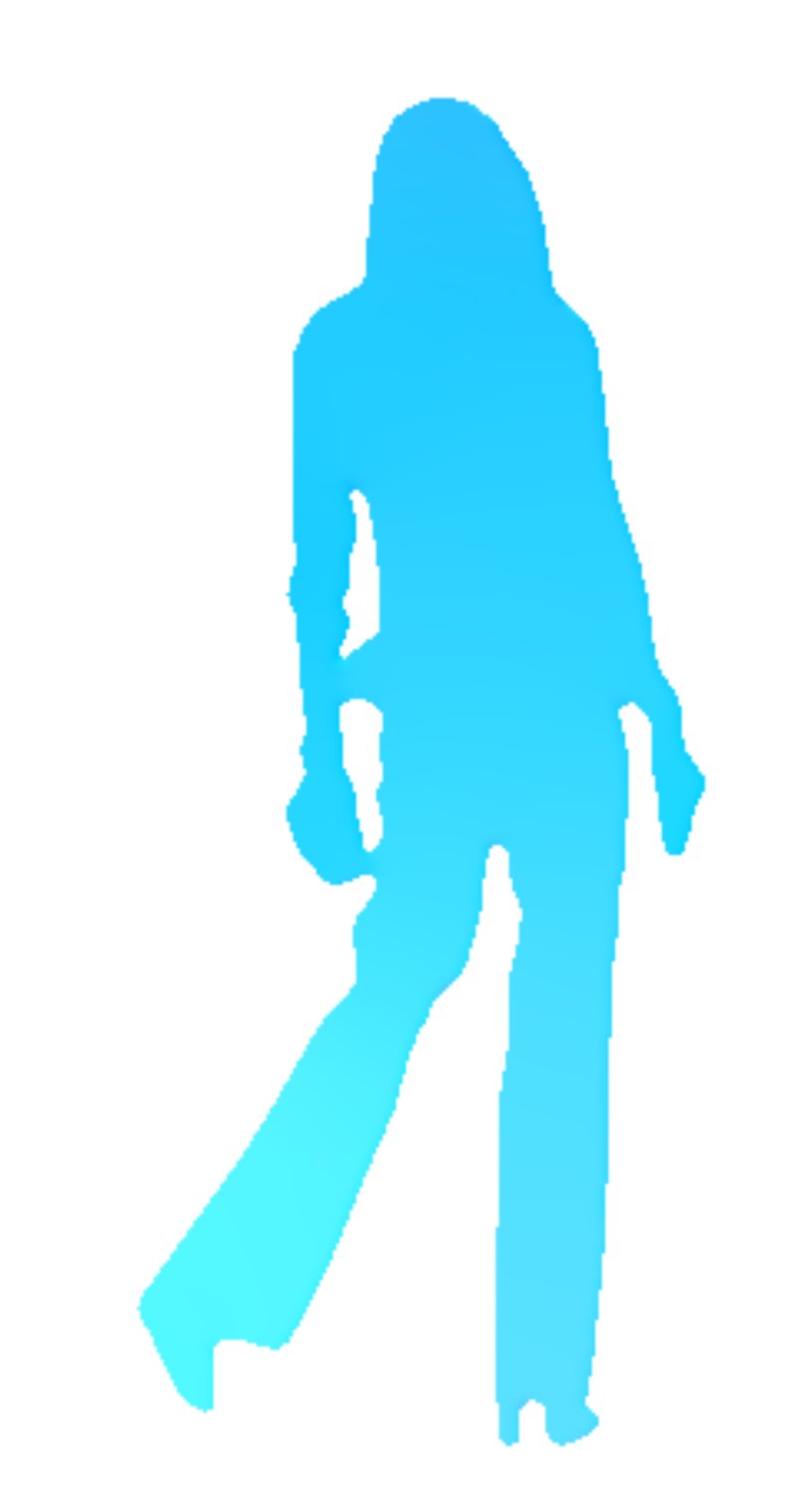} &
    \includegraphics[width=\fWidth,height=\fHeight]{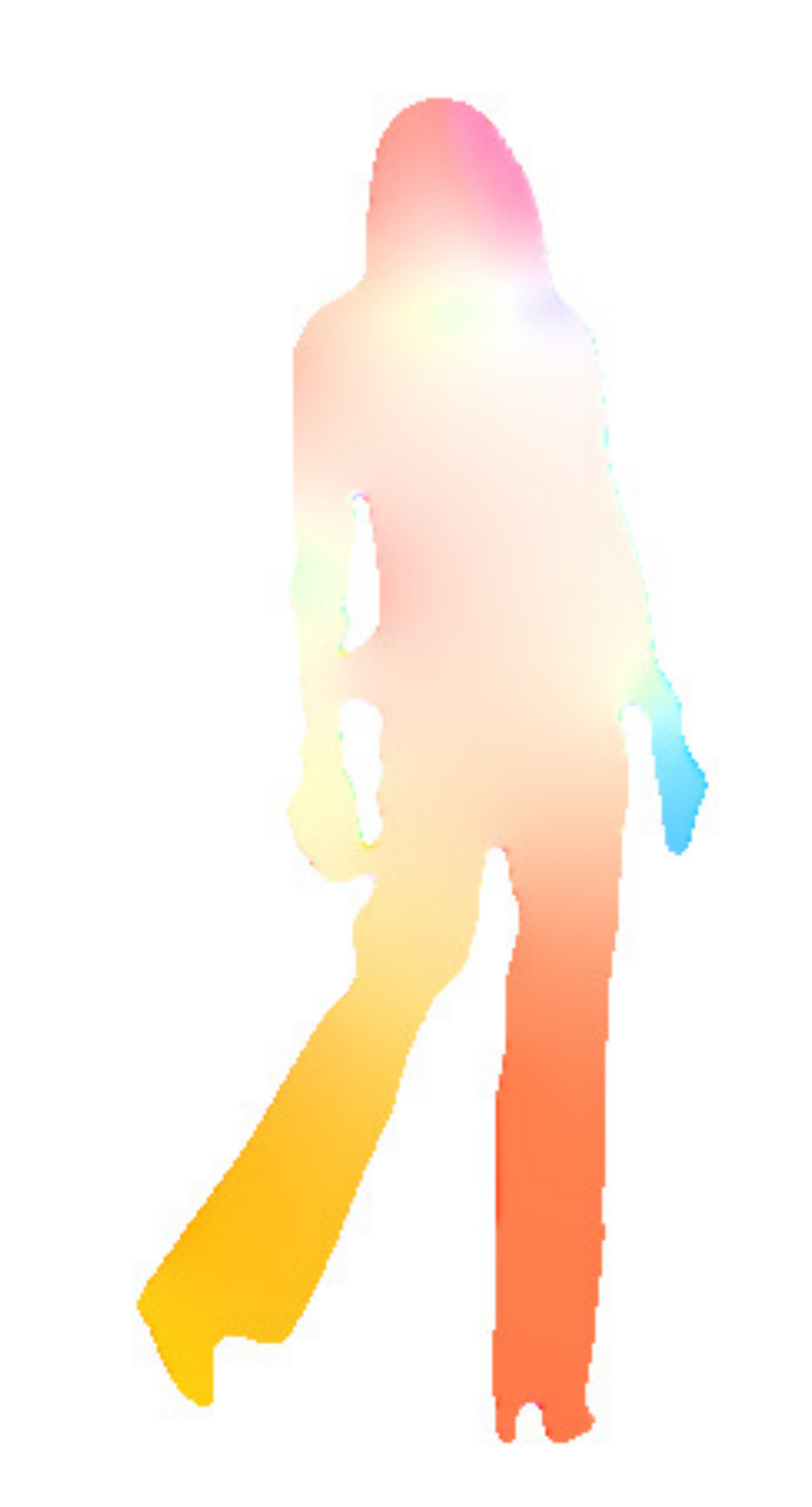} &
    \includegraphics[width=\fWidth,height=\fHeight]{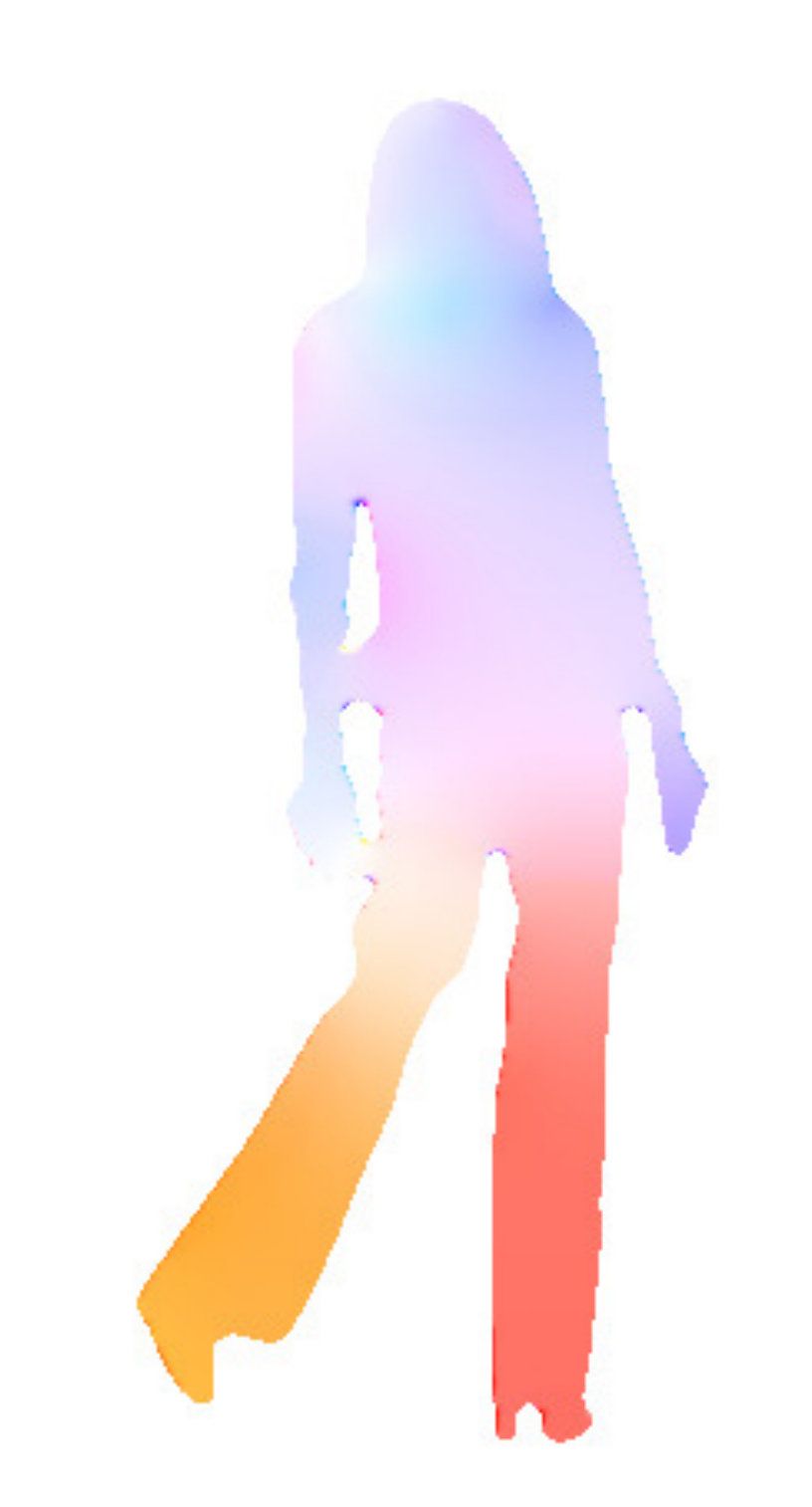}
    \\
    \multicolumn{7}{c} { Energy regularization, $\gamma=500$ } \\
    \includegraphics[width=\fWidth,height=\fHeight]{\fPatha/coutourevolve001} &
    \includegraphics[width=\fWidth,height=\fHeight]{\fPatha/coutourevolve002} &
    \includegraphics[width=\fWidth,height=\fHeight]{\fPatha/coutourevolve003} &

    \includegraphics[width=\fWidth,height=\fHeight]{\fPatha/coutourevolve004} &
    \includegraphics[width=\fWidth,height=\fHeight]{\fPatha/colormap001} &
    \includegraphics[width=\fWidth,height=\fHeight]{\fPatha/colormap002} &
    \includegraphics[width=\fWidth,height=\fHeight]{\fPatha/colormap003} \\
    \multicolumn{7}{c} { Coarse-to-Fine Region-Based Sobolev Descent (\emph{Parameter-free}) } \\
    \includegraphics[width=\fWidth,height=\fHeight]{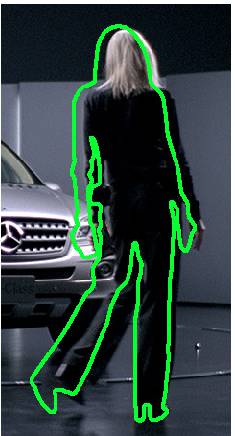} &
    \includegraphics[width=\fWidth,height=\fHeight]{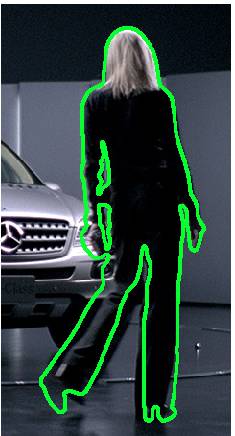} &
    \includegraphics[width=\fWidth,height=\fHeight]{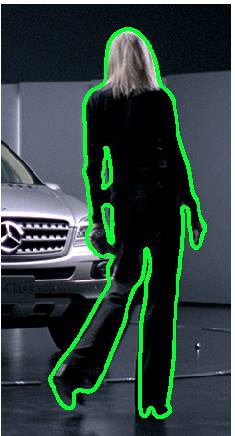} &
    \includegraphics[width=\fWidth,height=\fHeight]{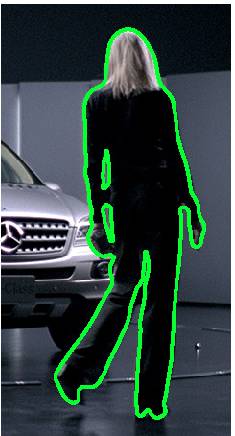} &
    \includegraphics[width=\fWidth,height=\fHeight]{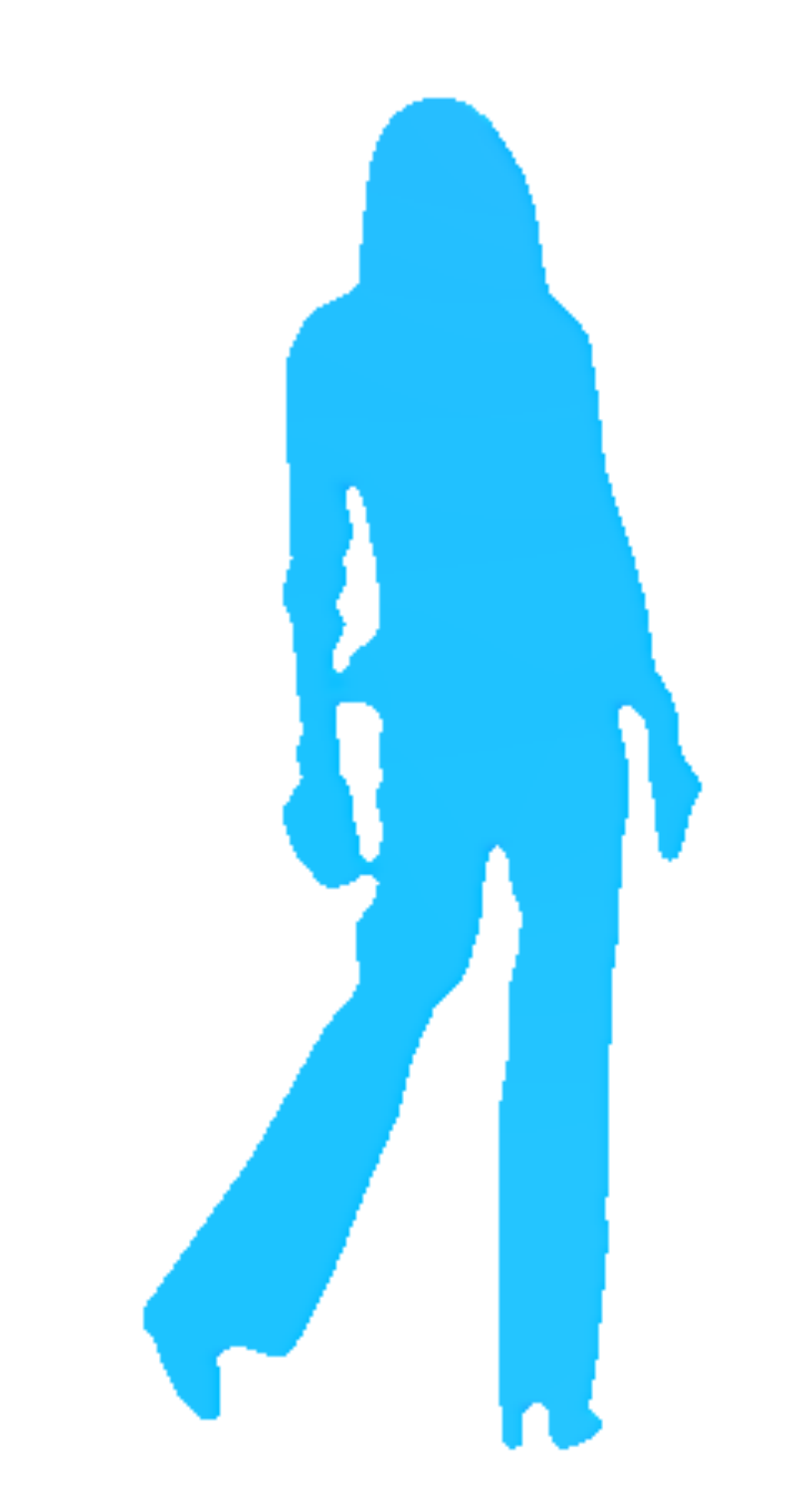} &
    \includegraphics[width=\fWidth,height=\fHeight]{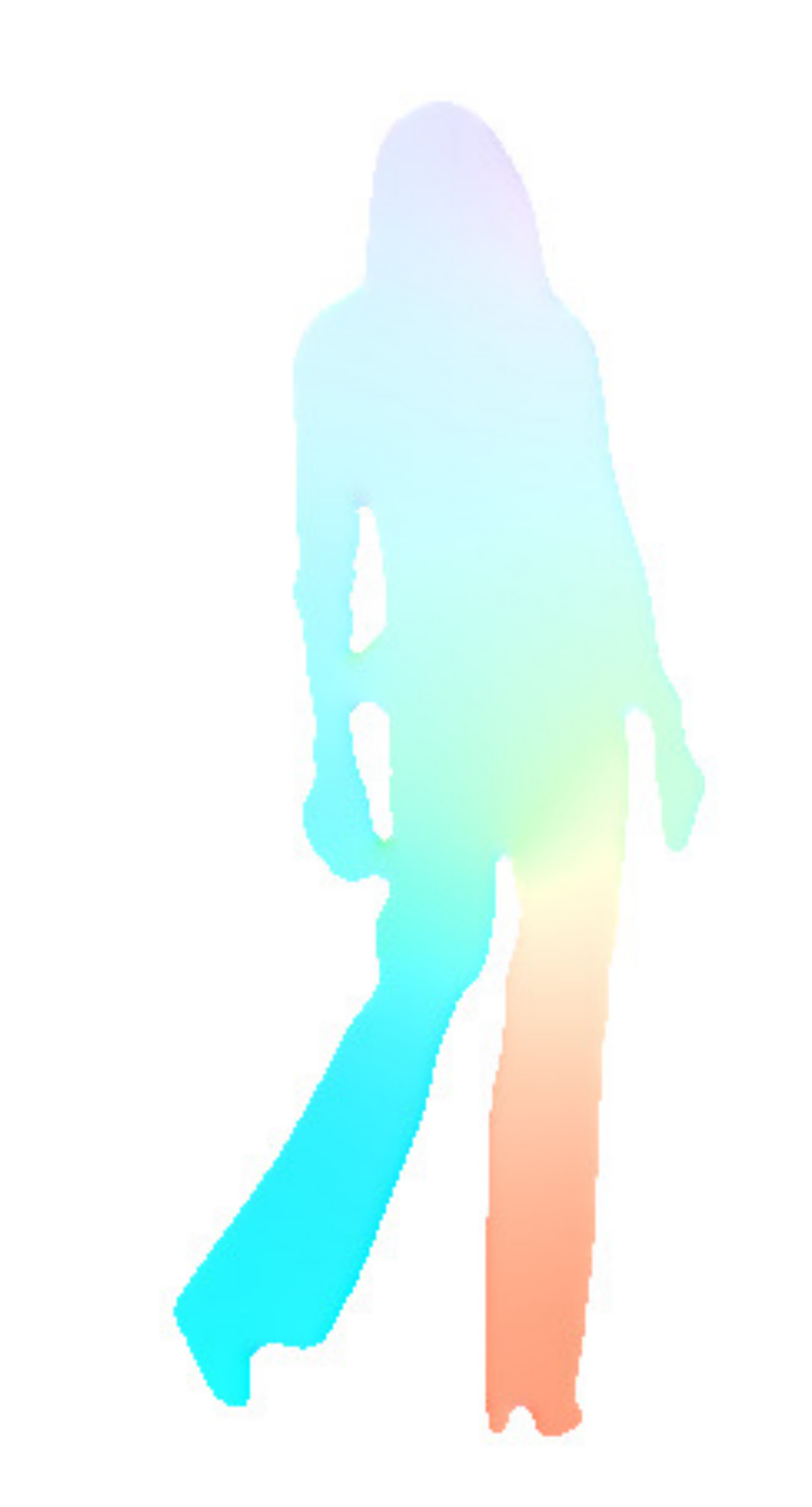} &
    \includegraphics[width=\fWidth,height=\fHeight]{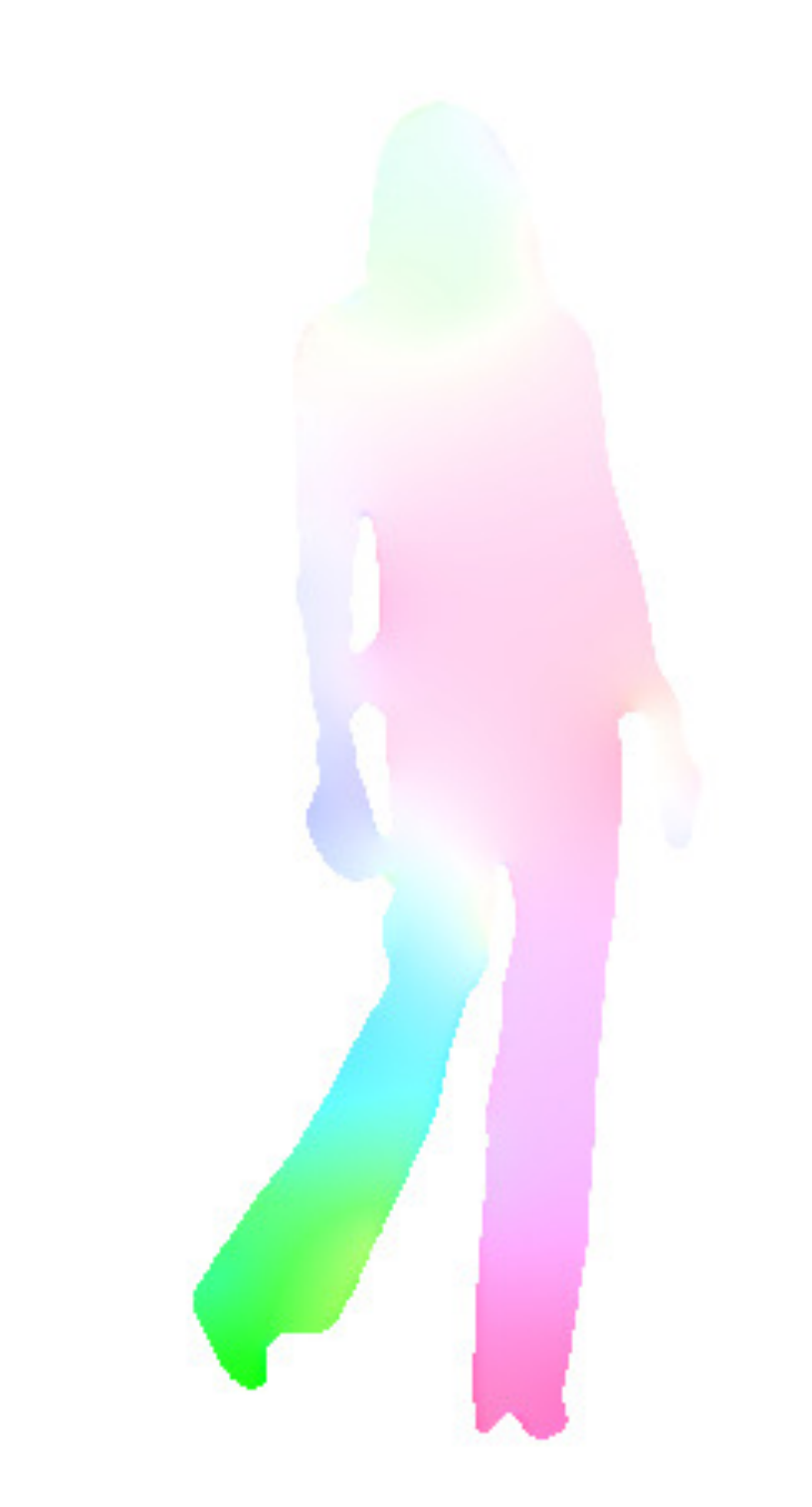}
  \end{tabular}
}
%\vspace{-0.15in}
\caption{{\bf Coarse-to-Fine Behavior of Region-Based Sobolev
    Descent}. Matching a template (obtained from $I_1$) to $I_2$ from
  Figure~\ref{fig:coarse_to_fine_images} using regularization of the
  velocity field in the energy, and Sobolev descent. In each row, the
  evolution (until convergence) is shown. [First four images]:
  $\partial R_{\tau}$ on $I_2$ for various snapshots $\tau$. [Last
  three images]: displacement of object between adjacent snapshots (in
  optical flow color code).  Small $\gamma$ favors fine deformations
  and is sensitive to intermediate structures, whereas large $\gamma$
  favors only coarse deformations and cannot capture regions with
  fine-scale deformations, e.g., legs. Sobolev descent captures all
  scales of deformation without being sensitive to intermediate
  structures.}
  \label{fig:coarse_to_fine}
\end{figure}

\def\fHeight{2in}
\def\fHeightt{1.92in}
\begin{figure}
  \centering
  \includegraphics[clip,trim=25 7 15 160, totalheight=\fHeightt]{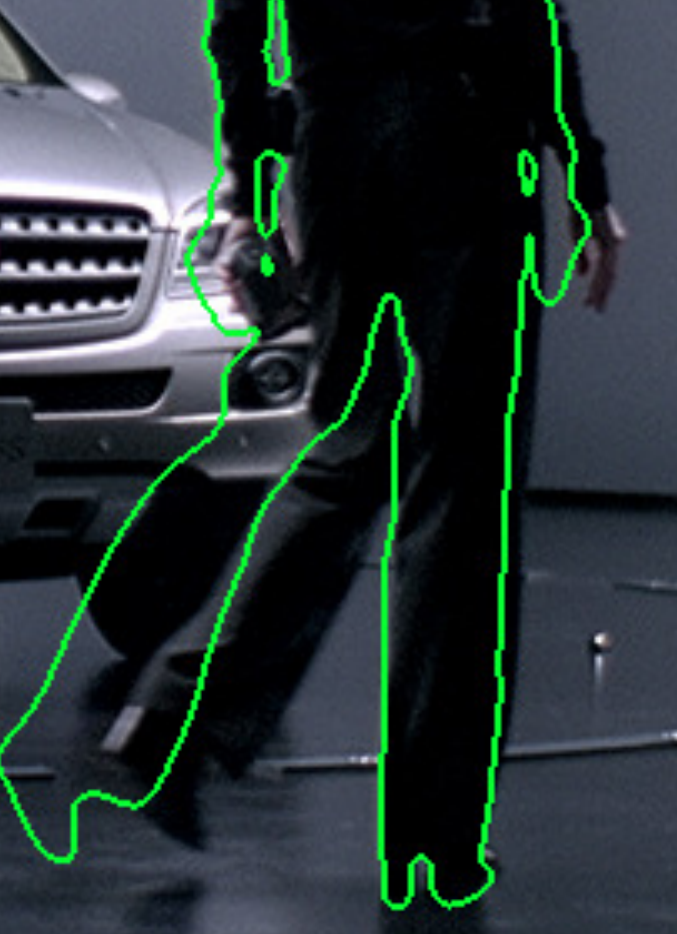}
  \includegraphics[clip,trim=25 7 15 160, totalheight=\fHeightt]{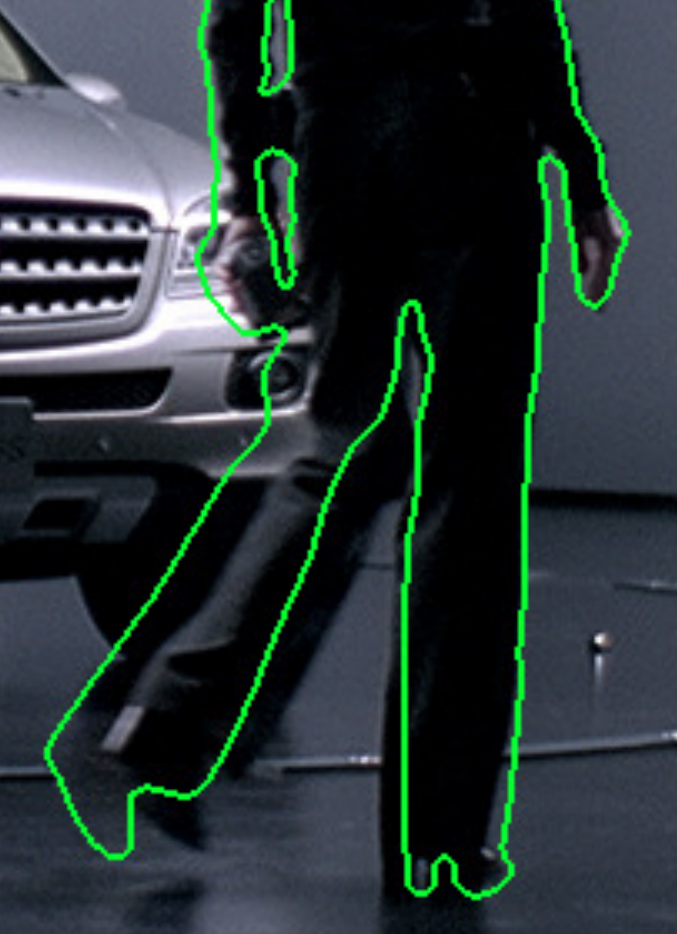}
  \includegraphics[clip,trim=25 7 15 160, totalheight=\fHeightt]{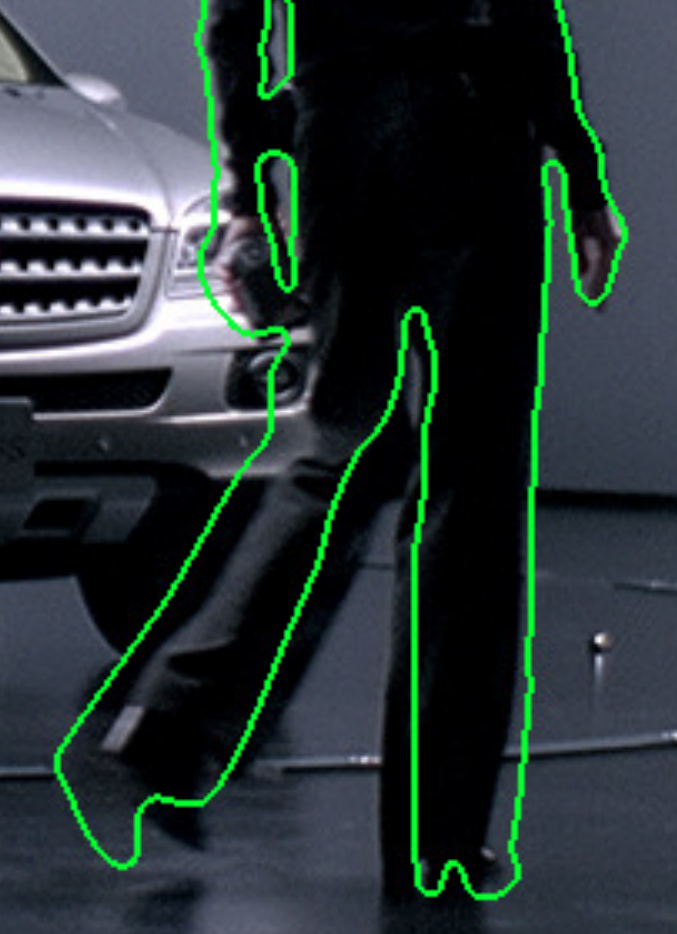}
  \includegraphics[clip,trim=25 7 15 160, totalheight=\fHeightt]{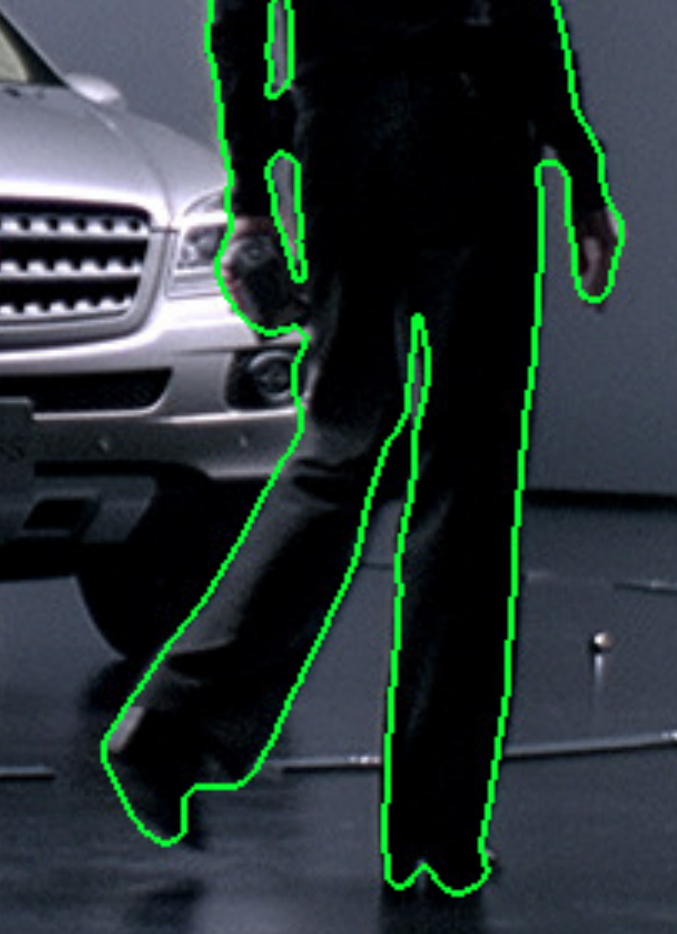}\\\vspace{0.05in}
  \includegraphics[clip,trim=25 7 15 160, totalheight=\fHeightt]{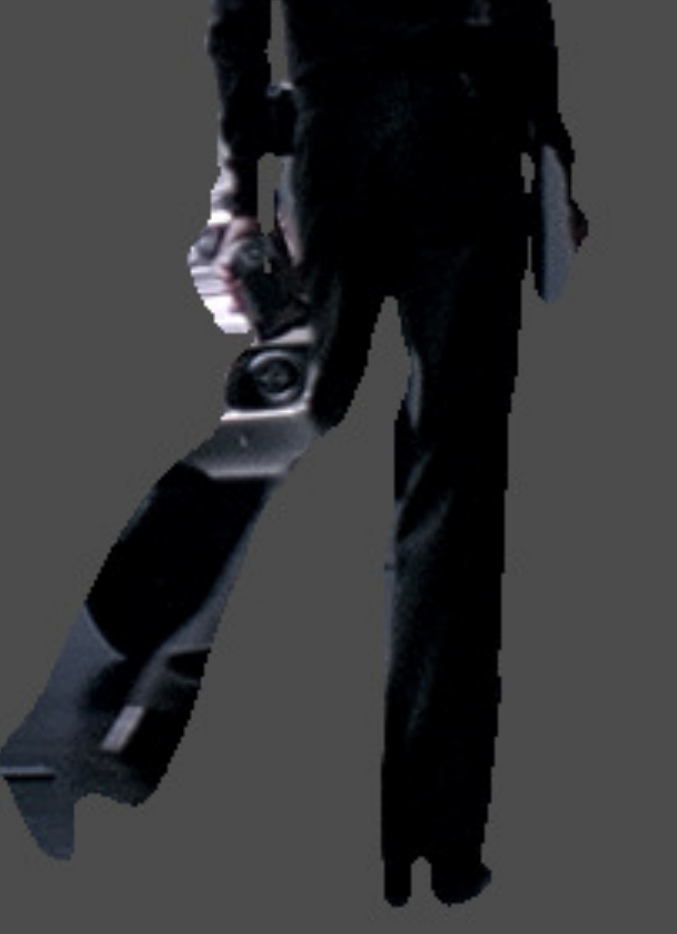}
  \includegraphics[clip,trim=25 7 15 160, totalheight=\fHeightt]{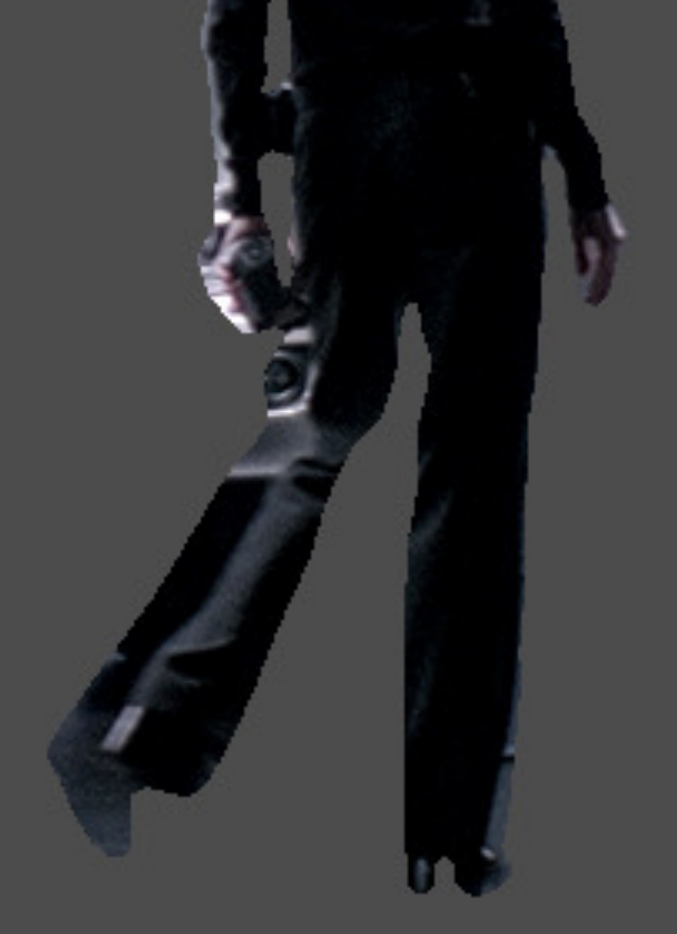}
  \includegraphics[clip,trim=25 7 15 160, totalheight=\fHeightt]{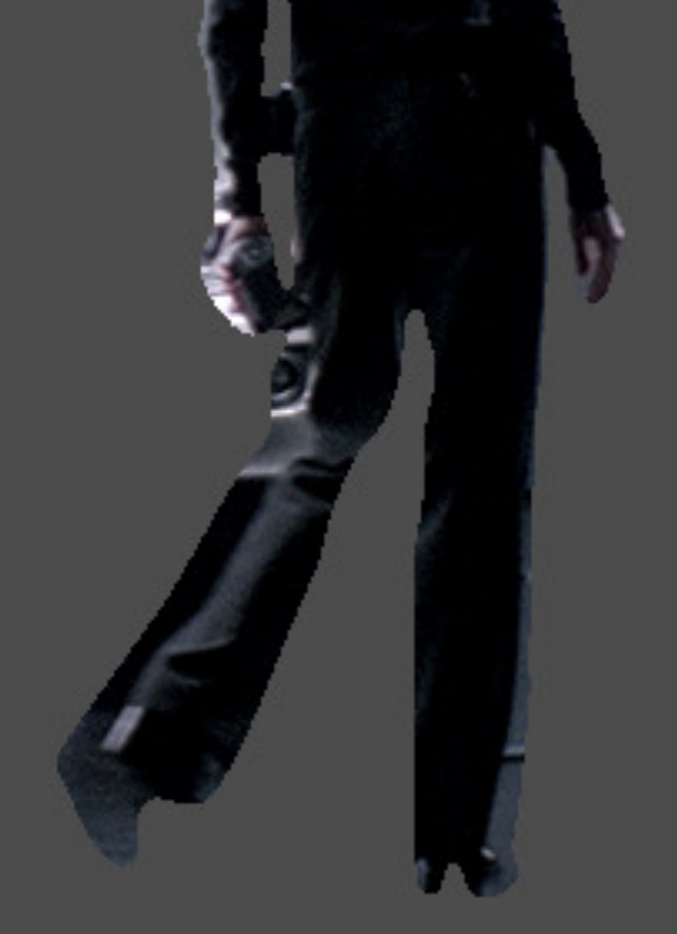}
  \includegraphics[clip,trim=25 7 15 160, totalheight=\fHeightt]{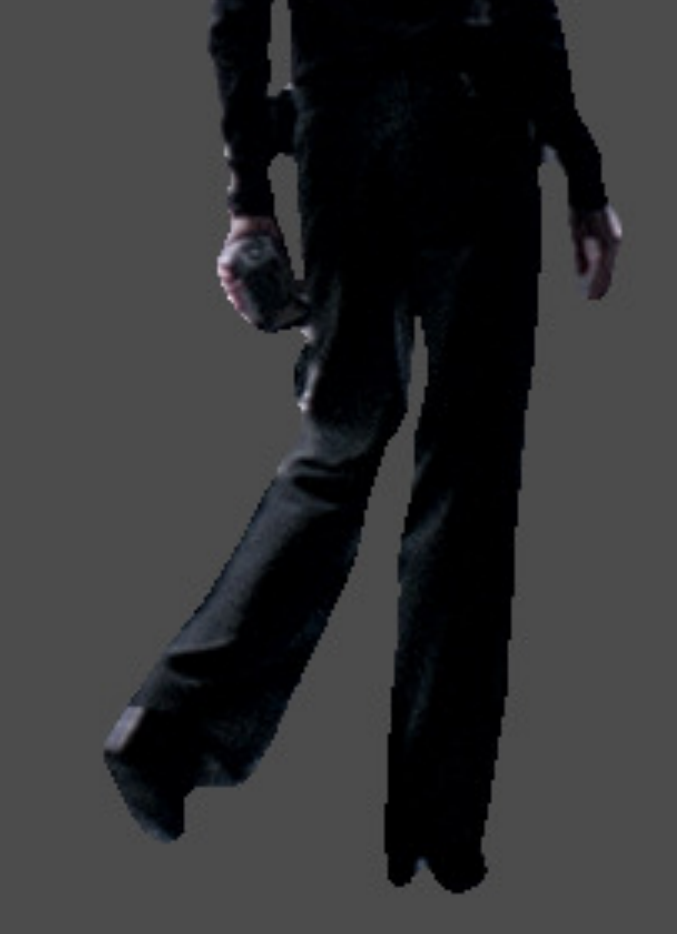}
  \caption{Zoom of final converged results of experiment of
    Figure~\ref{fig:coarse_to_fine}. [Top row]: boundary of converged
    region on $I_2$, [Bottom row]: cutout of object in $I_2$.  [Left]:
    energy regularization $\gamma=10$, [Middle-left]: energy
    regularization $\gamma=100$, [Middle-right]: energy regularization
    $\gamma=500$, [Right]: region-based Sobolev. Notice that small
    $\gamma$ misses regions of coarse motion, larger $\gamma$ obtains
    regions of coarse motion, but misses regions where finer
    deformation occurs. Sobolev obtains both coarse and fine
    deformations.}
  \label{fig:zoom_coarse_to_fine}
\end{figure}

\section{Occlusion/Dis-occlusion Computation and Alternating
  Optimization}
\label{sec:final_algorithm}

We now describe the alternating optimization scheme to optimize $E_o$,
combining the coarse-to-fine optimization scheme described in the
previous section, and optimization in the occlusion, which we describe
next. We then present the optimization scheme to determine the
dis-occlusion.

\subsection{Joint Occlusion and Warp Optimization}
\label{sec:final_optimization_warp_occlusion}

Note that given an estimate $w$, one can solve for a global optimizer
of the energy $E_o$. Indeed, the energy can be written as 
\begin{equation}
  E_o(O | w ; I, a, R) = 
  \int_{R\backslash O} \rho( (I(w(x))-a(x))^2 ) \ud x + \int_O \beta_o
  \ud x, \, O\subset R.
\end{equation}
The optimization problem can be thought of as an assigment problem
where points $x\in R$ are assigned to the occlusion $O$ or the
co-visible region $R\backslash O$. If $x$ is assigned to $O$,
then it adds to the energy an amount $\beta_o$, whereas, if it is
assigned to $R\backslash O$, it adds to the energy an amount $\rho(
(I(w(x))-a(x))^2 )$. Therefore to minimize the energy, we assign
pixels to the occlusion based on 
\begin{align}
  O & = \{ x\in R \, :\, \rho( (I(w(x))-a(x))^2 ) > \beta_o \} \\
  & = 
  w^{-1} \{ x\in w(R) \, : \, \rho( (I(x)-a(w^{-1}(x))^2 ) ) >
  \beta_o \}, 
\end{align}
which is a global optimizer of $E_o$ conditioned on $w$.

The alternating scheme to optimize $E_o$ in both $O$ and $w$ is then a
modification of the scheme presented in
Sub-Section~\ref{subsec:gradient_descent_warp} to update the occlusion
during the evolution. The scheme is as follows:
\begin{align}
  \label{eq:init_alternate_opt_w}
  \Psi_0(x) &= d_R(x), \, x\in B_2(R) \\
  \phi_{0}^{-1}(x) &= x, \, x\in R \\
  R_0 &= R \\
  \tilde O_0 &= \emptyset \\
  G_{\tau} &= \nabla_w E(\phi_{\tau}|O_{\tau}, R_{\tau}, I)\\
  \label{eq:backward_map_evolution}
  \partial_{\tau} \phi_{\tau}^{-1} &= \nabla\phi_{\tau}^{-1}(x)\cdot
  G_{\tau}(x), x\in R_{\tau}\\
  \label{eq:level_set_evolution}
  \partial_{\tau} \Psi_{\tau} &= \nabla \Psi_{\tau}(x) \cdot
  G_{\tau}(x), x\in B_2(R_{\tau}) \\
  R_{\tau} &= \{ \Psi_{\tau} < 0 \} \\
  \label{eq:end_alternate_opt_w}
  \tilde O_{\tau} &= \{ x\in R_{\tau} \, : \, \rho( (I(x)-a(\phi_{\tau}^{-1}(x))^2 ) ) >
  \beta_o \}, 
\end{align}
where $\tilde O_{\tau}$ indicates the current estimate of the warped
occlusion $O_{\tau}$, i.e., $\tilde
O_{\tau}=\phi_{\tau}(O_{\tau})$. Note that only $\tilde O_{\tau}$ is
needed to compute the gradient $G_{\tau}$, and thus we do not
explicitly compute $O_{\tau}$.  Note that $G_{\tau}$ is specified by
$\mean{G_{\tau}}$ and $\tilde G_{\tau}$, where $\tilde G_{\tau}$
satisfies
\begin{equation}
  \begin{cases}
    -\Delta \tilde G_{\tau}(x) = f_1(x,\phi_{\tau}^{-1}(x))
    \det{(\nabla \phi_{\tau}^{-1}(x))}^{-1} - 
      \mean{ f_1(\cdot,\phi_{\tau}^{-1}(\cdot)) \det{(\nabla
          \phi_{\tau}^{-1} (\cdot))}^{-1}  } & x\in R_{\tau} \\
      \nabla \tilde G_{\tau}(x) \cdot N = 0  & x\in \partial R_{\tau} \\
      \mean{\tilde G_{\tau}}=0 & 
    \end{cases}.
\end{equation}
where 
\begin{equation}
  \mean{G_{\tau}} = \int_{R} f_1(x,\phi_{\tau}^{-1}(x)) \det{(\nabla
    \phi_{\tau}^{-1}(x))}^{-1} \ud x,
\end{equation}
and 
\begin{align}
  f_1(x,\phi_{\tau}^{-1}(x)) &= \rho'( |I(x)-a_{\tau}(x) )|^2 )
  (I(x)-a_{\tau}(x) )\nabla I(x) \chi_{\tilde O_{\tau}}(x), \quad x\in R_{\tau} \\
  a_{\tau}(x) &= a(\phi_{\tau}^{-1}(x)), \quad x\in R_{\tau}
\end{align}
Discretization of
\eqref{eq:init_alternate_opt_w}-\eqref{eq:end_alternate_opt_w} and
numerical implementation is given in Appendix~\ref{app:numerics}.

Let $\tau=\tau_{\infty}$ be the time of convergence,
$R_{\tau_{\infty}}$ - a warping of $R$ includes a warping of the
occluded region $O_{\tau_{\infty}}$, and thus the warping of the
un-occluded region is $ w(R\backslash
O_{\tau_{\infty}})=R'_{_{\tau_{\infty}}}=R_{_{\tau_{\infty}}}\backslash
\tilde O_{_{\tau_{\infty}}}$, and does not include the disoccluded region,
which is computed in the next section from $R'_{_{\tau_{\infty}}}$. To
ensure spatial regularity of $R'_{_{\tau_{\infty}}}$, at convergence
of \eqref{eq:init_alternate_opt_w}-\eqref{eq:end_alternate_opt_w}, we
induce spatial regularity into $O_{\tau_{\infty}}$ by using the
estimate
\begin{align}
  \label{eq:occl_thres}
  \tilde O_{\tau_{\infty}} &=  \{ x\in R_{\tau_{\infty}} \, : \,
  (G_{\sigma}\ast \mbox{Res} )(x) >
  \beta_o \} \\
  \mbox{Res}(x) &= \rho( (I(x)-a(\phi_{\tau_{\infty}}^{-1}(x)))^2  )
\end{align}
where $G_{\sigma}$ denotes an isotropic Gaussian kernel.

Fig.~\ref{fig:occl_def_estimation} shows the evolution
\eqref{eq:init_alternate_opt_w}-\eqref{eq:end_alternate_opt_w} on an
example, and the final co-visible region $R'_{_{\tau_{\infty}}}$.

\def\fHobEx{figures2/example_hobits}
\def\fHobExS{0.48in}
\def\fHobExSc{0.8in}

\begin{figure}
  \centering
  \begin{minipage}{7cm}
  \includegraphics[clip,trim=230 0 120 0,totalheight=\fHobExSc]{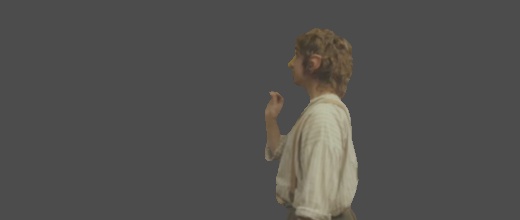}
  \includegraphics[clip,trim=230 0 120 0,totalheight=\fHobExSc]{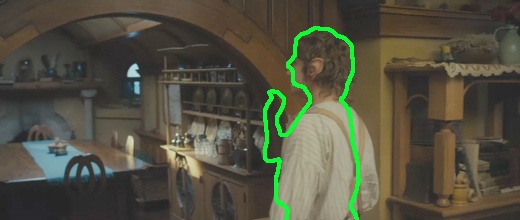}
  \includegraphics[clip,trim=230 0 120 0,totalheight=\fHobExSc]{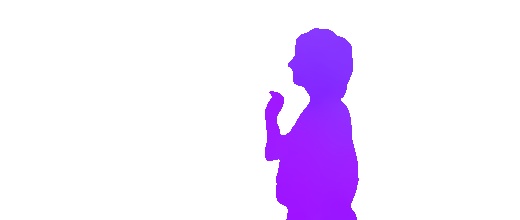}
  \includegraphics[clip,trim=230 0 120 0,totalheight=\fHobExSc]{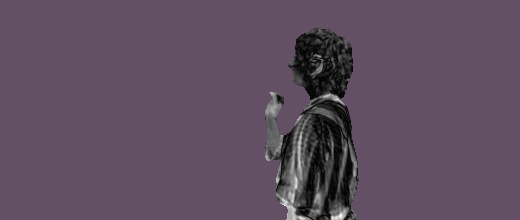}\\
  \includegraphics[clip,trim=230 0 120 0,totalheight=\fHobExSc]{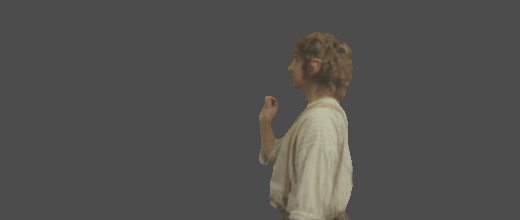}
  \includegraphics[clip,trim=230 0 120 0,totalheight=\fHobExSc]{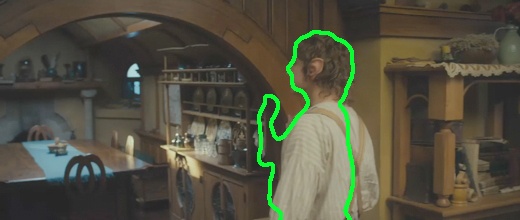}
  \includegraphics[clip,trim=230 0 120 0,totalheight=\fHobExSc]{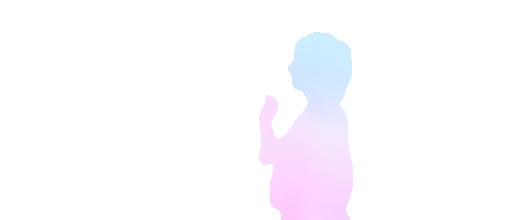}
  \includegraphics[clip,trim=230 0 120 0,totalheight=\fHobExSc]{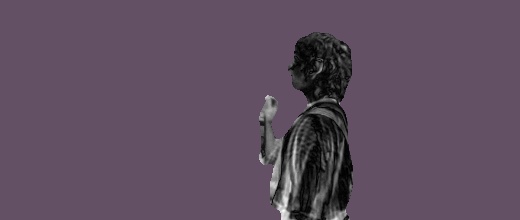}\\
  \includegraphics[clip,trim=230 0 120 0,totalheight=\fHobExSc]{\fHobEx/c/templateevolve/templateevolve076}
  \includegraphics[clip,trim=230 0 120 0,totalheight=\fHobExSc]{\fHobEx/c/contourevolve/coutourevolve076}
  \includegraphics[clip,trim=230 0 120 0,totalheight=\fHobExSc]{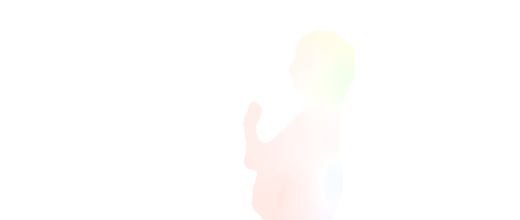}
  \includegraphics[clip,trim=230 0 120 0,totalheight=\fHobExSc]{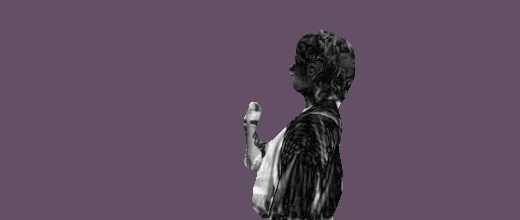}
  \end{minipage}%
  \begin{minipage}{0.6cm}
    \includegraphics[clip,trim=120 0 0 0,angle=180,totalheight=0.7in]{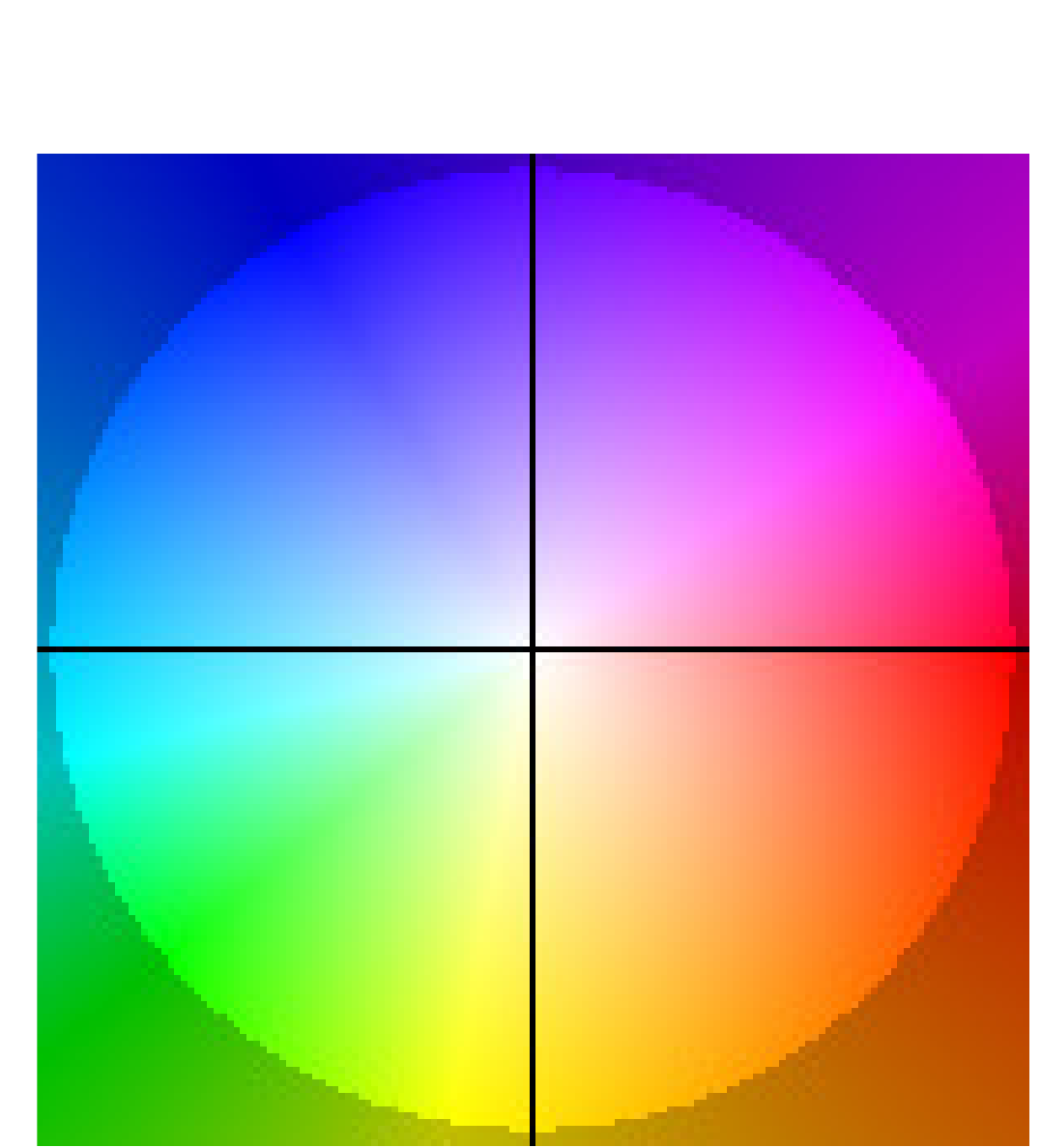}
  \end{minipage}
  \caption{{\bf Occlusion estimation and warping.} [Top to bottom]:
    Beginning ($\tau=0$), intermediate, and final stages of evolution.
    [$1^{\text{st}}$ column]: radiance $a_{\tau}$,
    [$2^{\text{nd}}$]: target image $I$ and boundary of $R_{\tau}$,
    [$3^{\text{rd}}$]: velocity $-G_{\tau}$, [$4^{\text{th}}$]:
    occlusion estimation $\mbox{Res}$ at time $\tau$, [$5^{\text{th}}$]:
    optical flow color code. The final occluded region is shown in
    Fig.~\ref{fig:overview}(b). }
  \label{fig:occl_def_estimation}
\end{figure}

\subsection{Dis-Occlusion Optimization}
\label{subsec:disocclusion_optimization}

We show how to optimize the dis-occlusion energy $E_d$
\eqref{eq:energy_d}.  The global minimum of $E_d$ is computed in a
thresholding step from the likelihood $p$. Since $p$ decreases
exponentially with distance to $R'$, we assume that $D \subset \{ 0 <
d_{R'} < \varepsilon \}$. The dis-occlusion is computed as
\begin{equation} \label{eq:disocc_comp}
  D = \{ 
  x \, : \, d_{R'}(x)\in (0,\varepsilon] , (G_{\sigma} \ast p)(x) > \beta_d \}
\end{equation}
where $\sigma=0$ corresponds to the global optimum, but to ensure
spatial regularity of $D$, we choose $\sigma>0$. The choice of
$\beta_d$ is based on the frame-rate of the camera and the speed of
the object (the more the speed and the less the frame-rate, the
smaller $\beta_d$).  Fig.~\ref{fig:disocclusion_example} shows an
example of $p$, the dis-occlusion detected, and the final estimate of
the region.

Computation of $d_{R'}$ in $\{ 0 < d_{R'} < \varepsilon \}$ is done
efficiently with the Fast Marching Method \cite{sethian1996fast}, and
$\mbox{cl}(x)$ at each point is simultaneously propagated as the front
in the Fast Marching Method evolves. Then $p$ is readily computed.

\def\fHobExSb{0.8in}
\begin{figure}
  \centering
  \includegraphics[clip,trim=230 0 120 0,totalheight=\fHobExSb]{\fHobEx/c/template_after_occlusiondetection}
  \includegraphics[clip,trim=230 0 120 0,totalheight=\fHobExSb]{\fHobEx/a/Image_t+1}
  \includegraphics[clip,trim=230 0 120 0,totalheight=\fHobExSb]{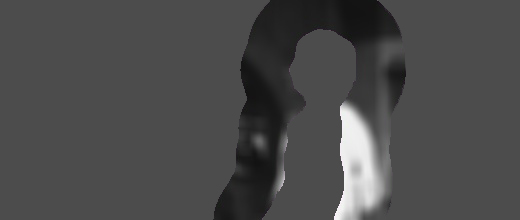}
  \includegraphics[clip,trim=230 0 120 0,totalheight=\fHobExSb]{\fHobEx/d/detected_disocclusion}
    \includegraphics[clip,trim=230 0 120 0,totalheight=\fHobExSb]{\fHobEx/d/Final_Appearance}
    \caption{{\bf Illustration of disocclusion detection.}
      [$1^{\text{st}}$]: warped un-occluded radiance defined on $R'$
      (after occlusion and deformation computation),
      [$2^{\text{nd}}$]: target image $I$, [$3^{\text{rd}}$]:
      likelihood of dis-occlusion map $p$ (defined in
      $B_{R'}(\varepsilon)$), [$4^{\text{th}}$]: computed dis-occlusion
      $D$ (white), and [$5^{\text{th}}$]: final radiance. Boundary of
      final region super-imposed on $I$ is in Fig.~\ref{fig:overview}
      (d).}
  \label{fig:disocclusion_example}
\end{figure}

\section{Filtering Radiance Across Frames}
\label{sec:filtering}
We integrate the results of occlusion/deformation estimation and
dis-occlusion estimation into a final estimate of the shape and
radiance in each frame. To deal with modeling noise (specified in
\eqref{eq:appearance_model}), we filter the radiance in time.

Given the image sequence $I_t, \, t=1\ldots, N$ and an initial
template $R_0\subset \Omega, \, a_0 : R_0 \to \R^k$, the
final algorithm is as follows. For $t=1,\ldots,N$, the following steps
are repeated:
\begin{enumerate}
\item Compute the warping of $R_{t-1}$ and $O_{t-1}$:
  $w_{t-1}(R_{t-1})$ and $w_{t-1}(O_{t-1})$, resp., and
  $a_t'=a_{t-1}\circ w_{t-1}^{-1}$ defined on $w_{t-1}(R_{t-1})$ using
  the optimization scheme described in
  Section~\ref{sec:final_optimization_warp_occlusion} with input
  $R_{t-1}, a_{t-1}$ and $I_t$.
\item Given $R'_t = w_{t-1}(R_{t-1})\backslash w_{t-1}(O_{t-1})$, the
  warping of the un-occluded part of $R_{t-1}$, and the image $I_t$,
  compute the dis-occlusion $D_t$ using \eqref{eq:disocc_comp}. The
  estimate of $R_t$ is then $R_t'\cup D_t$.
\item The radiance is then updated as
\begin{equation} \label{eq:appearance_update} 
  \!\!a_{t}(x) =
  \begin{cases} (1-K_a) a_t'(x) +
    K_a I_{t}(x)& x\in R_t' \\ 
    I_{t}(x) & x\in  D_t
  \end{cases}
\end{equation}
where $K_a \in [0,1]$ is the gain.
\end{enumerate}
The averaging of the warped radiance and the current image
\eqref{eq:appearance_update} combats modeling noise $\eta$ in
\eqref{eq:appearance_model}. In practice, $K_a$ is chosen large if the
image is reliable (e.g., no specularities, illumination change, noise,
or any other deviations from brightness constancy), and small
otherwise.

\section{Experiments and Comparisons}

\label{sec:experiments}

We demonstrate our method on a variety of videos that contain
self-occlusions/disocclusions. All examples shown have over 100
frames\footnote{Videos for all experiments and comparisons are
  available at
  \url{http://vision.ucla.edu/∼ganeshs/articulated object tracking html/ ObjectTrackingSelfOcclusions.html}}.
To demonstrate that occlusion/dis-occlusion modeling aids joint
shape/appearance tracking, we compare to Adobe After Effects CS6 2013
(AAE) (based on \cite{bai2009video}, but significantly extended over
several years), which employs localized joint shape and appearance
information without explicit occlusion modeling. Note that AAE has an
interactive component to correct errors in the automated component; we
compare to the automated component to show less interaction would be
required with our approach. To show advantages over tracking using
global statistics, we compare to Scribbles \cite{Fan:PAMI12} (publicly
available code), which is a state-of-the-art technique that employs
global statistics in addition to other advanced techniques.

Parameters are chosen as: $\sigma = 5$ in \eqref{eq:disocc_comp} and
\eqref{eq:occl_thres}, $\sigma_d=100$ in the likelihood, $p$ in
\eqref{eq:p_disc}, the band thickness for the domain of $p$ is
$\varepsilon=30$, and the radius of $B_r$ in $p_{f,x}$ and $p_{b,x}$
is $r=3\varepsilon$ (i.e., a $6\varepsilon\times 6\varepsilon$
window). The threshold for the occlusion stage is $\beta_o =
\mbox{Res}_{min} + 0.3\times (\mbox{Res}_{max} - \mbox{Res}_{min})$
where $\mbox{Res}_{max}$ ($\mbox{Res}_{min}$) denotes the maximum
(minimum) value of smoothed residual. The threshold for the
dis-occlusion stage is $\beta_d=0.5$ when $p$ is normalized to be a
probability. The gain in the radiance update
\eqref{eq:appearance_update} is $K_a=0.8$. Most parameters can be
fixed for the whole video, and work on a wide range. Most significant
parameters are the $\beta$'s, and sensitivity analysis is shown later
(near the end of the Section).

\begin{figure}
  \centering
  \includegraphics[clip,trim=264 0 171 0, totalheight=0.965in]{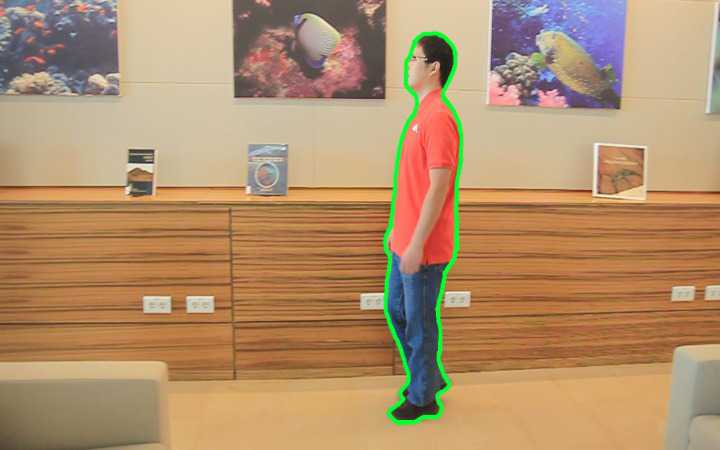}
  \includegraphics[clip,trim=264 0 171 0, totalheight=0.965in]{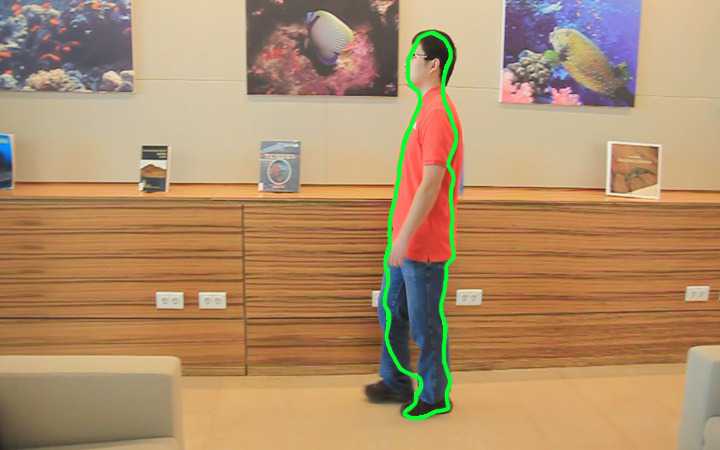}
  \includegraphics[clip,trim=264 0 171 0, totalheight=0.965in]{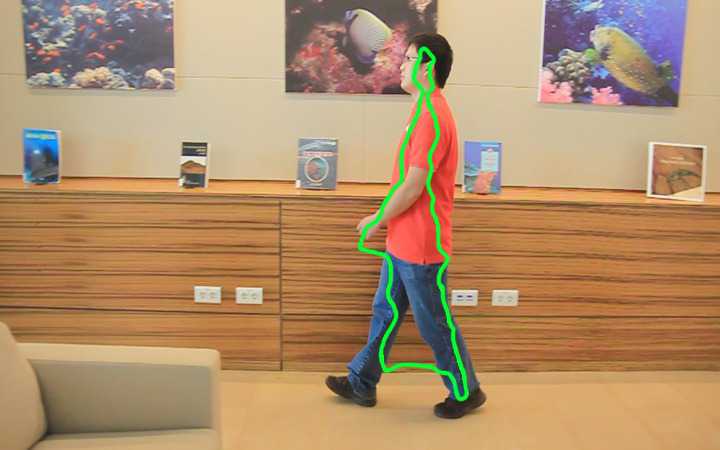}
  \includegraphics[clip,trim=264 0 171 0, totalheight=0.965in]{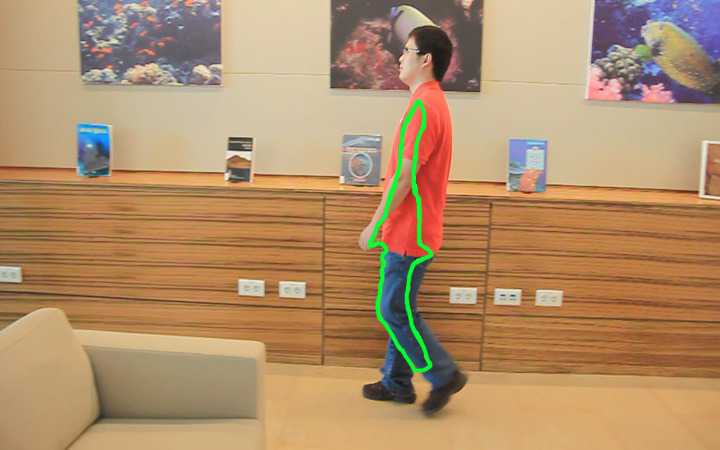}
  \includegraphics[clip,trim=264 0 171 0, totalheight=0.965in]{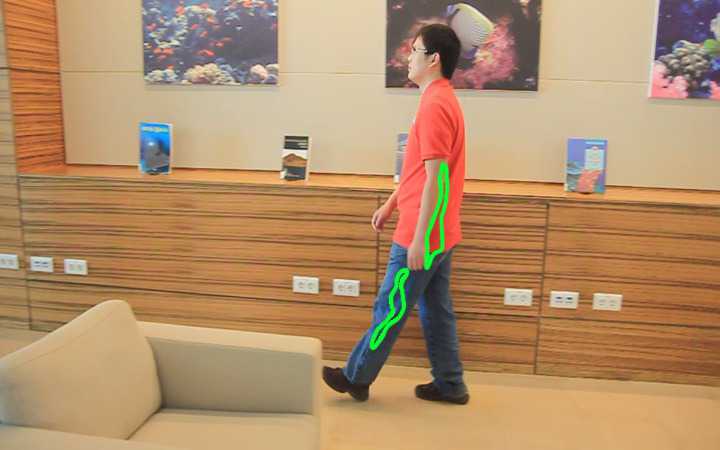}
  \\
  \includegraphics[clip,trim=264 0 171 0, totalheight=0.965in]{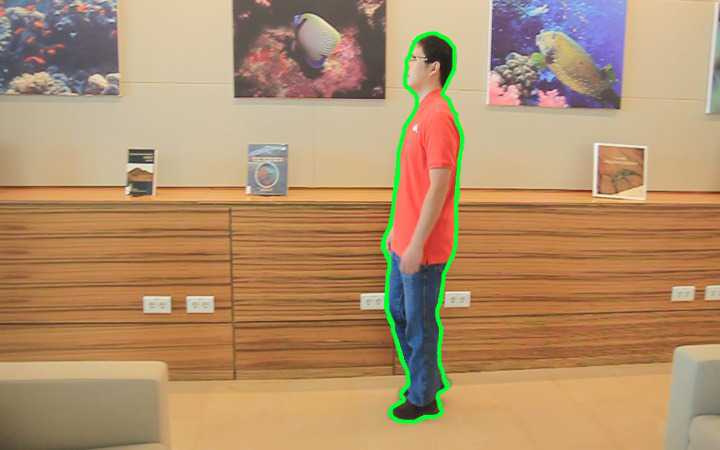}
  \includegraphics[clip,trim=264 0 171 0, totalheight=0.965in]{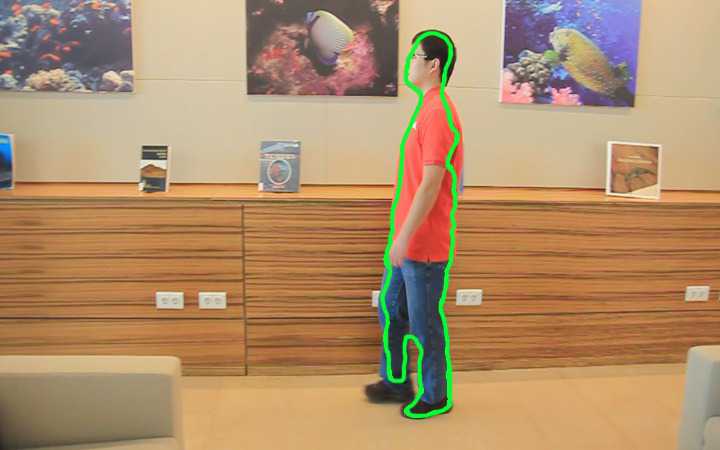}
  \includegraphics[clip,trim=264 0 171 0, totalheight=0.965in]{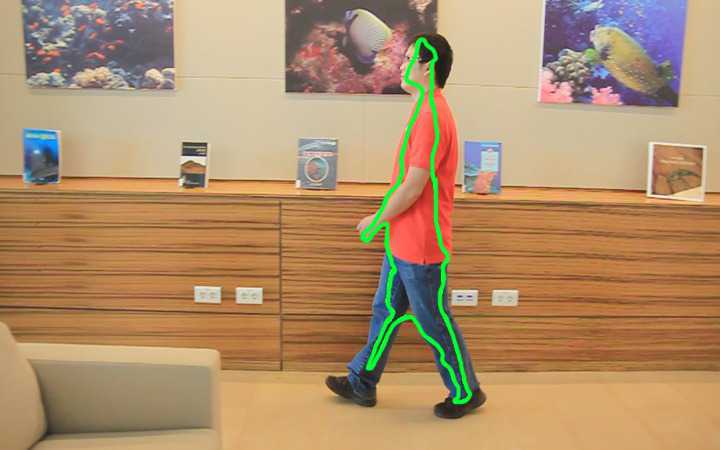}
  \includegraphics[clip,trim=264 0 171 0, totalheight=0.965in]{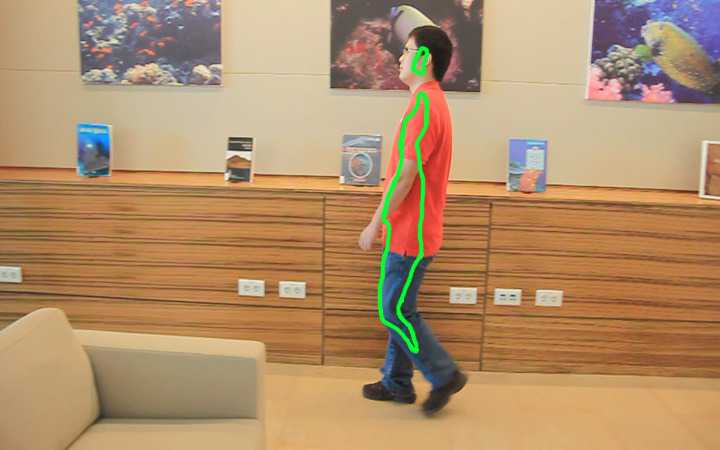}
  \includegraphics[clip,trim=264 0 171 0, totalheight=0.965in]{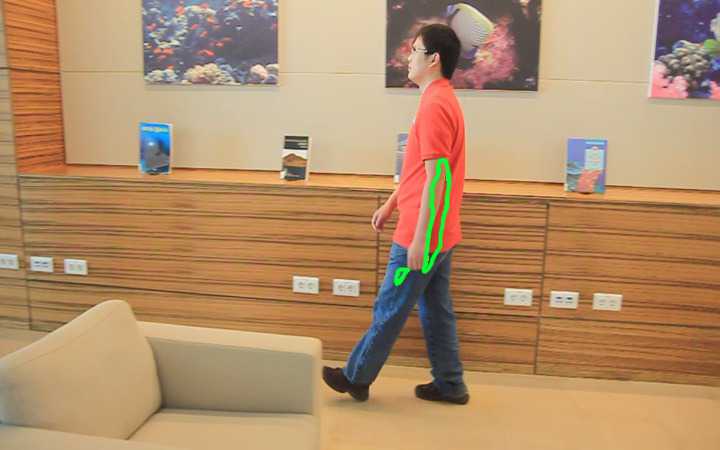}
  \\
  \includegraphics[clip,trim=264 0 171 0, totalheight=0.965in]{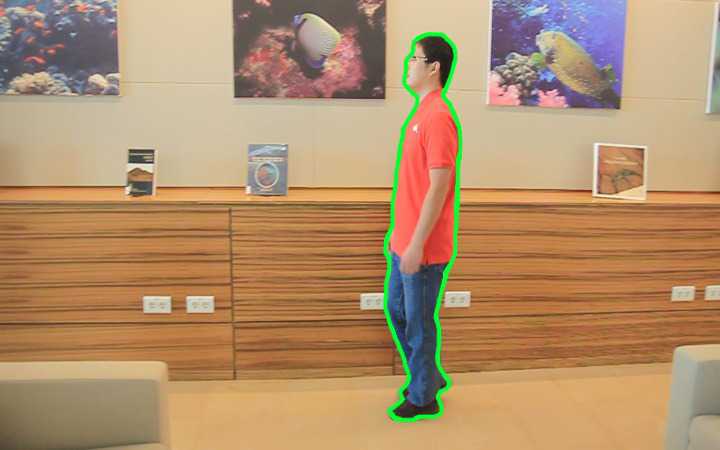}
  \includegraphics[clip,trim=264 0 171 0, totalheight=0.965in]{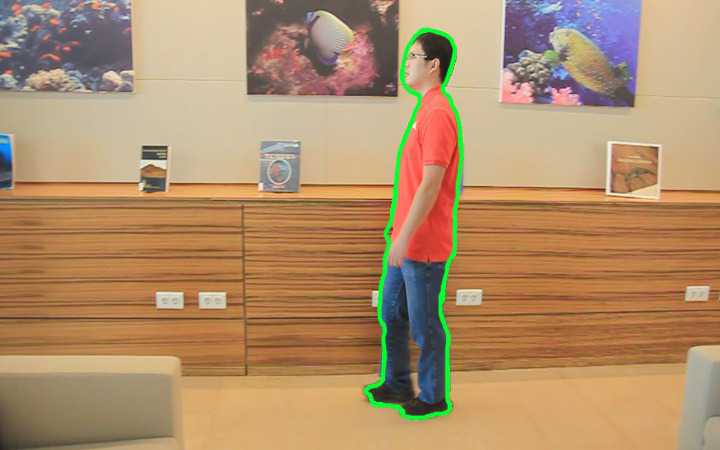}
  \includegraphics[clip,trim=264 0 171 0, totalheight=0.965in]{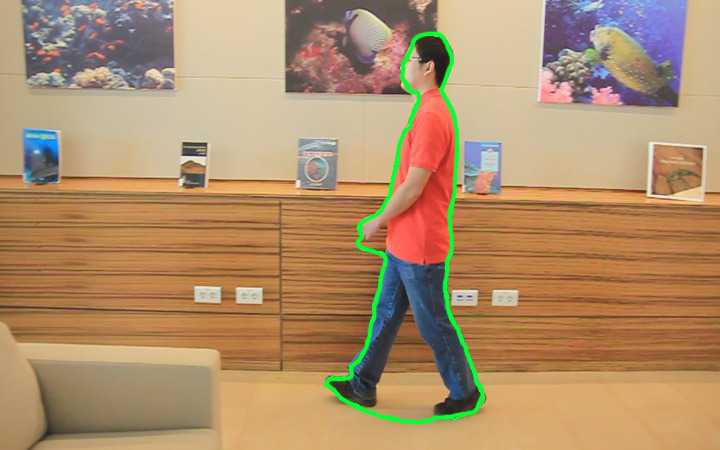}
  \includegraphics[clip,trim=264 0 171 0, totalheight=0.965in]{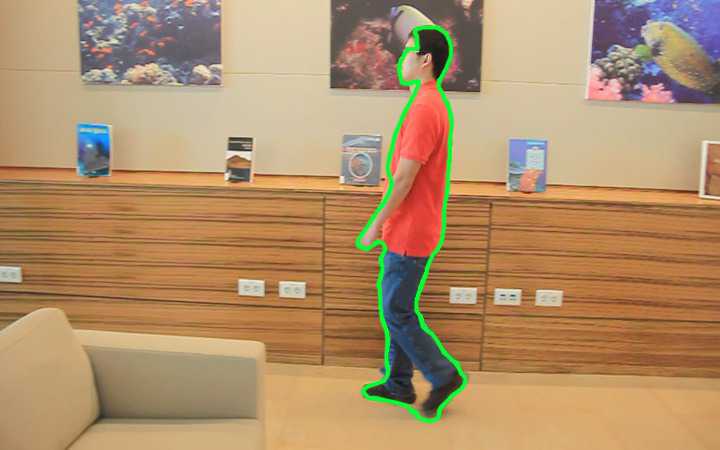}
  \includegraphics[clip,trim=264 0 171 0,totalheight=0.965in]{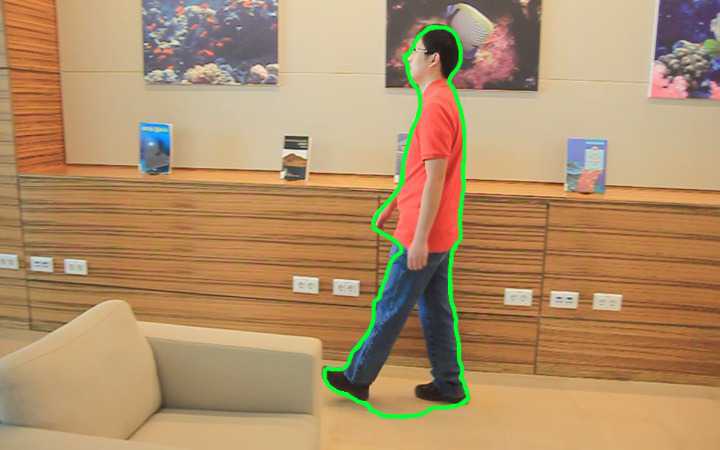}
  \\ 
  \includegraphics[clip,trim=264 0 171 0, totalheight=0.965in]{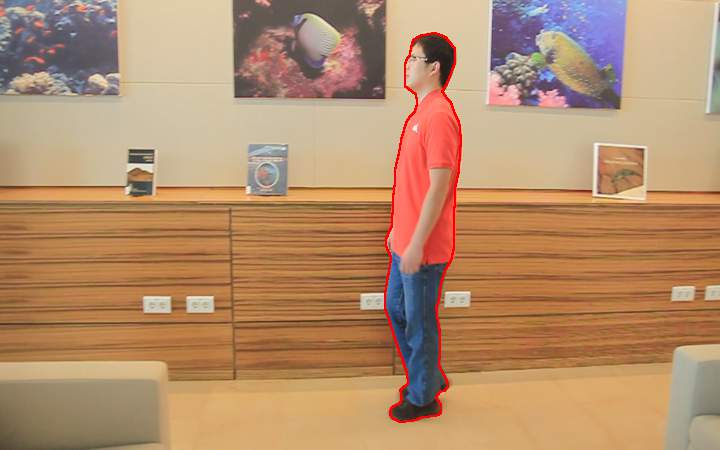}
  \includegraphics[clip,trim=264 0 171 0, totalheight=0.965in]{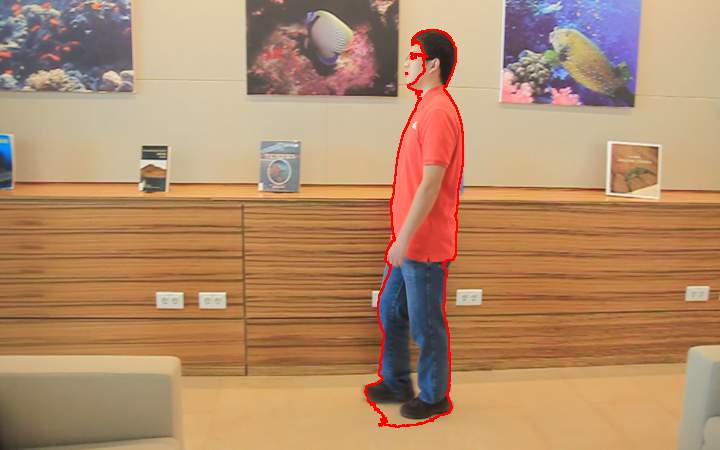}
  \includegraphics[clip,trim=264 0 171 0, totalheight=0.965in]{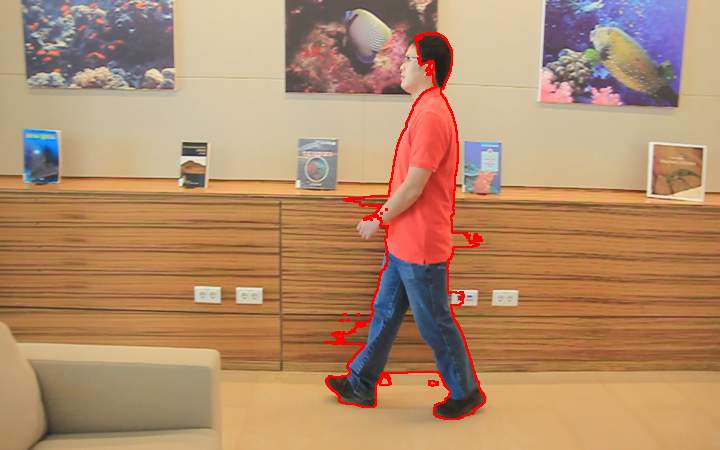}
  \includegraphics[clip,trim=264 0 171 0, totalheight=0.965in]{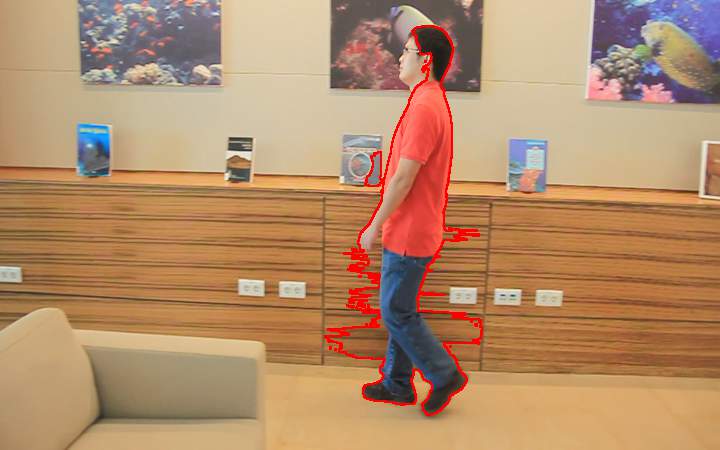}
  \includegraphics[clip,trim=264 0 171 0, totalheight=0.965in]{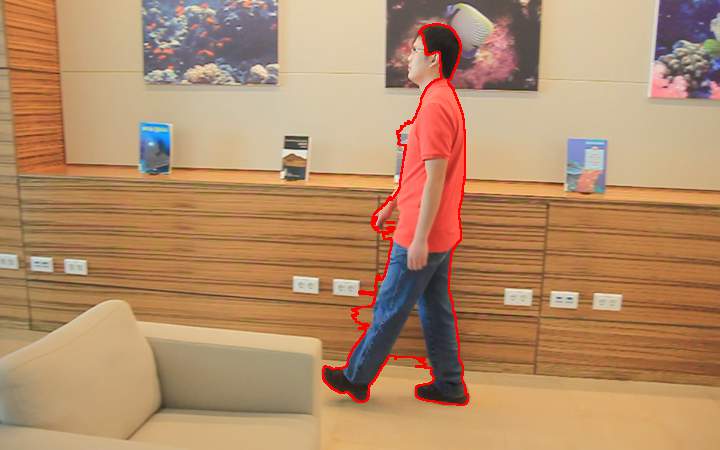}
  \\
  \includegraphics[clip,trim=195 0 127 0, totalheight=0.965in]{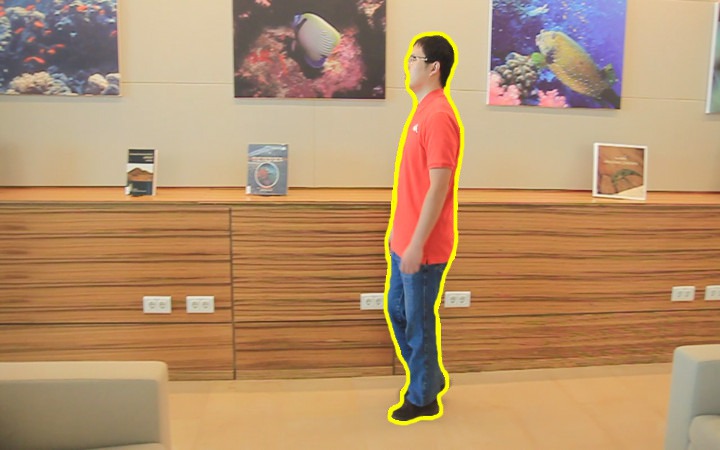}
  \includegraphics[clip,trim=195 0 127 0, totalheight=0.965in]{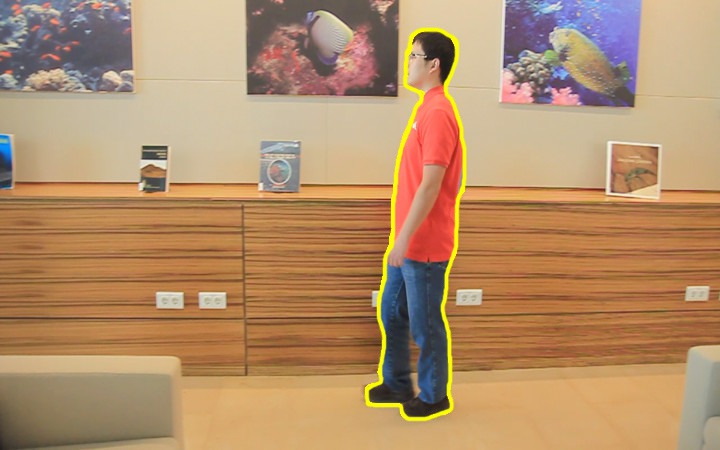}
  \includegraphics[clip,trim=195 0 127 0, totalheight=0.965in]{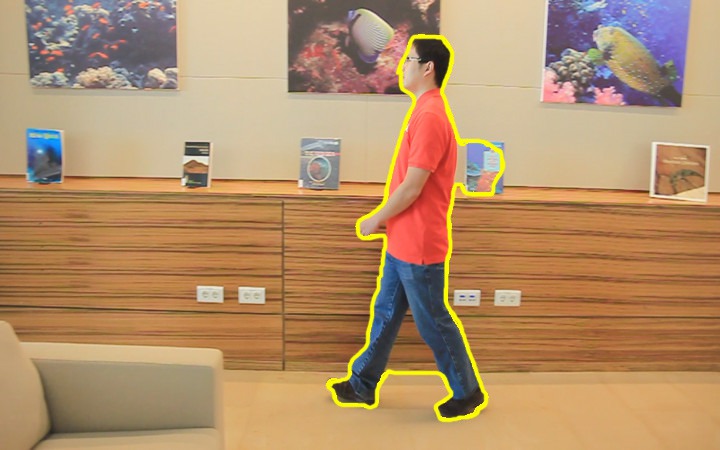}
  \includegraphics[clip,trim=195 0 127 0, totalheight=0.965in]{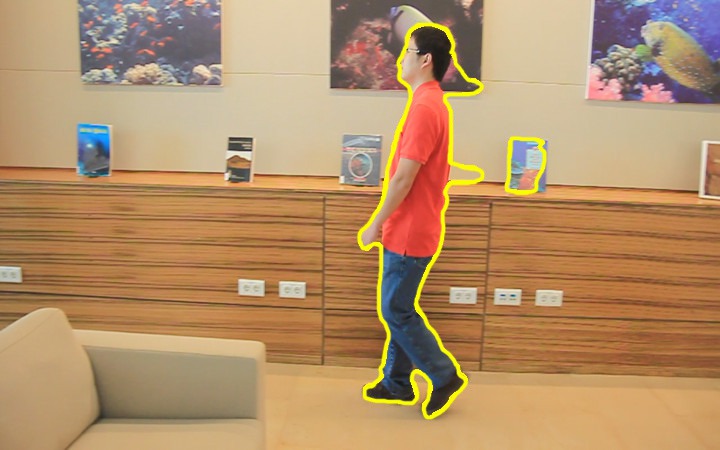}
  \includegraphics[clip,trim=195 0 127 0, totalheight=0.965in]{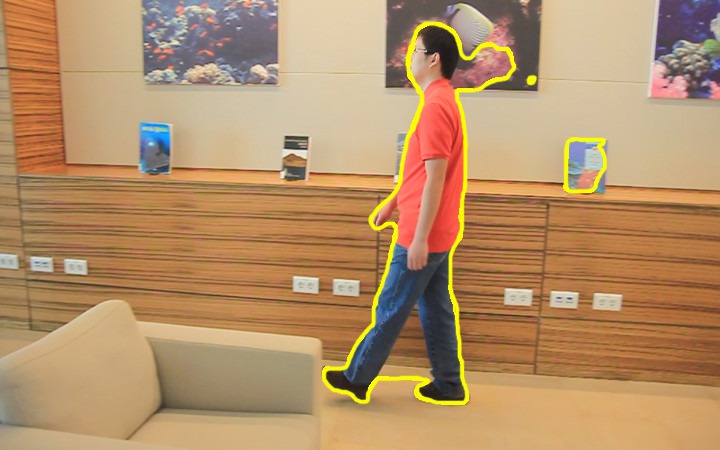}\\
  \includegraphics[clip,trim=264 0 171 0, totalheight=0.965in]{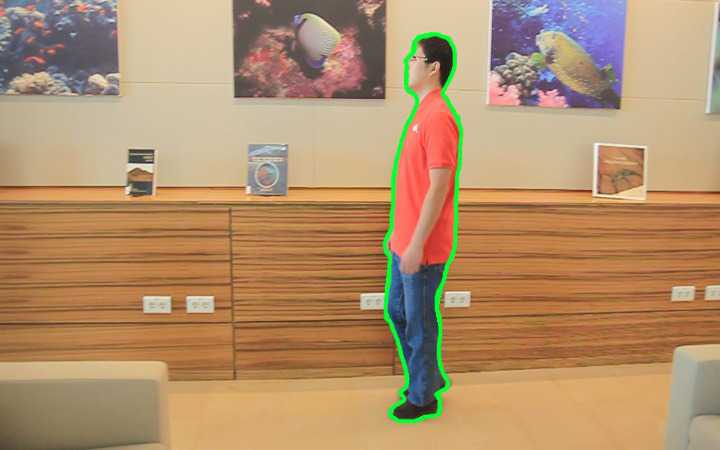}
  \includegraphics[clip,trim=264 0 171 0, totalheight=0.965in]{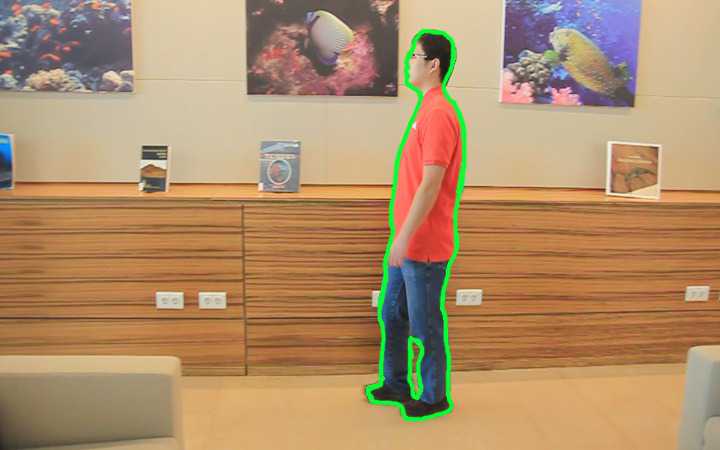}
  \includegraphics[clip,trim=264 0 171 0, totalheight=0.965in]{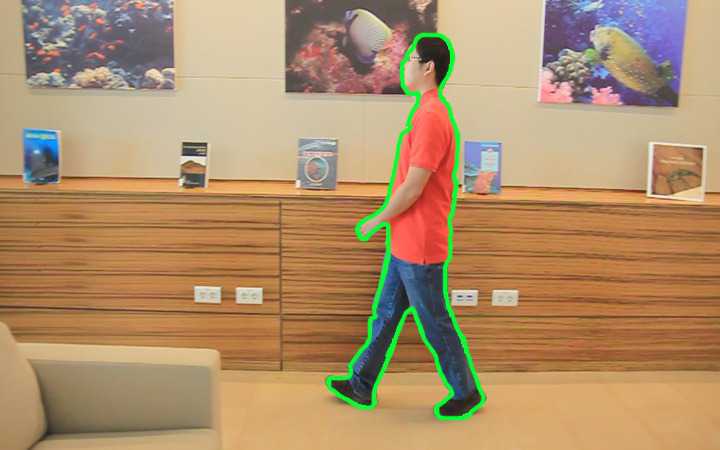}
  \includegraphics[clip,trim=264 0 171 0, totalheight=0.965in]{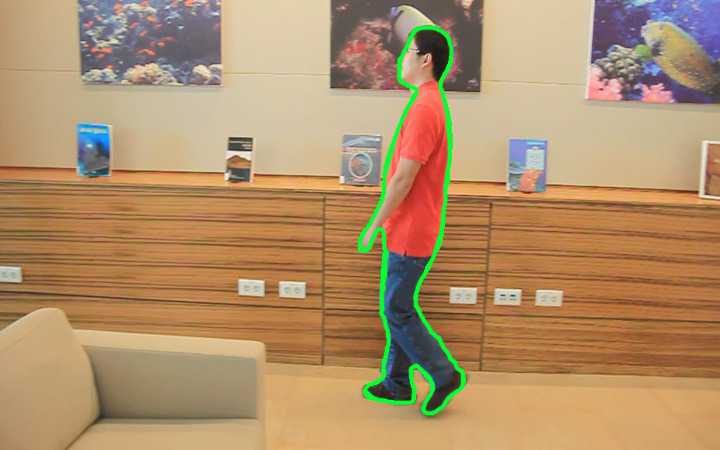}
  \includegraphics[clip,trim=264 0 171 0, totalheight=0.965in]{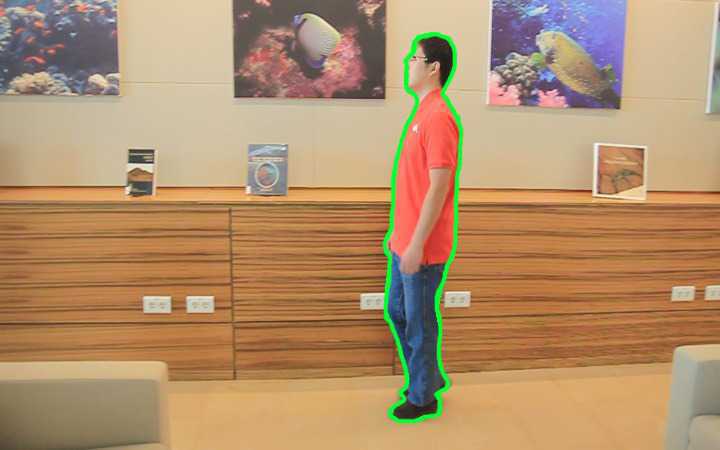}
  \caption{{\bf Modeling Occlusions/Dis-Occlusions is Necessary.}
    [$1^{\text{st}}$ row]: occlusion/dis-occlusion detection are
    turned off in our method. [$2^{\text{nd}}$]: occlusion
    modeling done, but not dis-occlusions in our
    method. [$3^{\text{rd}}$]: dis-occlusions detected but not
    occlusions.  [$4^{\text{th}}$]: result of Scribbles.
    [$5^{\text{th}}$]: result of AAE.
    [$6^{\text{th}}$]: accurate tracking when both
    occlusion and dis-occlusion modeling is performed (our final
    result).}
  \label{fig:expt_library}
\end{figure}

The first experiment (Fig.~\ref{fig:expt_library}) shows that
occlusion and dis-occlusion modeling is vital. As the man in the
sequence walks forward, his legs, arms and back are
self-occluded/disoccluded.  Ignoring occlusions (setting
$\tilde O_{\tau}=\emptyset$ in
Section~\ref{sec:final_optimization_warp_occlusion}) and dis-occlusion
detection, the shape is inaccurate (first row). Using occlusion
modeling but not dis-occlusions (second row), it is possible to
discard the portion of the background between the legs, and the
occluded right hand in the first frame is removed. Using the
dis-occlusion modeling but not occlusions (third row), disoccluded
parts of the body are detected. However, irrelevant regions of the
background (that can be removed in the occlusion stage) are
captured. Best results (last row) are achieved when both the occlusion
and dis-occlusions are modeled. The fourth row shows the result of
Scribbles, which has trouble discriminating between face and
the background, which share similar radiance.  The fifth row shows the
result of Adobe After Effects 2013 (AAE), which captures irrelevant
background.

\def\fPathLibST{figures2/new_expts/lib}

Fig.~\ref{fig:expt_fish} shows tracking of a fish and a skater. When
foreground/ background global histograms are easily separable,
Scribbles performs well, and when occlusions are minor AAE, performs
well as does the proposed method.

\def\fPathFishST{figures2/new_expts/fish}
\def\fishH{0.79in}
\def\skateH{0.79in}
\begin{figure}
  \centering
  \includegraphics[clip,trim=300 120 110 40, width=\fishH]{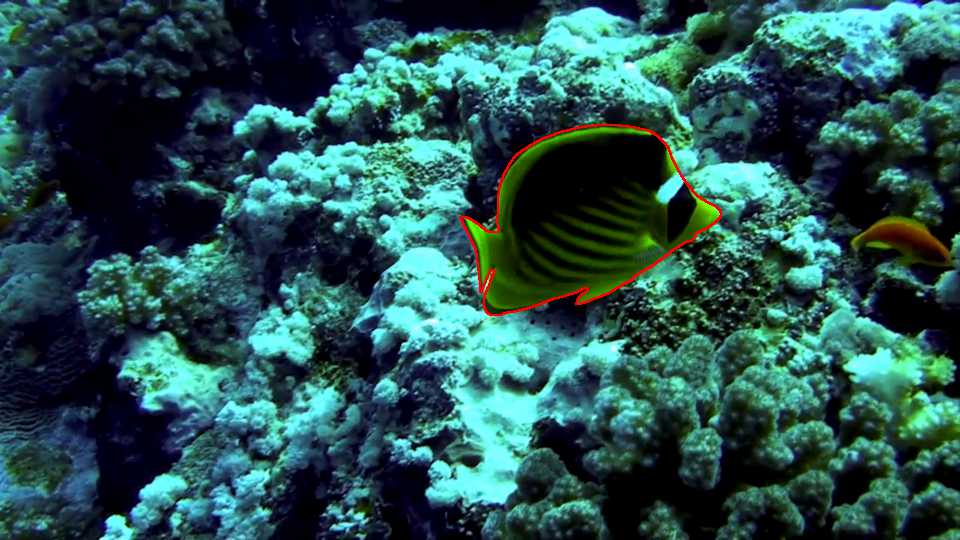}
  \includegraphics[clip,trim=300 120 110 40, width=\fishH]{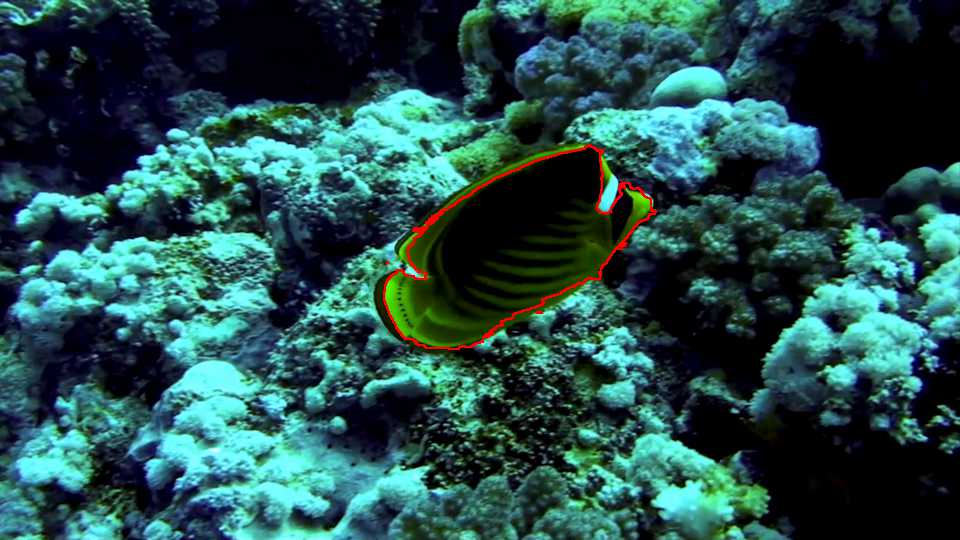}
  \includegraphics[clip,trim=300 120 110 40, width=\fishH]{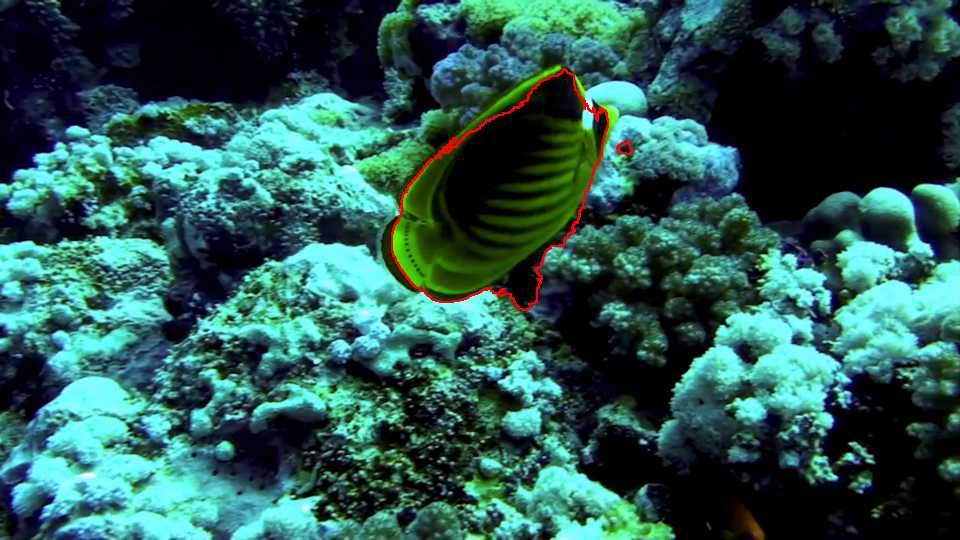}
  \includegraphics[clip,trim=300 120 110 40, width=\fishH]{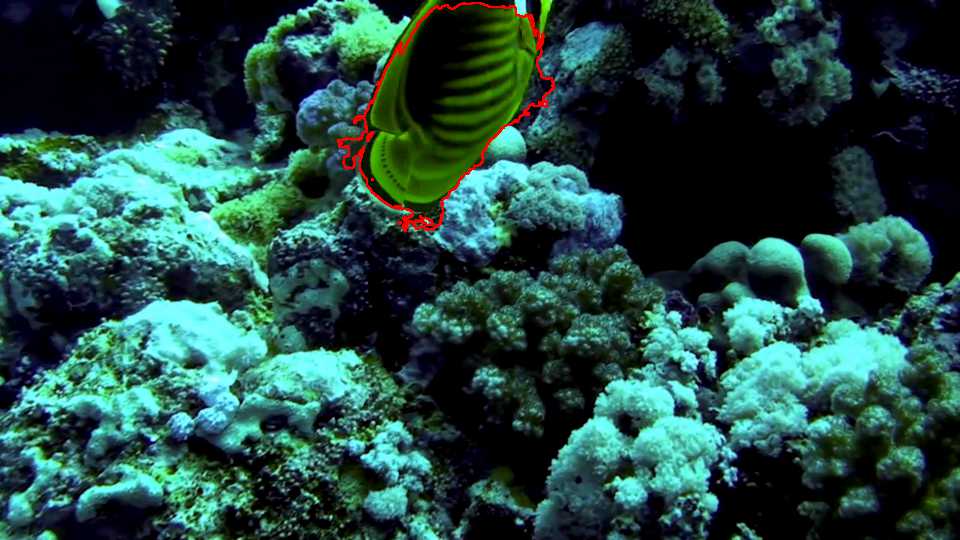}\\
  \includegraphics[clip,trim=225 90 82.5 30, width=\fishH]{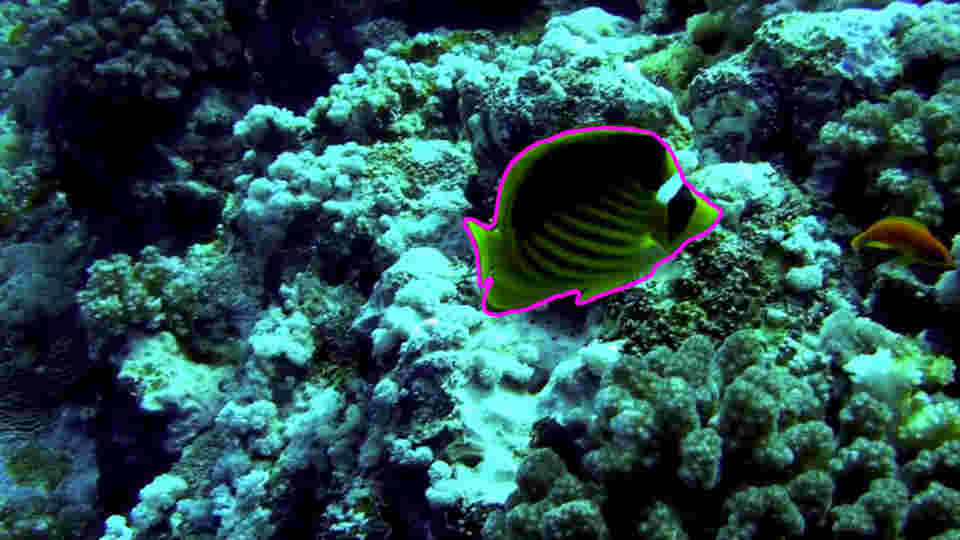}
  \includegraphics[clip,trim=225 90 82.5 30, width=\fishH]{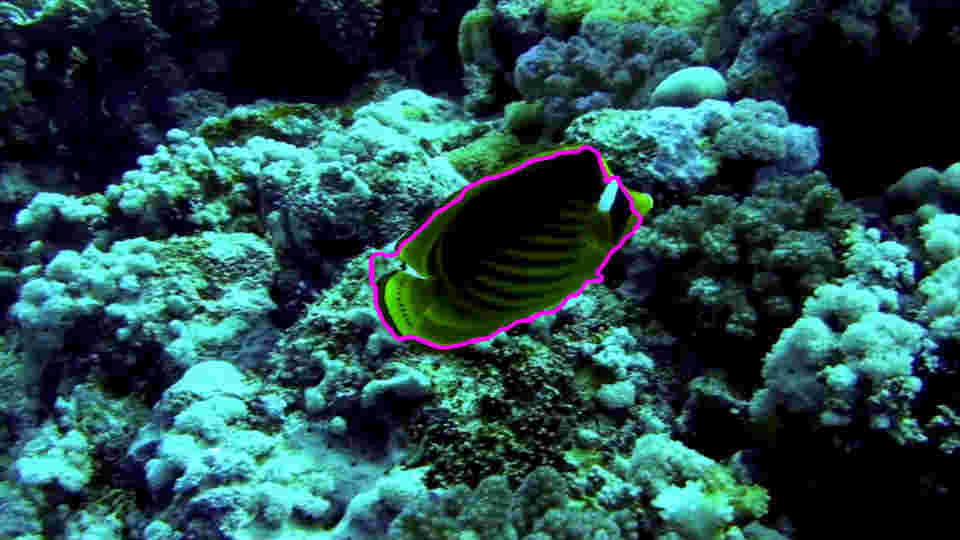}
  \includegraphics[clip,trim=225 90 82.5 30, width=\fishH]{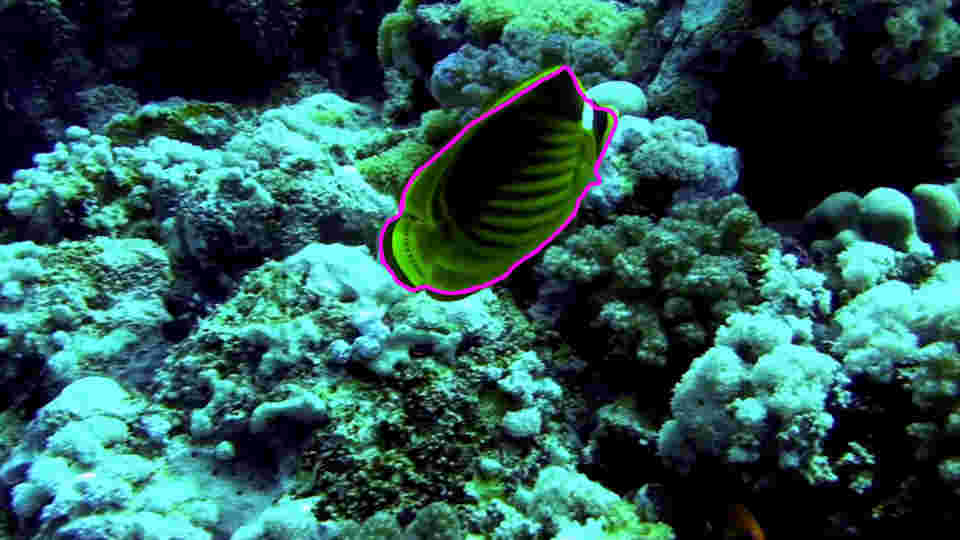}
  \includegraphics[clip,trim=225 90 82.5 30, width=\fishH]{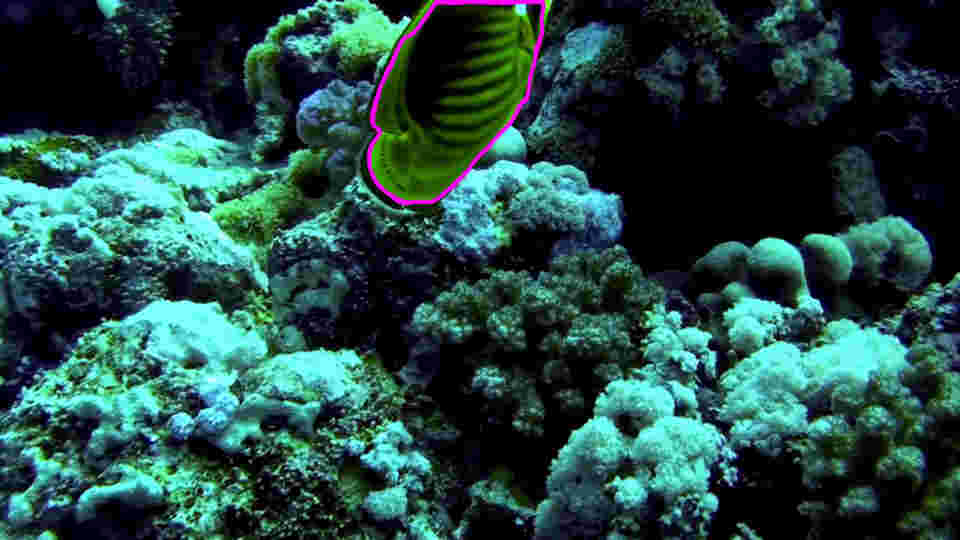}\\
  \includegraphics[clip,trim=300 120 110 40, width=\fishH]{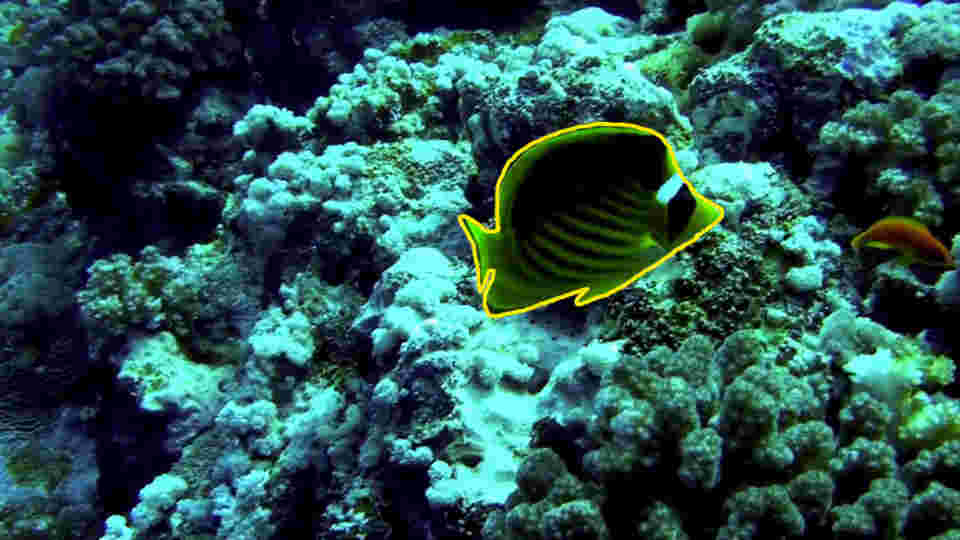}
  \includegraphics[clip,trim=300 120 110 40, width=\fishH]{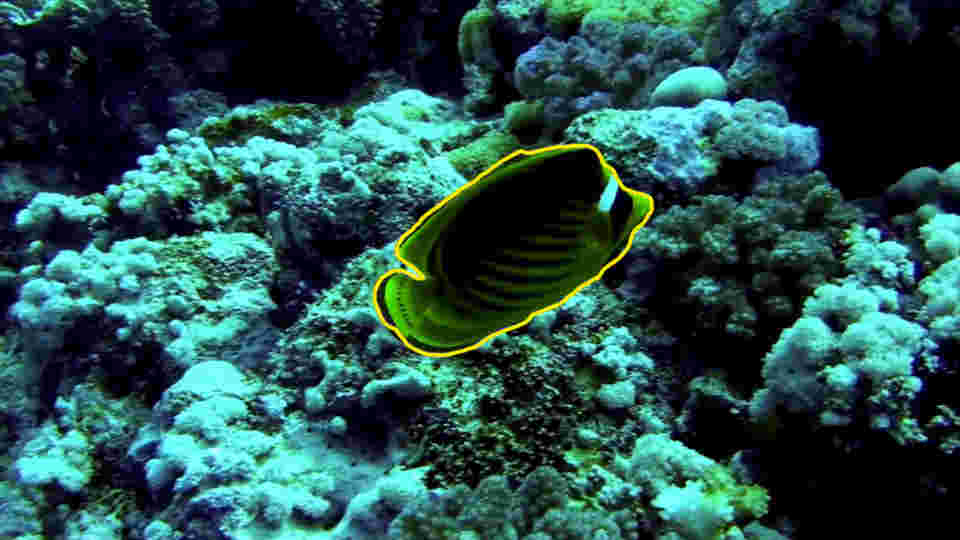}
  \includegraphics[clip,trim=300 120 110 40, width=\fishH]{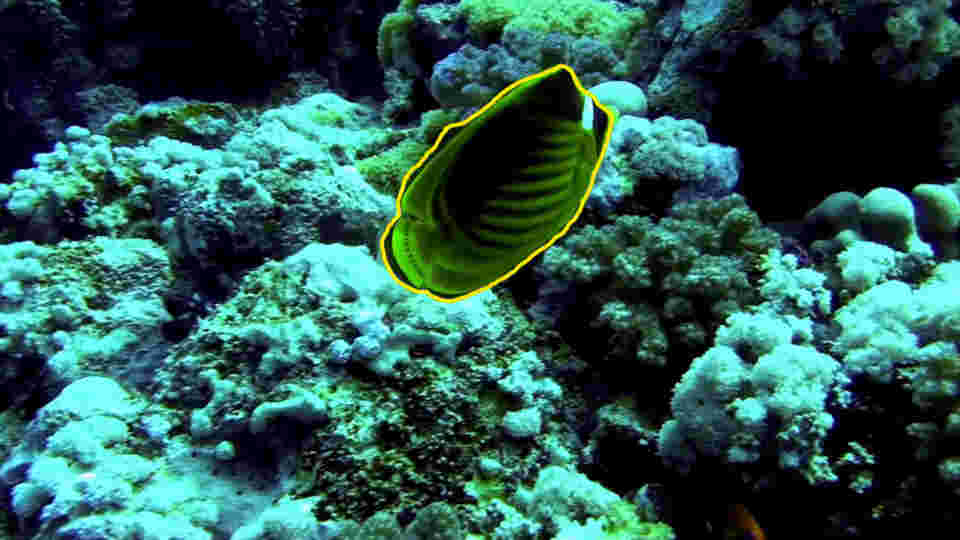}
  \includegraphics[clip,trim=300 120 110 40,
  width=\fishH]{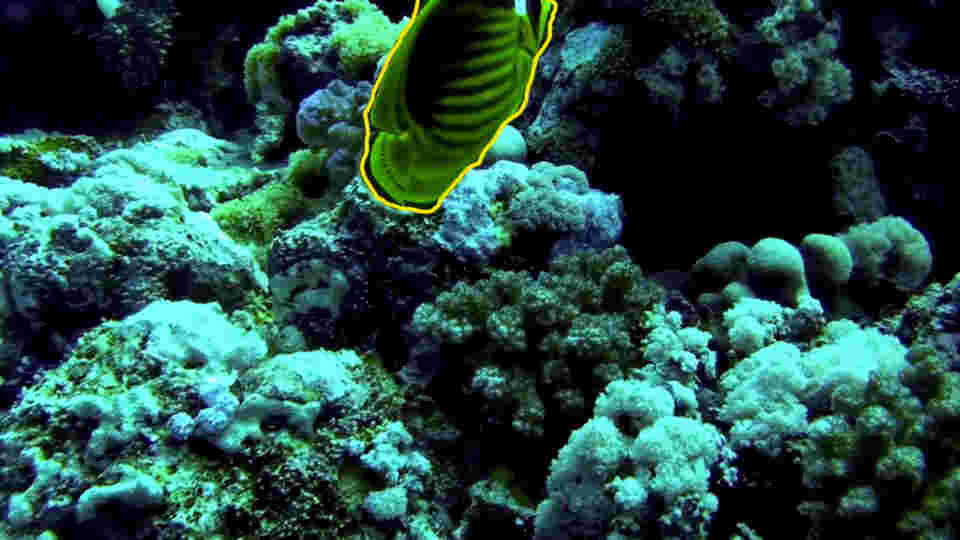}\\\vspace{0.05in}
  \includegraphics[clip,trim=167 63 125 63, width=\skateH]{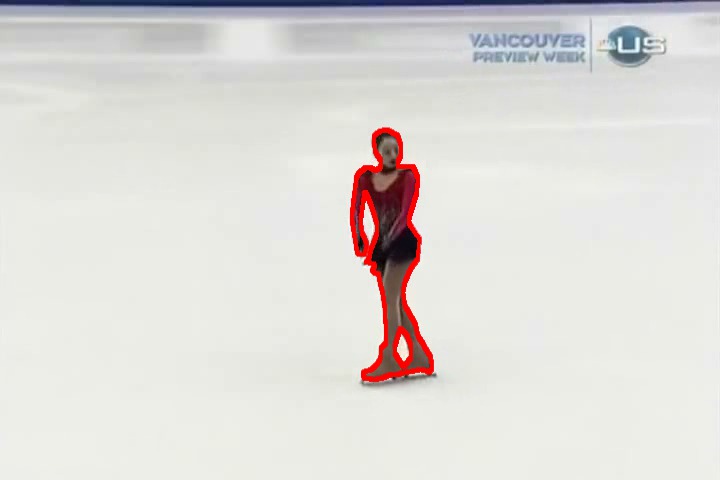}
  \includegraphics[clip,trim=167 63 125 63, width=\skateH]{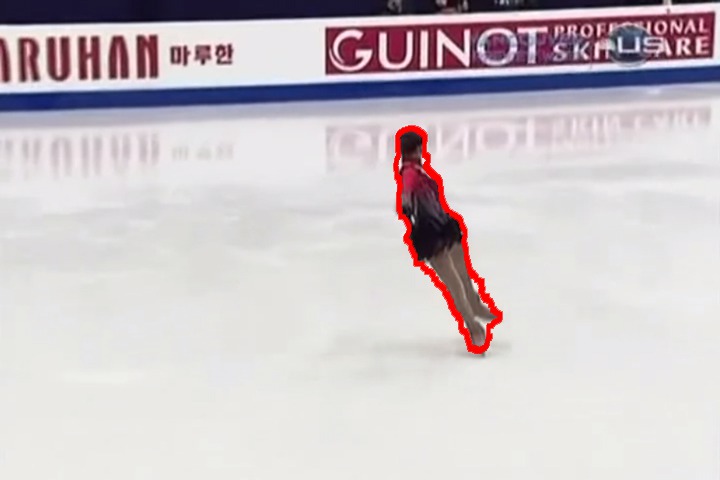}
  \includegraphics[clip,trim=167 63 125 63, width=\skateH]{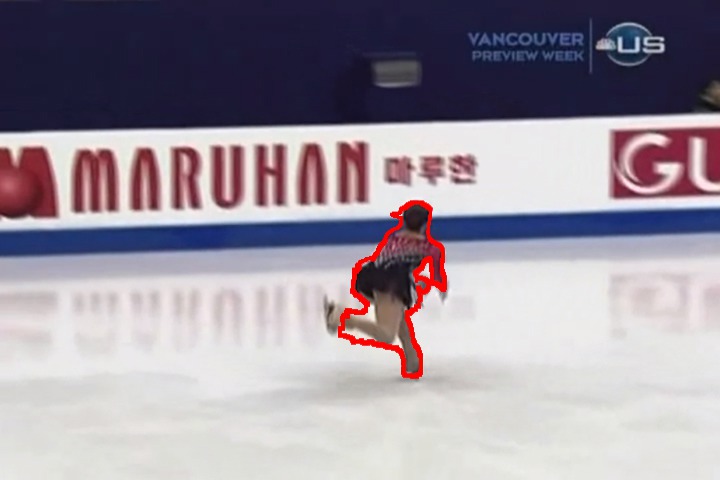}
  \includegraphics[clip,trim=167 63 125 63, width=\skateH]{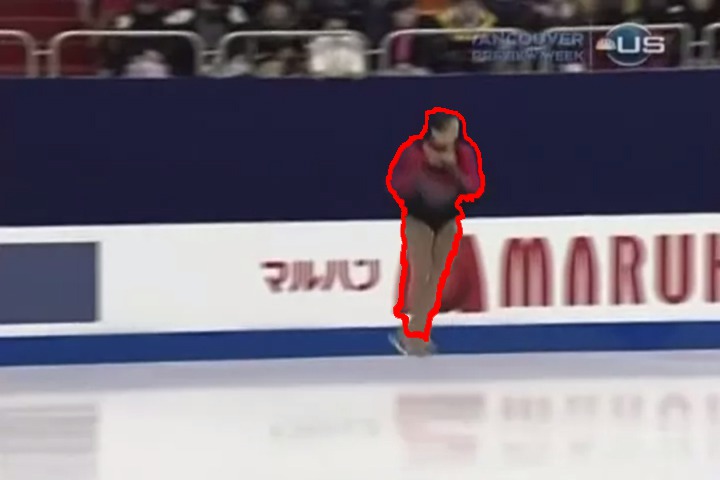}\\
  \includegraphics[clip,trim=167 63 125 63, width=\skateH]{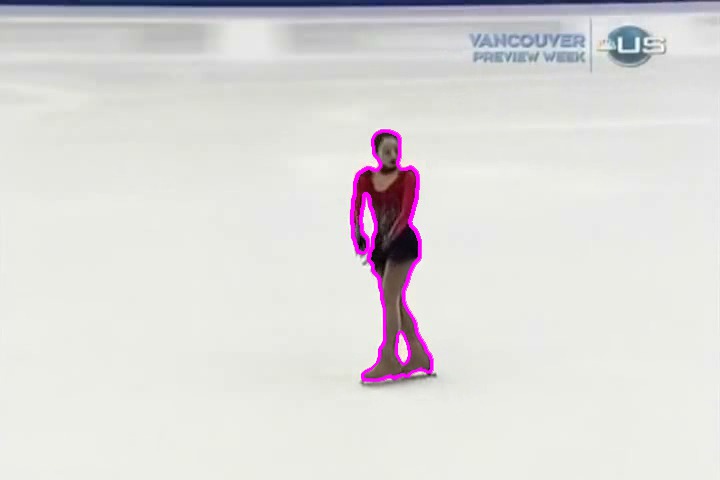}
  \includegraphics[clip,trim=167 63 125 63, width=\skateH]{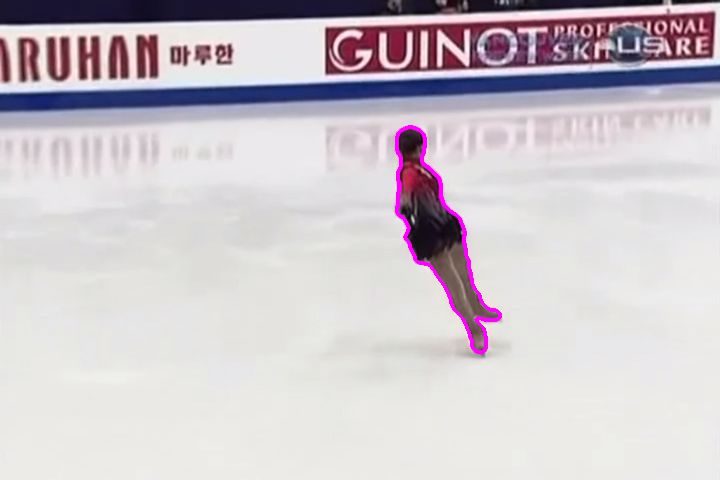}
  \includegraphics[clip,trim=167 63 125 63, width=\skateH]{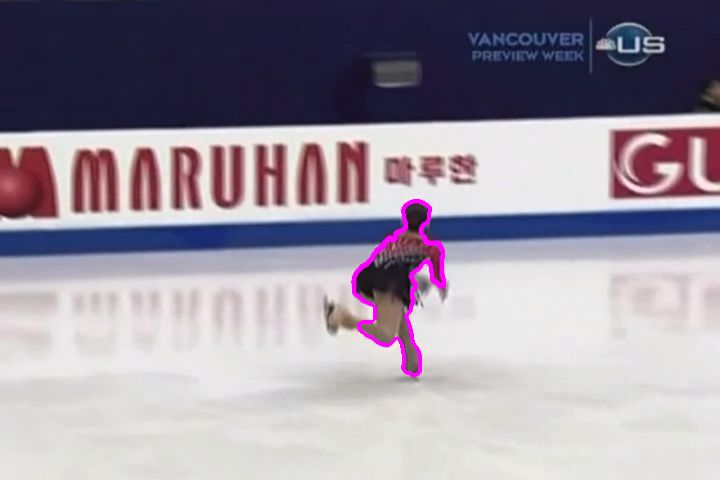}
  \includegraphics[clip,trim=167 63 125 63, width=\skateH]{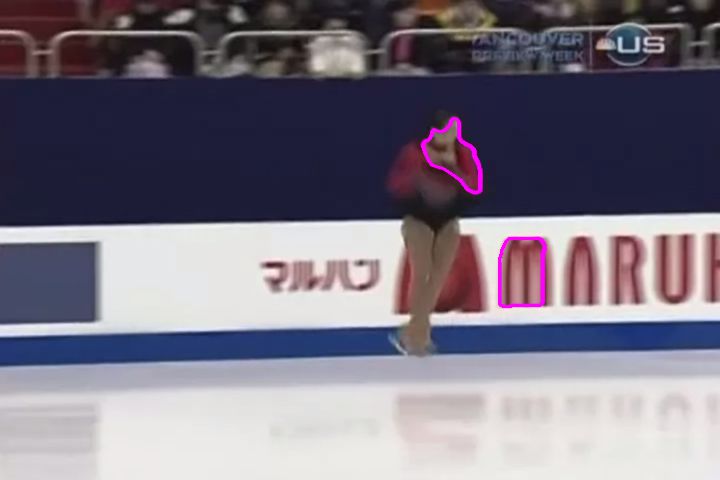}\\
  \includegraphics[clip,trim=167 63 125 63, width=\skateH]{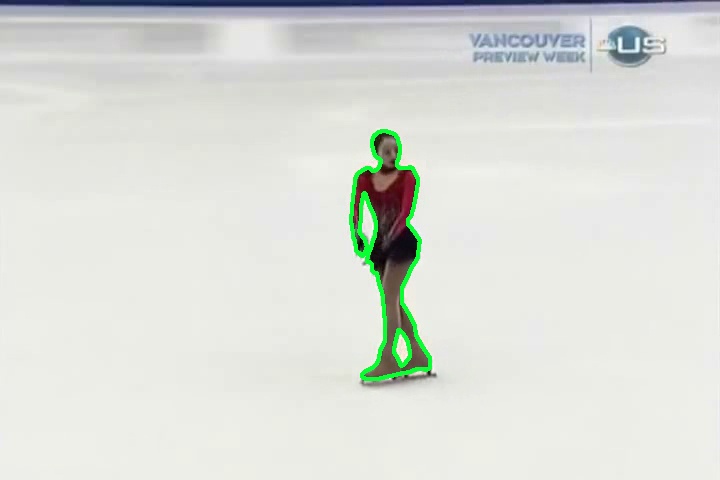}
  \includegraphics[clip,trim=167 63 125 63, width=\skateH]{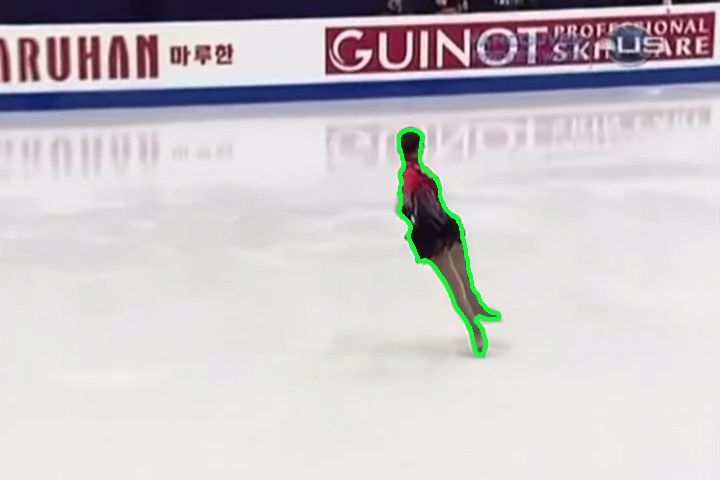}
  \includegraphics[clip,trim=167 63 125 63, width=\skateH]{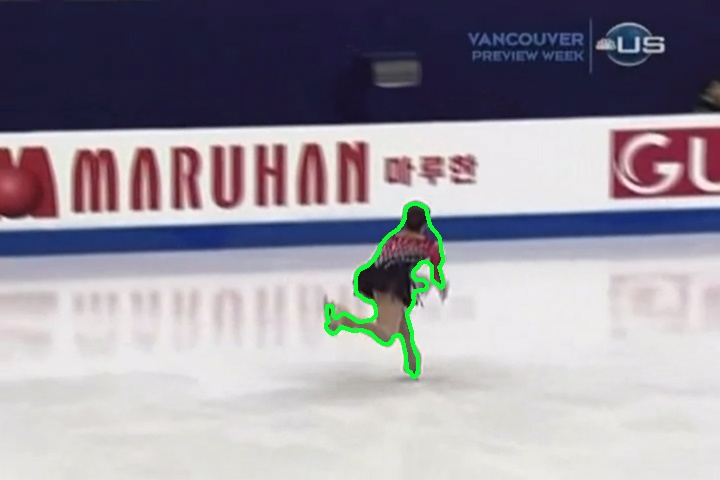}
  \includegraphics[clip,trim=167 63 125 63, width=\skateH]{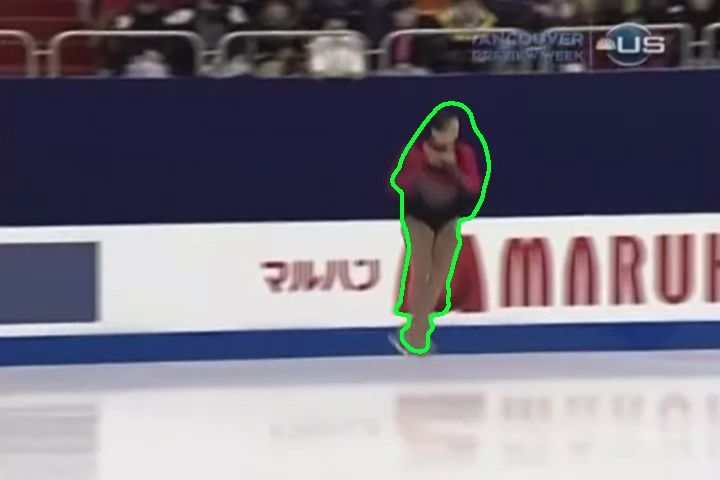}
  \caption{{\bf Distinctive foreground/background global statistics.}
    [Top]: \cite{Fan:PAMI12}, [Middle]: AAE, [Bottom]: proposed
    method. When fore/background global statistics are separable,
    \cite{Fan:PAMI12}, and AAE, for minor occlusions, performs well.}
  \label{fig:expt_fish}
\end{figure}

In Fig.~\ref{fig:expt_lady_mercedes}, we have tested our algorithm on
challenging video (more than 100 frames per sequence) exhibiting
self-occlusions and dis-occlusion (crossing legs, viewpoint change,
rotations in depth), complex object radiance and background in which
it becomes difficult to discriminate between foreground and background
global statistics (e.g., the woman's pants have same radiance as car
tires). Deviations from brightness constancy are clearly visible
(small illumination change, specular reflections, and even
shadows). The latter are handled with our dynamic radiance update. In
these sequences, Scribbles and Adobe After Effects 2013 (AAE) have
trouble discriminating between object and background which share
portions of similar intensity, and occlusions (e.g., crossing of
legs). In the ``Lady Mercedes,'' sequence (top left), after a few
frames, Scribbles can only track the head of the lady. This is because
the lady's clothing shares similar intensity as the tires of the car
and some of the background. Thus, the tracker confuses the clothing
with the background and only tracks the head, which has different
statistics from the rest of the images.  Our method is able to capture
the shape of the objects quite well (quantitative assessment is in
Table~\ref{fig:quantative_assesment}). The man at the station (top
right group) at the fourth column shows a limitation of our
dis-occlusion detection: dis-occluded parts of the object that do not
share similar radiance as the current template (sole of shoe) are not
detected. A variety of other videos are processed, and our method
performs quite well.

\def\fPathLadyST{figures2/new_expts/lady}
\def\fPathStationST{figures2/new_expts/station}
\def\fLadyH{0.77in}
\begin{figure*}
  \centering
  \includegraphics[clip,trim=170 0 20 13, width=\fLadyH]{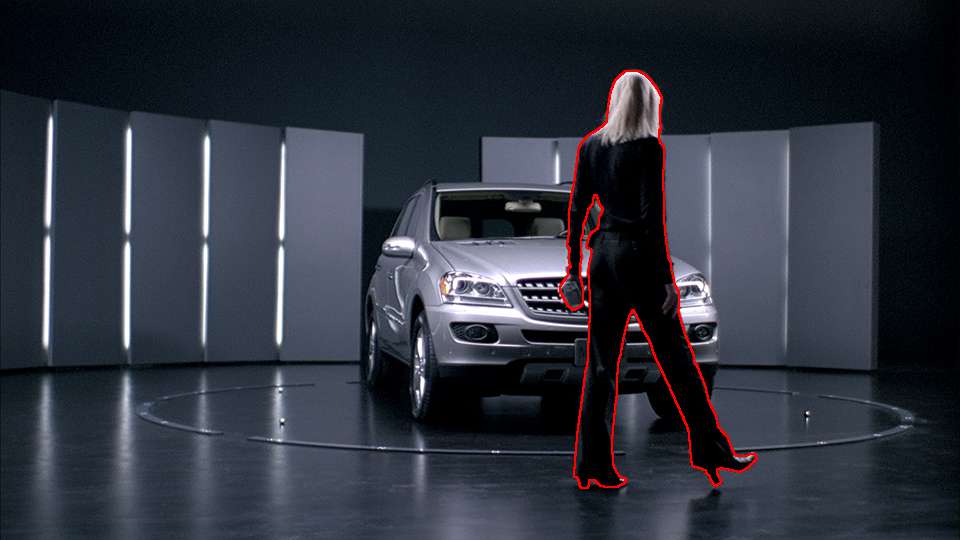}
  \includegraphics[clip,trim=170 0 20 13, width=\fLadyH]{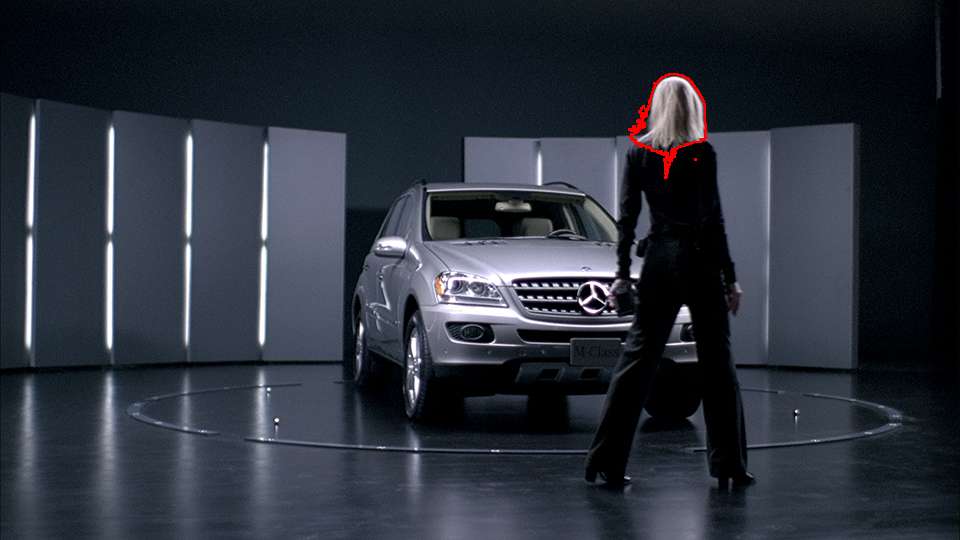}
  \includegraphics[clip,trim=170 0 20 13, width=\fLadyH]{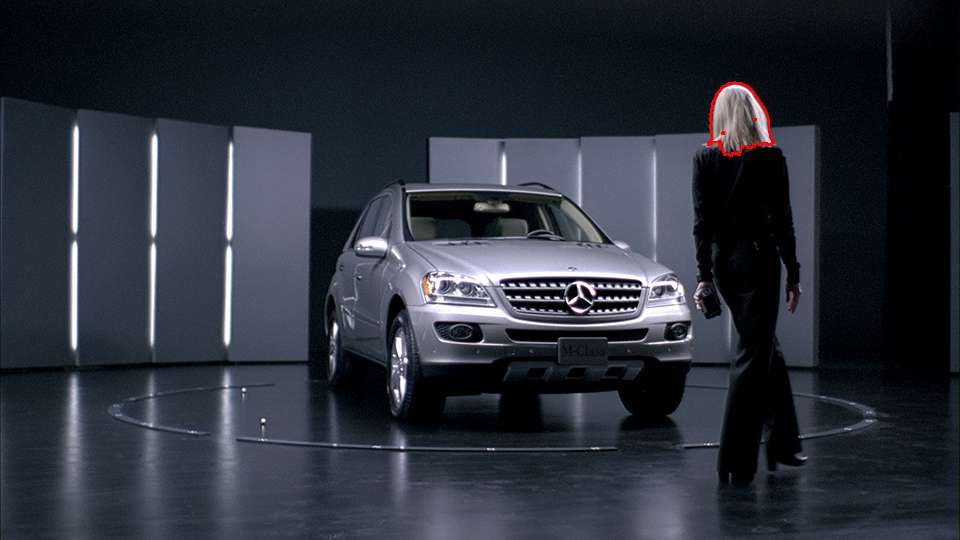}
  \includegraphics[clip,trim=170 0 20 13, width=\fLadyH]{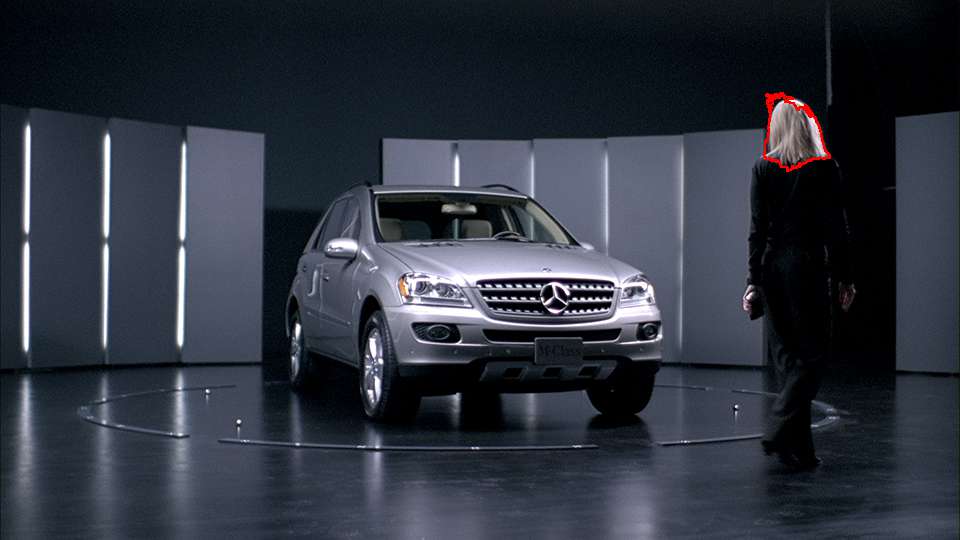}
  \includegraphics[clip,trim=100 50 0 70, width=\fLadyH]{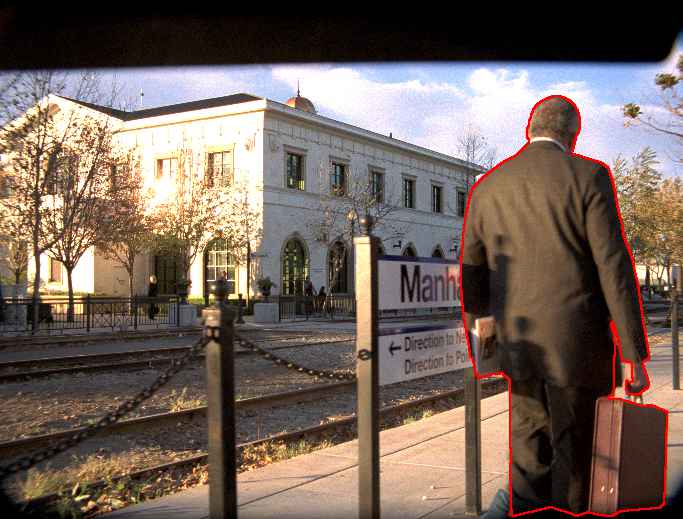}
  \includegraphics[clip,trim=100 50 0 70, width=\fLadyH]{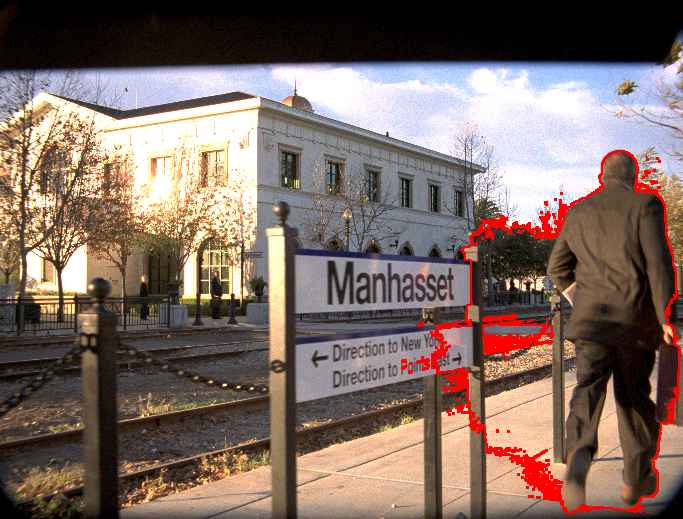}
  \includegraphics[clip,trim=100 50 0 70, width=\fLadyH]{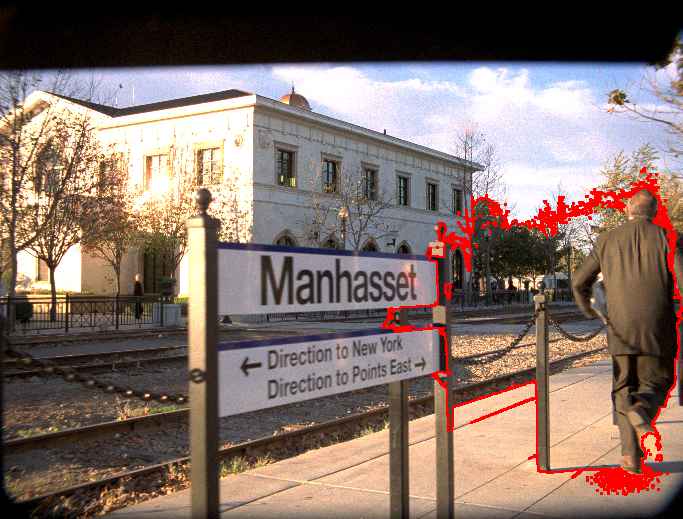}
  \includegraphics[clip,trim=100 50 0 70, width=\fLadyH]{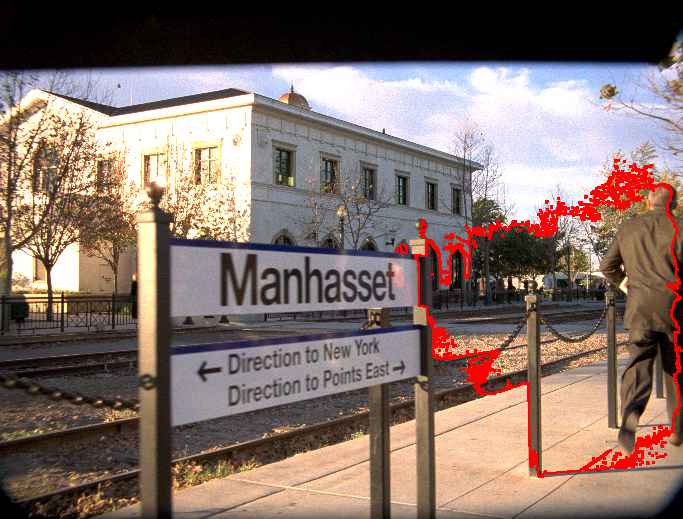}\\
  \includegraphics[clip,trim=170 0 20 13, width=\fLadyH]{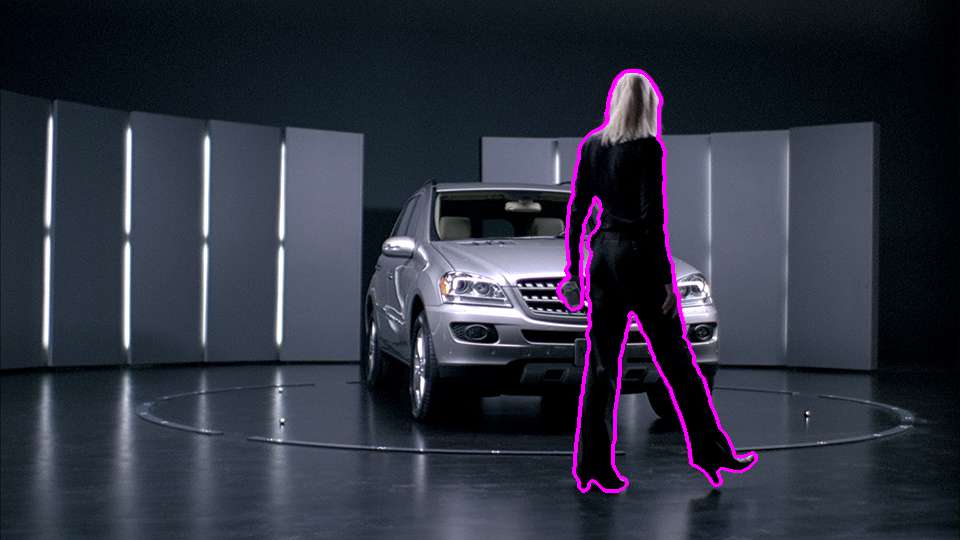}
  \includegraphics[clip,trim=170 0 20 13, width=\fLadyH]{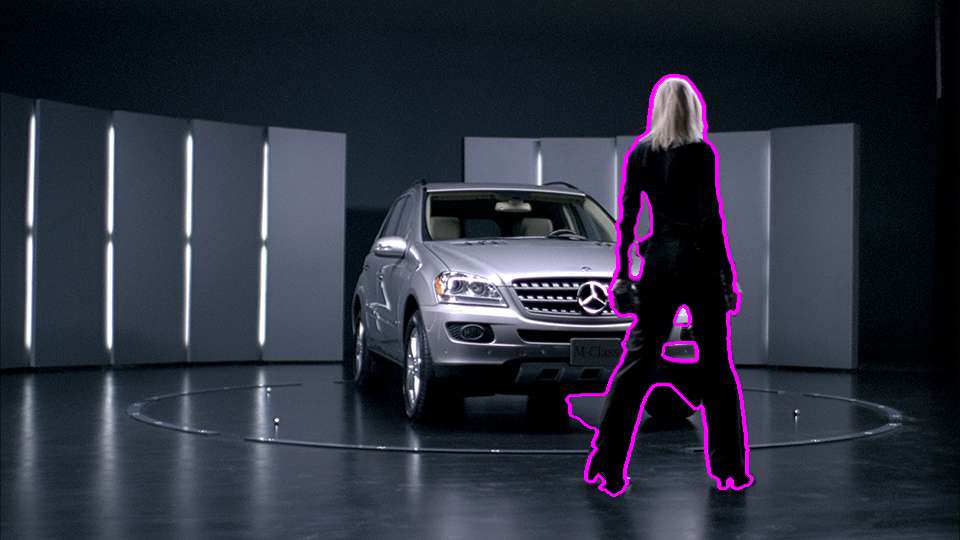}
  \includegraphics[clip,trim=170 0 20 13, width=\fLadyH]{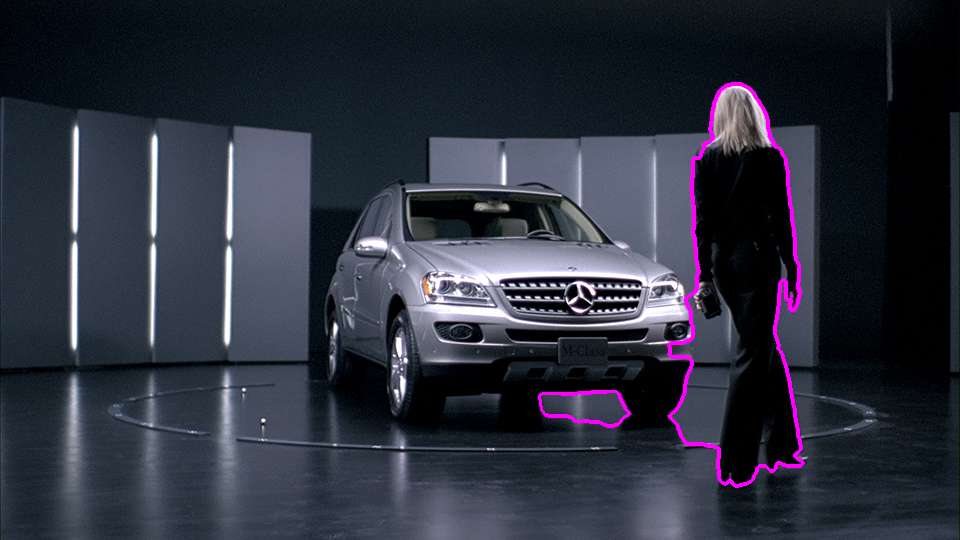}
  \includegraphics[clip,trim=170 0 20 13, width=\fLadyH]{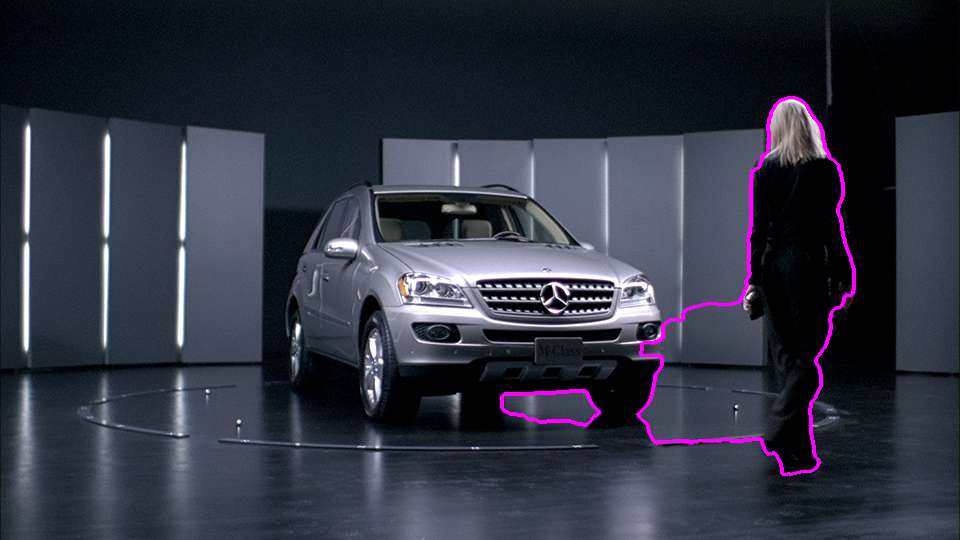}
  \includegraphics[clip,trim=75 38 0 52, width=\fLadyH]{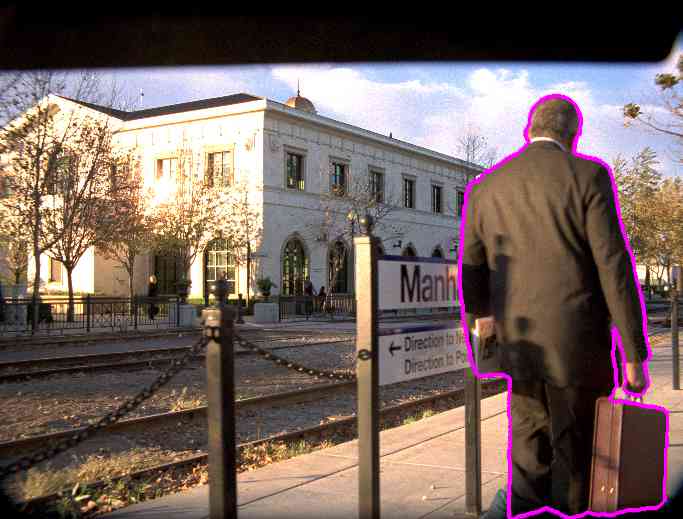}
  \includegraphics[clip,trim=75 38 0 52, width=\fLadyH]{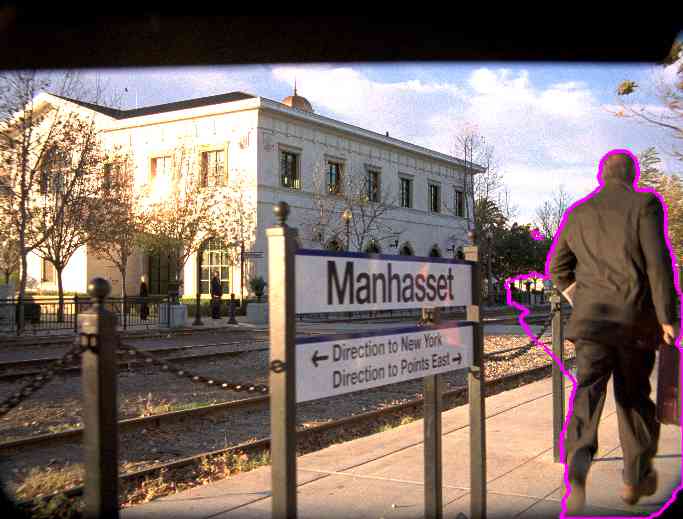}
  \includegraphics[clip,trim=75 38 0 52, width=\fLadyH]{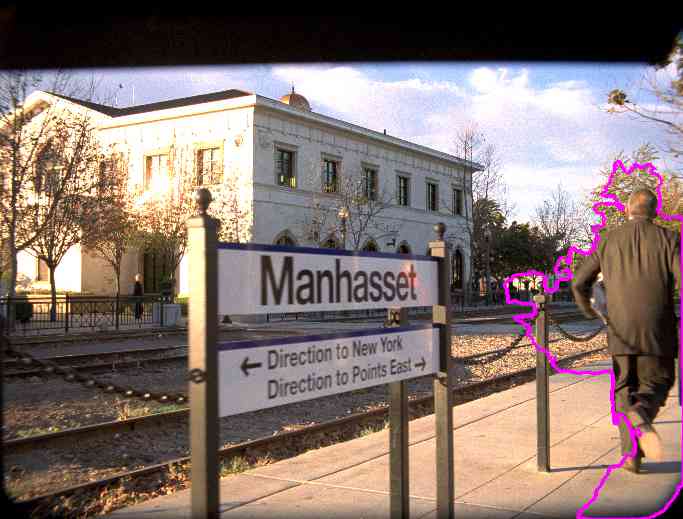}
  \includegraphics[clip,trim=75 38 0 52, width=\fLadyH]{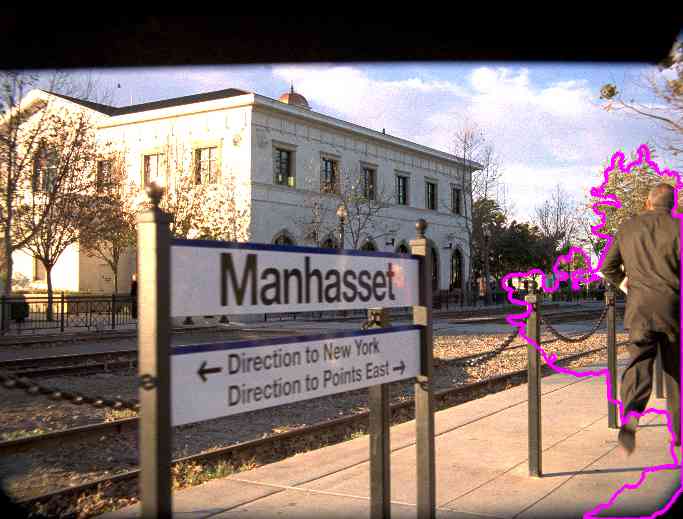}\\
  \includegraphics[clip,trim=170 0 20 13, width=\fLadyH]{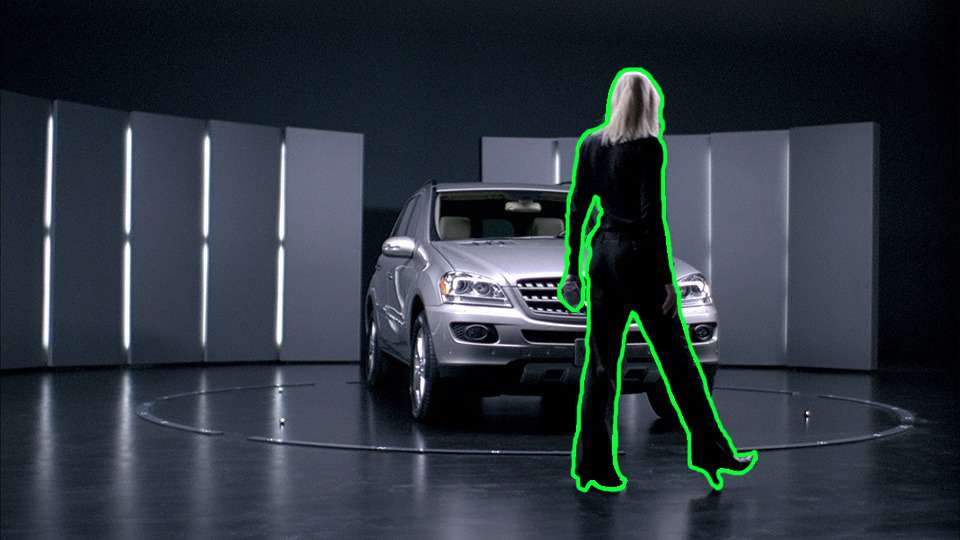}
  \includegraphics[clip,trim=170 0 20 13, width=\fLadyH]{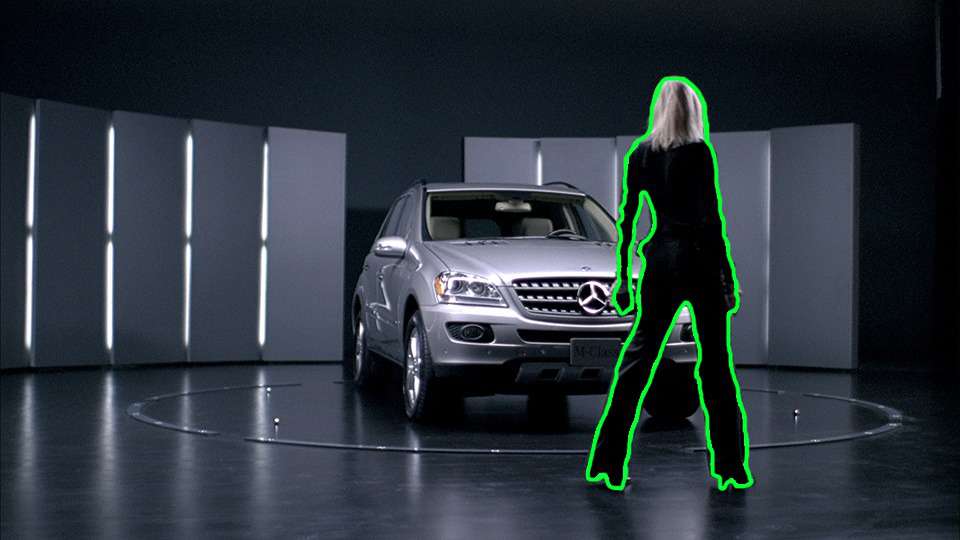}
  \includegraphics[clip,trim=170 0 20 13, width=\fLadyH]{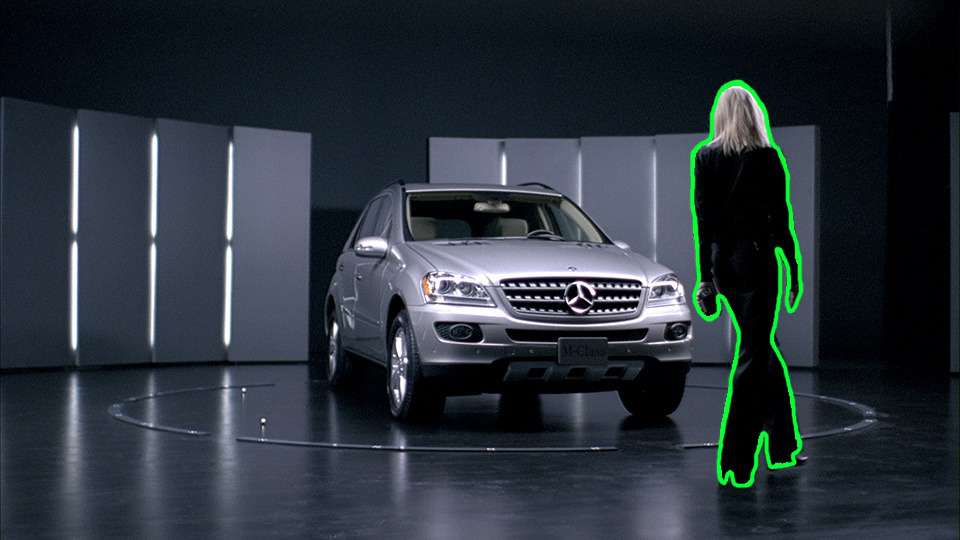}
  \includegraphics[clip,trim=170 0 20 13, width=\fLadyH]{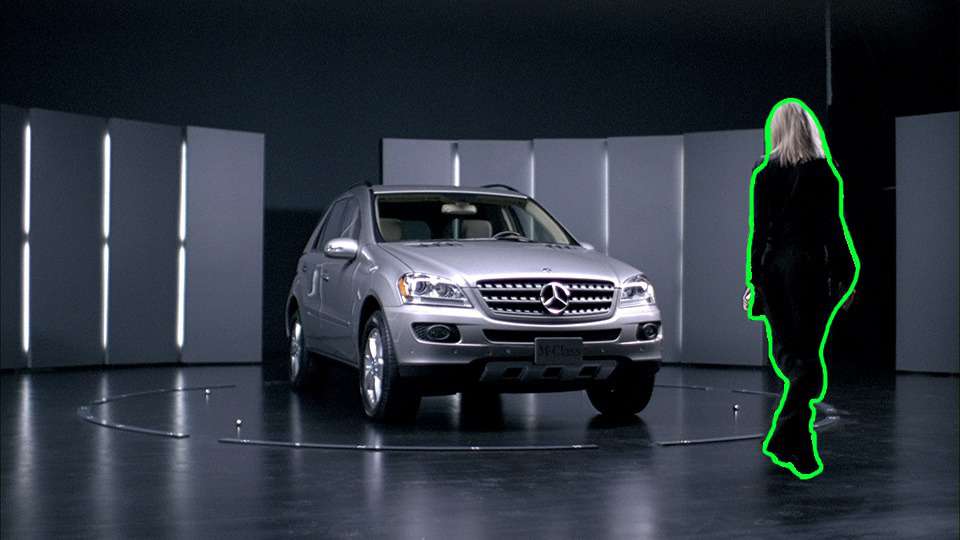}
  \includegraphics[clip,trim=100 50 0 70, width=\fLadyH]{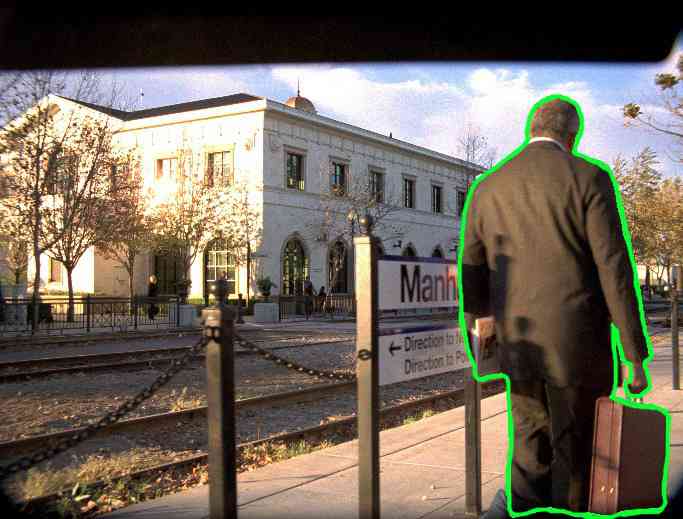}
  \includegraphics[clip,trim=100 50 0 70, width=\fLadyH]{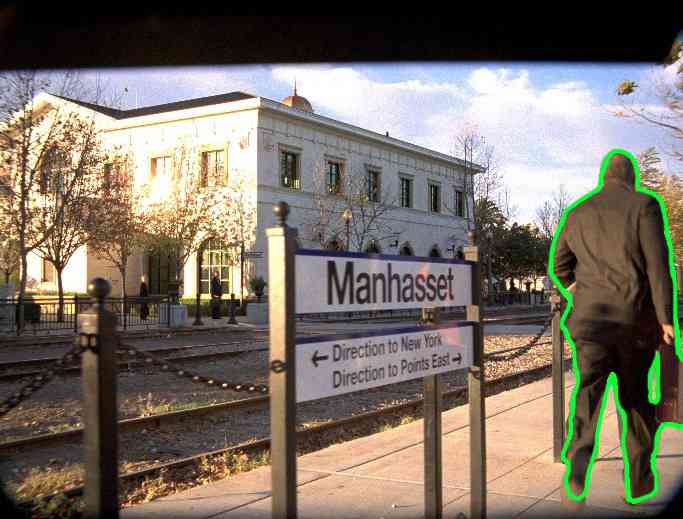}
  \includegraphics[clip,trim=100 50 0 70, width=\fLadyH]{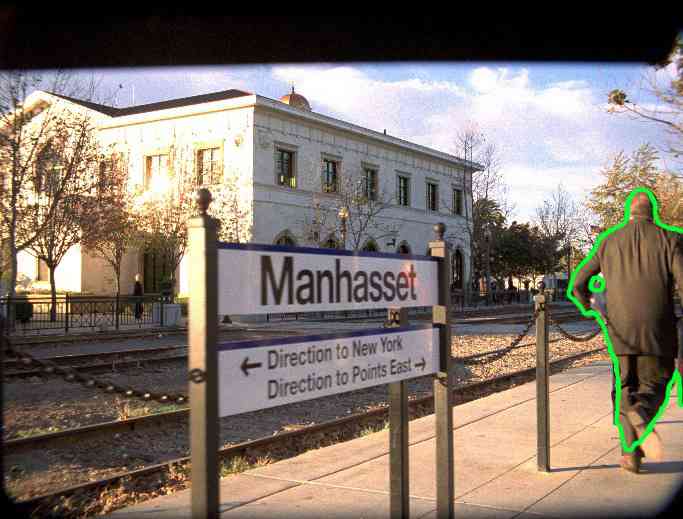}
  \includegraphics[clip,trim=100 50 0 70, width=\fLadyH]{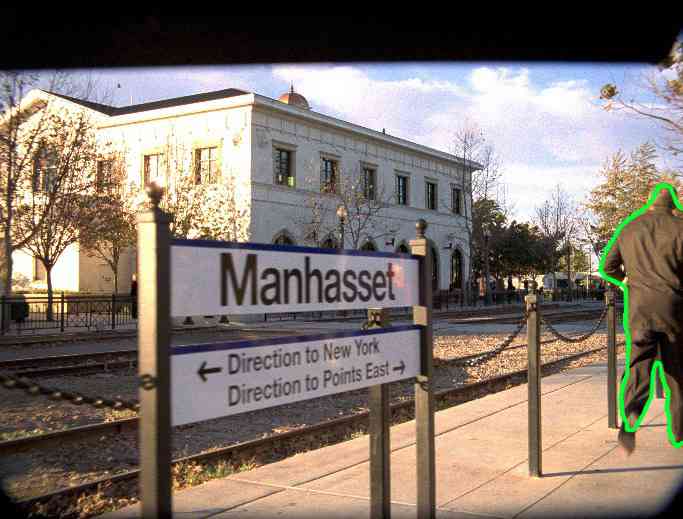}\\\vspace{0.05in}
  \def\hobHeight{0.77in}
  \def\marHeight{0.77in}
  \includegraphics[clip,trim=150 0 50 0, width=\hobHeight]{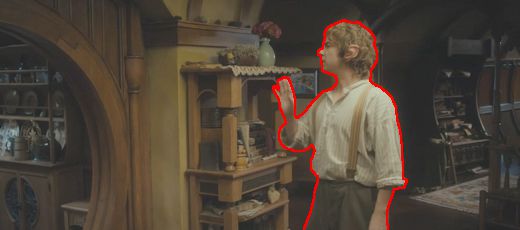}
  \includegraphics[clip,trim=150 0 50 0, width=\hobHeight]{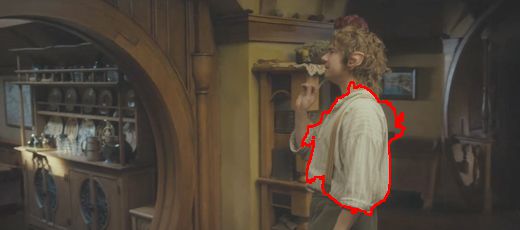}
  \includegraphics[clip,trim=150 0 50 0,width=\hobHeight]{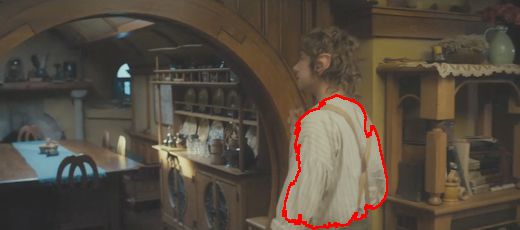}
  \includegraphics[clip,trim=150 0 50 0, width=\hobHeight]{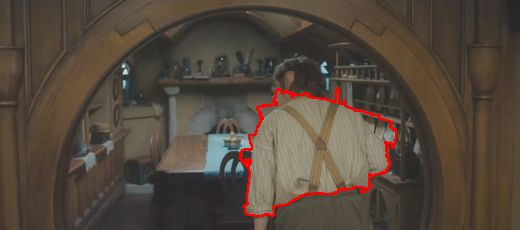}
  \includegraphics[clip,trim=0 17 0 10, width=\marHeight]{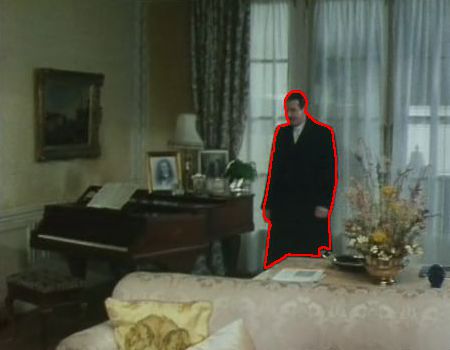}
   \includegraphics[clip,trim=0 17 0 10, width=\marHeight]{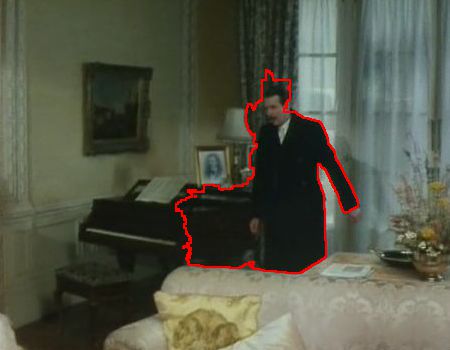}
   \includegraphics[clip,trim=0 17 0 10, width=\marHeight]{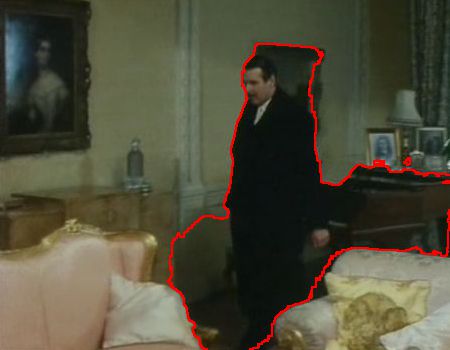}
   \includegraphics[clip,trim=0 17 0 10, width=\marHeight]{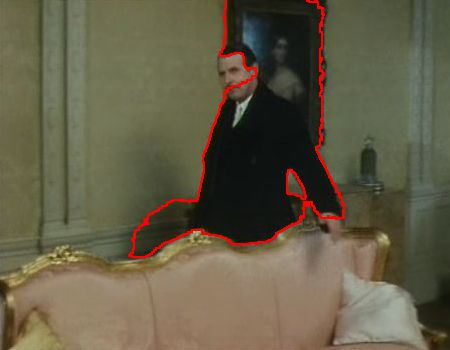}\\
  \includegraphics[clip,trim=36 0 12 0, width=\hobHeight]{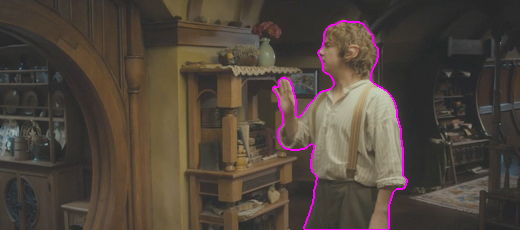}
  \includegraphics[clip,trim=36 0 12 0, width=\hobHeight]{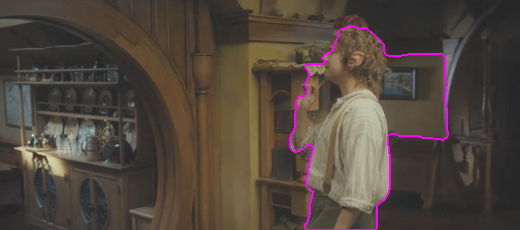}
  \includegraphics[clip,trim=36 0 12 0,width=\hobHeight]{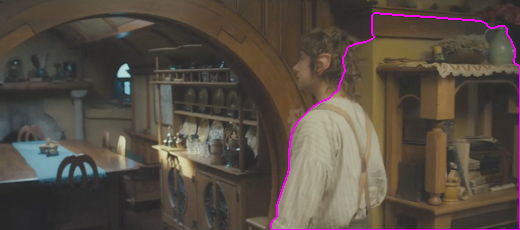}
  \includegraphics[clip,trim=36 0 12 0, width=\hobHeight]{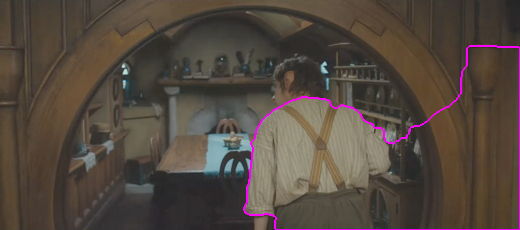}
   \includegraphics[clip,trim=0 17 0 10, width=\marHeight]{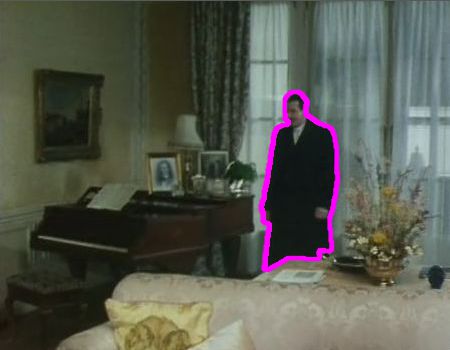}
  \includegraphics[clip,trim=0 17 0 10, width=\marHeight]{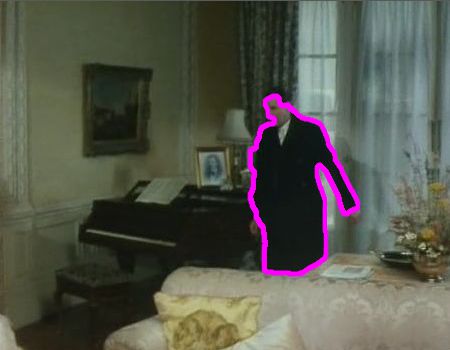}
  \includegraphics[clip,trim=0 17 0 10, width=\marHeight]{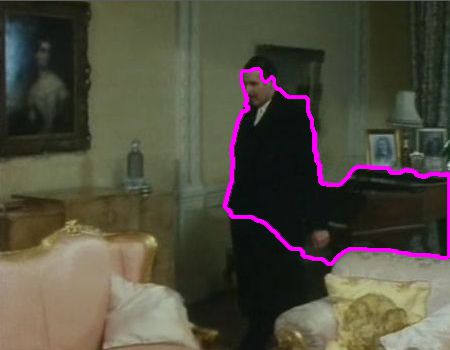}
  \includegraphics[clip,trim=0 17 0 10, width=\marHeight]{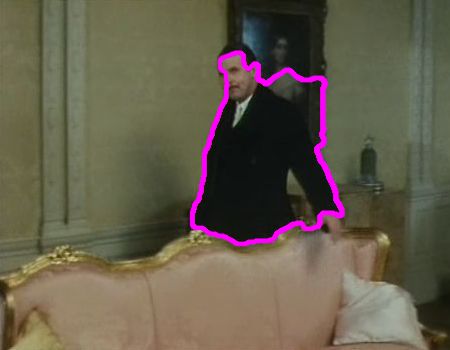}\\
  \includegraphics[clip,trim=150 0 50 0, width=\hobHeight]{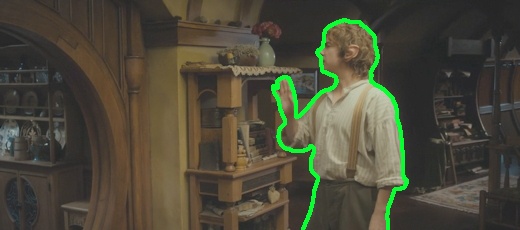}
  \includegraphics[clip,trim=150 0 50 0, width=\hobHeight]{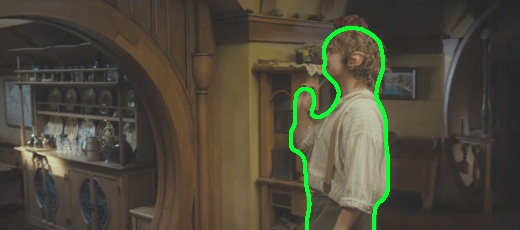}
  \includegraphics[clip,trim=150 0 50 0,width=\hobHeight]{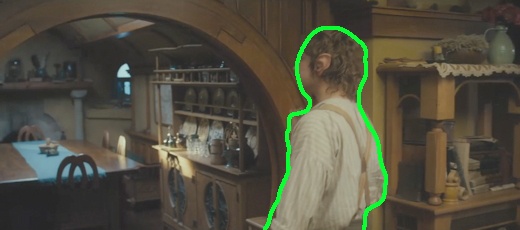}
  \includegraphics[clip,trim=150 0 50 0, width=\hobHeight]{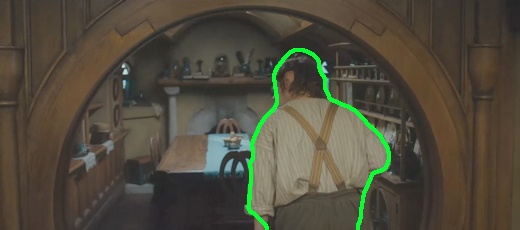}
  \includegraphics[clip,trim=0 17 0 10, width=\marHeight]{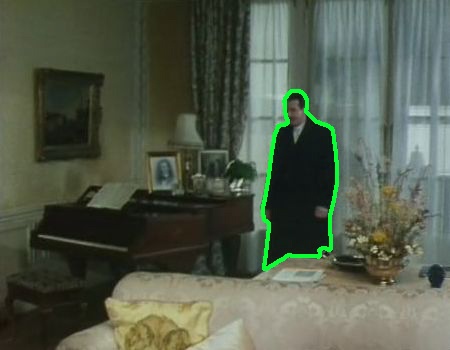}
  \includegraphics[clip,trim=0 17 0 10, width=\marHeight]{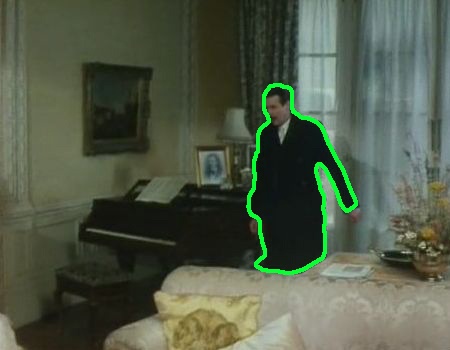}
  \includegraphics[clip,trim=0 17 0 10, width=\marHeight]{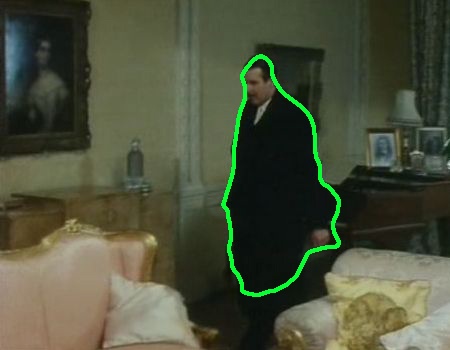}
  \includegraphics[clip,trim=0 17 0 10,
  width=\marHeight]{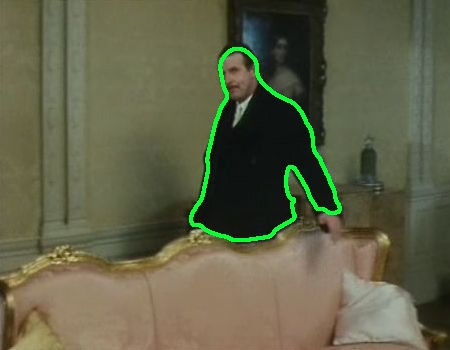}\\\vspace{0.05in}
  \def\fPathfatladyours{figures2/lady2/ours}
  \def\fPathfatladysapiro{figures2/lady2/sapiro}
  \def\fPathfatladyST{figures2/lady2/ST}
  \def\fPathpsyours{figures2/psy/ours}
  \def\fPathpsysapiro{figures2/psy/sapiro}
  \def\fPathpsyST{figures2/psy/ST}
  \def\fWidthnew{0.77in}
  \includegraphics[clip,trim=0 0 0 0, width=\fWidthnew]{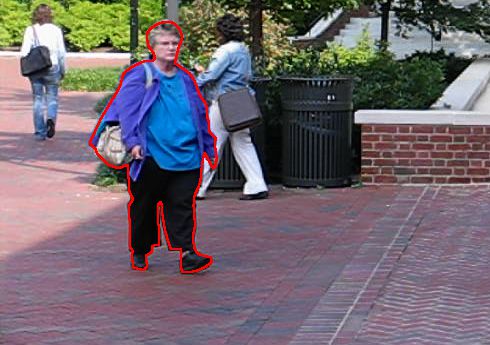}
  \includegraphics[clip,trim=0 0 0 0, width=\fWidthnew]{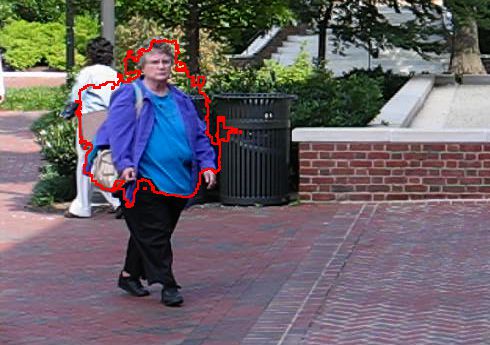}
  \includegraphics[clip,trim=0 0 0 0, width=\fWidthnew]{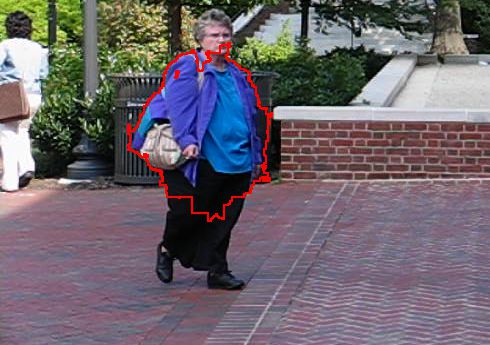}
  \includegraphics[clip,trim=0 0 0 0, width=\fWidthnew]{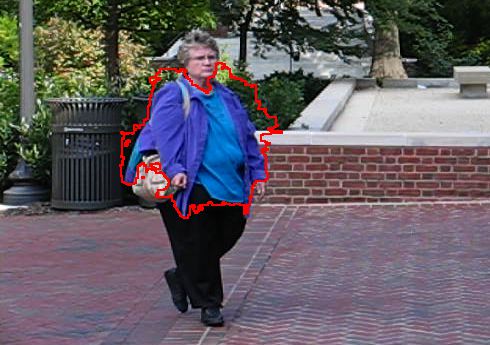}
  \includegraphics[clip,trim=0 0 0 0, width=\fWidthnew]{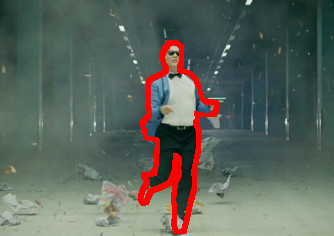}
  \includegraphics[clip,trim=0 0 0 0, width=\fWidthnew]{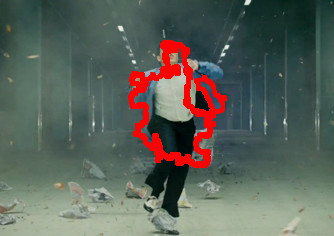}
  \includegraphics[clip,trim=0 0 0 0, width=\fWidthnew]{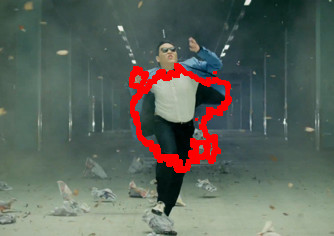}
  \includegraphics[clip,trim=0 0 0 0, width=\fWidthnew]{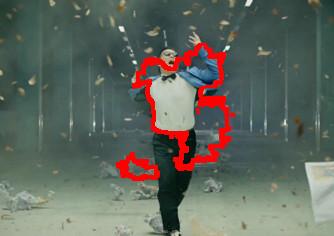}\\
  \includegraphics[clip,trim=0 0 0 0, width=\fWidthnew]{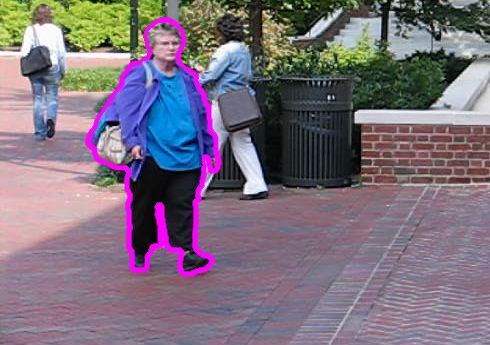}
  \includegraphics[clip,trim=0 0 0 0, width=\fWidthnew]{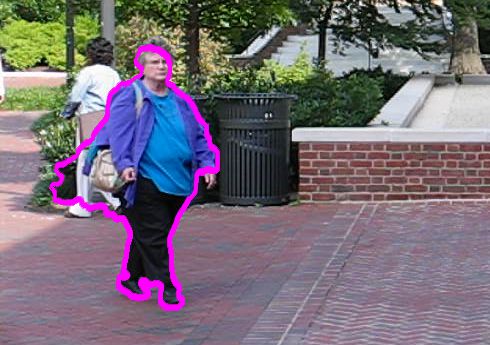}
  \includegraphics[clip,trim=0 0 0 0, width=\fWidthnew]{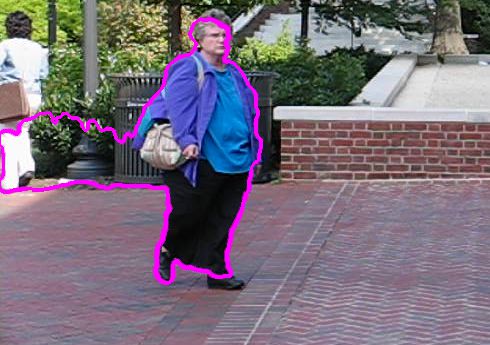}
  \includegraphics[clip,trim=0 0 0 0, width=\fWidthnew]{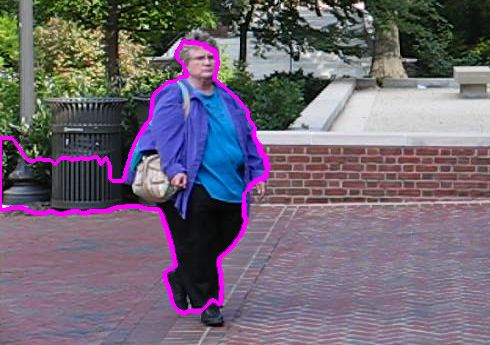}
  \includegraphics[clip,trim=0 0 0 0, width=\fWidthnew]{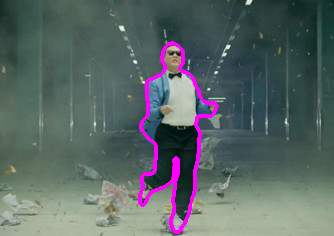}
  \includegraphics[clip,trim=0 0 0 0, width=\fWidthnew]{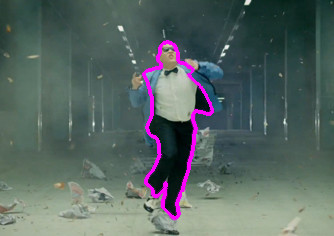}
  \includegraphics[clip,trim=0 0 0 0, width=\fWidthnew]{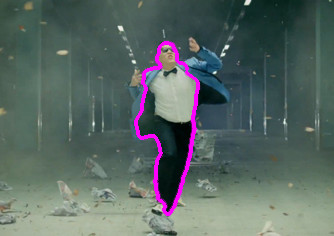}
  \includegraphics[clip,trim=0 0 0 0, width=\fWidthnew]{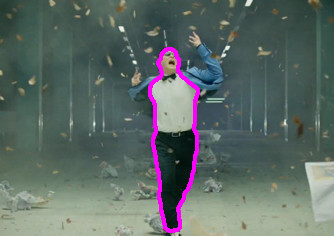}\\
  \includegraphics[clip,trim=0 0 0 0, width=\fWidthnew]{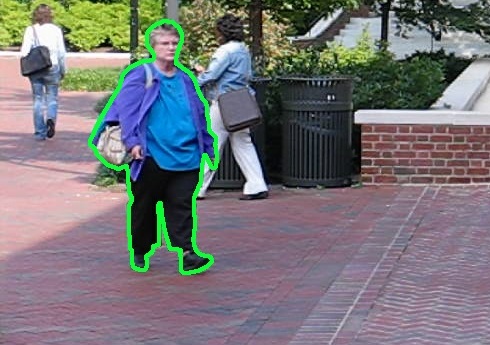}
  \includegraphics[clip,trim=0 0 0 0, width=\fWidthnew]{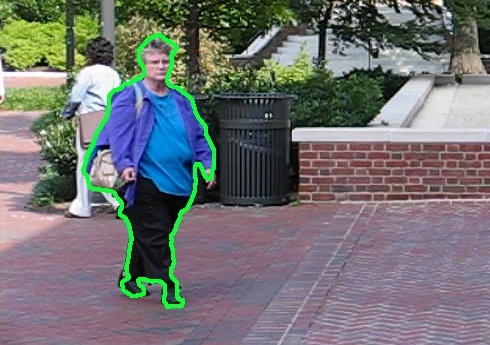}
  \includegraphics[clip,trim=0 0 0 0, width=\fWidthnew]{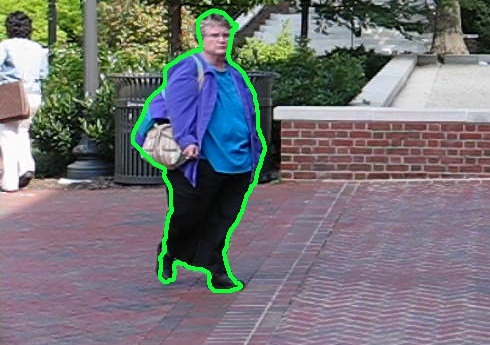}
  \includegraphics[clip,trim=0 0 0 0, width=\fWidthnew]{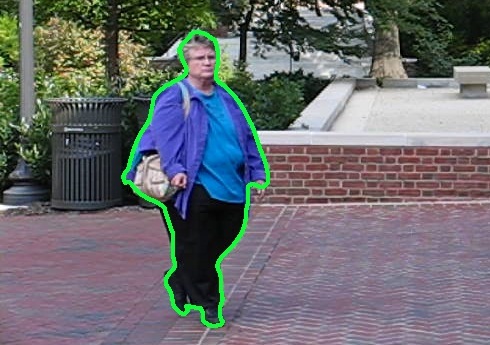}
  \includegraphics[clip,trim=0 0 0 0, width=\fWidthnew]{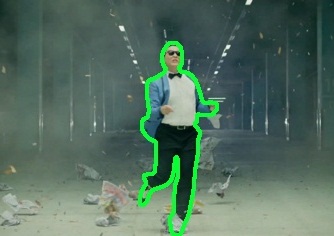}
  \includegraphics[clip,trim=0 0 0 0, width=\fWidthnew]{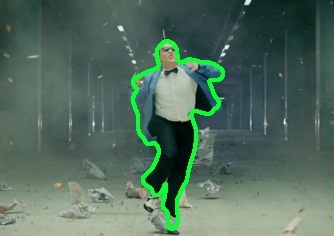}
  \includegraphics[clip,trim=0 0 0 0, width=\fWidthnew]{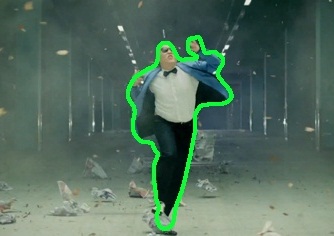}
  \includegraphics[clip,trim=0 0 0 0, width=\fWidthnew]{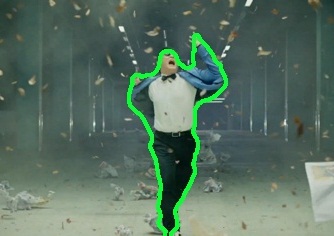}
  \caption{{\bf Occlusions/dis-occlusions, violations of brightness
      constancy, and foreground/background not easily separable.}
    [Top]: Scribbles, [Middle]: Adobe After Effects 2013, [Bottom]:
    proposed method. Methods based on foreground/background image
    statistic discrimination leak into the background. Note 4 (out of
    about 100-200) frames are selected for display in each sequence (see
    video on website).}
  \label{fig:expt_lady_mercedes}
\end{figure*}

\begin{table}
  \centering
  \footnotesize
  \begin{tabular}{|l|c|c|c|} \hline
    Sequence & Scribbles \cite{Fan:PAMI12} & Adobe Effects 2013
    \cite{bai2009video} & Ours \\ \hline\hline
    Library  & 0.8926 & 0.9193 & 0.9654  \\ \hline
    Fish &  0.9239	& 0.9513	& 0.9792 \\ \hline
    Skater & 0.8884 &0.6993 & 0.9086 \\ \hline
    Lady & 0.2986	& 0.8243	& 0.9508 \\ \hline
    Station &0.5367 &0.8258 &0.9216  \\ \hline
    Hobbit & 0.7312	& 0.5884	& 0.9335 \\ \hline
    Marple & 0.6942	& 0.8013	& 0.9186 \\ \hline
    Lady 2 & 0.7457 & 0.7909 & 0.9584 \\ \hline
    Psy & 0.6163 & 0.8845 & 0.9329 \\ \hline
  \end{tabular}
  \caption{{\bf Quantitative performance analysis.} Average F-measure
    (over all frames) computed from ground truth are shown. Larger
    F-measure means better performance.}
  \label{fig:quantative_assesment}
\end{table}

\begin{figure}
  \centering
  \includegraphics[totalheight=1.7in]{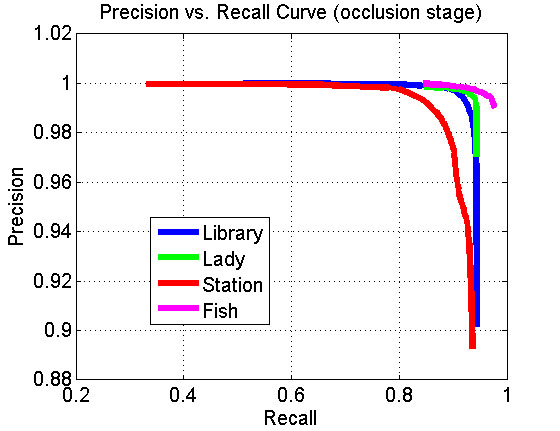}
  \includegraphics[totalheight=1.7in]{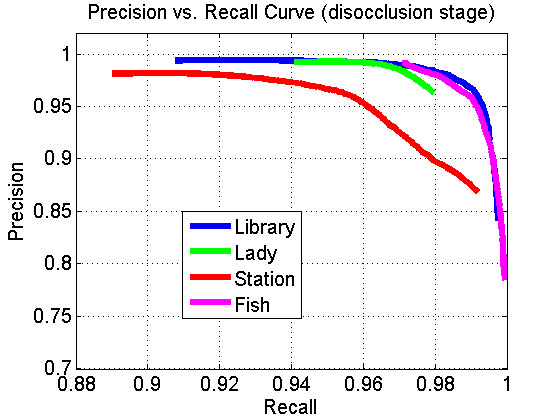}
  \caption{{\bf Sensitivity of Key Parameters}. The figure shows a
    quantitative assessment of the sensitivity of the key parameters
    (i.e., thresholds), $\beta_o$ and $\beta_d$, of the proposed
    algorithm for the occlusion and disocclusion stages in the
    sequences above. The Precision/Recall curves indicate that the
    parameters are robust to a wide range of thresholds that result in
    high values of precision and recall.}
  \label{fig:roc}
\end{figure}

In Figure~\ref{fig:roc}, we show a quantitative analysis of the
sensitivity of the key parameters of the proposed method. We analyze
the sensitivity of the thresholds $\beta_o$ and $\beta_d$ in the occlusion and
disocclusion detection stages using an precision / recall (PR)
curve. For four image sequences, we choose a pair of images so that
significant occlusion and disocclusion are present between the frames,
and significant deformation and motion is present. Typically, the pair
is separated by 5 frames on these sequences that have a frame rate of
30 frames per second. Given a hand cutout in the first frame, we run
our algorithm (both occlusion and disocclusion stages) to obtain the
cutout in the next frame. The first image in Figure~\ref{fig:roc}
shows the PR curve as the parameter $\beta_o$ is varied between its valid
range (the minimum value of the residual, $\mbox{Res}$, and its
maximum value), and the threshold of the disocclusion stage $\beta_d$ is
kept fixed. The second image in Figure~\ref{fig:roc} shows the
precision / recall curve as the parameter $\beta_d$ in the disocclusion
stage is varied between its valid range (the minimum and maximum value
of $p$), and the threshold in the occlusion stage $\beta_o$ is kept
fixed. Note high precision and recall is maintained for a wide range
of thresholds.

Lastly, we state the running time of our algorithm on a standard Intel
2.8GHz dual core processor. Note that the speed will depend on a
variety of factors such as the size of the object and amount of
deformation between frames. On HD 720 video, it is on average 8
seconds per frame for sequences in Fig.~\ref{fig:expt_lady_mercedes}
(in C++), while AAE takes 1 second. Speed-ups are possible, e.g., the
joint velocity and occlusion computation can be sped up using a
multi-scale procedure.

\section{Conclusion}
The proposed technique for shape tracking is based on jointly matching
shape and complex radiance (defined as a function on the region) of
the object across frames. Self-occlusions and dis-occlusions pose a
challenge for joint shape/appearance tracking, which were modeled and
computed in a principled framework in this work. 

In order to compute self-occlusions and the warp of a template to the
next frame, a joint energy was formulated, and a novel general
optimization scheme was derived that has an automatic coarse-to-fine
property, which is extremely useful in tracking. The method was based
on constructing a novel infinite dimensional Riemannian manifold of
parameterized regions and a novel Sobolev-type metric. The
optimization scheme is a gradient descent with respect to the
Sobolev-type metric, and empirical verification of the coarse-to-fine
property was given.

Experiments demonstrated the criticality of modeling occlusions and
dis-occlusions.  Comparison to recent methods built on global image
statistics foreground/background separation and joint shape/appearance
modeling without occlusion modeling demonstrated the effectiveness of
the proposed algorithm in situations of complex object/background
radiance, and self-occlusions/dis-occlusions.

Future work includes full occlusions of the object by other objects,
and improving dis-occlusion detection.

\appendices

\section{Computing Region-Based Sobolev Gradients}
\label{app:sobolev_gradients}

We now show how to compute the gradient of an energy with respect to
the Sobolev inner product defined in \eqref{eq:sobolev_inner}. For
generality, we compute the gradient of 
\begin{equation}
  E(w) = \int_{R} f( w(x), x ) \ud x
\end{equation}
where $f : \R^2 \times \R^2 \to \R$. The directional derivative in the
direction $h: R\to\R^2$ is
\begin{align}
  \ud E(w) \cdot h &= \int_{R} f_1 (w(x), x) \cdot h(x) \ud x \\
  &= \int_{w(R)} f_1(x,w^{-1}(x)) \cdot h\circ w^{-1}(x) \det{(\nabla w(x))} \ud x
\end{align}
and since by definition $\ud E(w)\cdot h = \ip{G}{h}{w}$ for all $h\in
T_wM$, where $G=\nabla_w E$ is the gradient with respect to the
Sobolev inner product, we have that
\begin{equation}
  \int_{w(R)} f_1(x,w^{-1}(x)) \cdot \hat h(x) \det{(\nabla w(x))} \ud x
  = \mean{G} \cdot \mean{\hat h} + \alpha \int_{w(R)} \mbox{tr}\left\{
    \nabla G(x)^T\nabla \hat h(x) \right\} \ud x
\end{equation}
By integrating by parts, one finds that
\begin{multline}
  \int_{w(R)} f_1(x,w^{-1}(x)) \cdot \hat h(x) \det{(\nabla w(x))} \ud x 
  = \\ \alpha \int_{\partial w(R)} (\nabla G(x) \cdot N)\cdot \hat h(x) \ud x - 
  \int_{w(R)} \left( \frac{1}{|w(R)|}\mean{G} -\alpha \Delta G(x) \right) \cdot 
  \hat h(x) \ud x.
\end{multline}
Therefore, $G$ can be obtained by solving
\begin{equation}
  \begin{cases}
    \frac{1}{w(R)|} \mean{G} -\alpha \Delta G(x) = f_1(x,w^{-1}(x)) \det{(\nabla w(x))}
    & x\in w(R) \\
    \nabla G(x) \cdot N = 0  & x\in \partial w(R)
  \end{cases}.
\end{equation}
Integrating both sides of the first equation above over $R$, we find that 
\begin{equation}
  \mean{G} = \int_{R} f_1(x,w^{-1}(x)) \det{(\nabla w(x))} \ud x.
\end{equation}
Therefore, the solution for $G$ is expressed as
\begin{equation}
  G = \mean{G} + \frac{1}{\alpha} \tilde G
\end{equation}
where $\tilde G$ (independent of $\alpha$) satisfies
\begin{equation}\label{eq:deformation_SG}
  \begin{cases}
    -\Delta \tilde G(x) = f_1(x,w^{-1}(x)) \det{(\nabla w(x))} -
    \mean{ f_1(\cdot,w^{-1}(\cdot) \det{(\nabla w(\cdot))} )
    } & x\in R \\
    \nabla \tilde G(x) \cdot N = 0  & x\in \partial w(R) \\
    \mean{\tilde G}=0 &
  \end{cases}.
\end{equation}

We consider $f$ of the form
\begin{equation}
  f( y, z ) = \frac 1 2 \rho( |I(y)-a(z)|^2 ) \chi_{O}(z)
\end{equation}
where $\rho : \R \to \R^+$. This gives
\begin{equation}
  f_1(y,z) = \rho'( |I(y)-a(z)|^2 ) (I(y)-a(z))\nabla I(y) \chi_{O}(z).
\end{equation}

\section{Numerical Implementation}
\label{app:numerics}

\subsection{Sobolev Gradient Computation}
\label{app:numerics_Poisson}

We show how to discretize \eqref{eq:deformation_SG}, the Poisson
equation. Let
\begin{equation}
F(x) = f_1(x,w^{-1}(x))\det{(\nabla w^{-1}(x))}^{-1} -
\mean{ f_1(\cdot,w^{-1}(\cdot)) \det{(\nabla w^{-1}(x))}^{-1} },
\end{equation}
then the discretization of the Laplacian is
\begin{equation}
  -\Delta \tilde G(x) = -\sum_{y\sim x} \tilde G(y) - \tilde G(x) =F(x),
\end{equation}
where $y\sim x$ indicates that $y$ is a 4-neighbor of
$x$. Discretizing the boundary condition $\nabla \tilde G(x)\cdot
N=\tilde G(y)-\tilde G(x) = 0$, when $y\sim x$,  and substituting it
above, we have that
\begin{equation}
  -\sum_{y\sim x, y\in R} \tilde G(y) - \tilde G(x) =F(x),
\end{equation}
and this can be solved using the conjugate gradient method. Indeed,
the operator on the left is positive definite on the set of mean zero
vector fields. One starts with an initialization such that
$\mean{\tilde G} = 0$.

\subsection{Discretization of Transport Equations}
We describe the discretizations of the transport equations used in the
gradient descent of the warp $\phi_{\tau}^{-1}$ and the warped region
$R_{\tau}$, which for the most part, are standard.

Let $\Psi_{\tau} : \Omega \to \R$ denote the level set function at
time $\tau$ such that $\{ x \in \Omega \, : \, \Psi_{\tau}(x) < 0 \} =
R_{\tau}$. The level set evolution equation
\eqref{eq:level_set_evolution} (shown here again for convenience):
\begin{equation}
  \partial_{\tau} \Psi_{\tau}(x) = \nabla G_{\tau}(x)\cdot \nabla \Psi_{\tau}(x)
\end{equation}
is discretized using an up-winding difference scheme:
\begin{equation} \label{eq:upwind_levelset}
  \Psi_{\tau_{i+1}}(x) = \Psi_{\tau_{i}}(x) + \Delta t \left(
    G_{\tau_i}^1(x) D_{x_1}[\Psi_{\tau_{i}}, G_{\tau_i}^1, x] + 
    G_{\tau_i}^2(x) D_{x_2}[\Psi_{\tau_{i}}, G_{\tau_i}^2, x]
  \right)
\end{equation}
where $\Delta t>0$ is the time step, and
\begin{equation}
  D_{x_1}[\Psi_{\tau_{i}}, G_{\tau_i}^1, x] = 
  \begin{cases}
    D_{x_1}^+\Psi_{\tau_{i}}(x)  & \mbox{ if } G_{\tau_i}^1(x) < 0 \\
    D_{x_1}^-\Psi_{\tau_{i}}(x)  & \mbox{ if } G_{\tau_i}^1(x) \geq 0 
  \end{cases}
\end{equation}
where $D^+_{x_j}$ ($D^-_{x_j}$) denotes the forward (backward, resp.)
difference with respect to the $j^{\text{th}}$ coordinate, and
$G_{\tau}(x) = ( G_{\tau}^1(x), G_{\tau}^2(x) )$. Note that
$G_{\tau}|\partial R_{\tau}$ is extended to the narrowband of the
level set function by choosing $G_{\tau}$ of a point $x$ in the
narrowband to be the same as that of the closest point on $\partial
R_{\tau}$ from $x$.

The discretization of the transport equation
\eqref{eq:backward_map_evolution} for the backward map:
\begin{equation}
  \partial_{\tau} \phi_{\tau}^{-1}(x) = \nabla G_{\tau}(x)\cdot \nabla \phi_{\tau}^{-1}(x)
\end{equation}
is
\begin{equation} \label{eq:appearance_discrete_update}
  \phi^{-1}_{\tau_{i+1}}(x) = 
  \begin{cases}
    \phi^{-1}_{\tau_i}(x) +\Delta t \left(  
      G_{\tau_i}^1(x) D_{x_1}[\phi^{-1}_{\tau_{i}}, G_{\tau_i}^1, x] + 
      G_{\tau_i}^2(x) D_{x_2}[\phi^{-1}_{\tau_{i}}, G_{\tau_i}^2, x]
    \right), & x\in R_{\tau_{i+1}} \cap R_{\tau_i} \\
    \frac{ \sum_{ y \in N_x \cap R_{\tau_i} } d_{\Psi_{\tau_i}}(x,y) \phi^{-1}_{\tau_i}(y) } 
    { \sum_{ y \in N_x \cap R_{\tau_i} } d_{\Psi_{\tau_i}}(x,y) },
      & x \in R_{\tau_{i+1}} \backslash R_{\tau_i}
  \end{cases}
\end{equation}
where $N_x$ denotes the eight neighbors of $x$, and
$d_{\Psi_{\tau_i}}(x,y)$ denotes the distance between $x$ and the zero
crossing of the level set $\Psi_{\tau_i}$ between $x$ and $y$ (zero if
there is no zero crossing). In the computation of the forward/backward
difference, if the relevant neighbor of $x$ is not in $R_{\tau_i}$,
then the difference is set to zero.  It should be noted that the step
size is chosen to satisfy the stability criteria, which means that the
level set may not move more than one pixel and thus $x$ will always
have a neighbor that is in $R_{\tau_i}$, and so the second case in
\eqref{eq:appearance_discrete_update} is well-defined.  The step size
$\Delta t$ is chosen to satisfy $\Delta t < 0.5 / \max_{x\in
  R_{\tau_i}, j=1,2} |G_{\tau_i}^j(x)|$. 

\bibliographystyle{IEEEtran}
\bibliography{tracking}

\end{document}